\documentclass[twoside,11pt]{article}

\usepackage{rotating}
\usepackage{jmlr2e}
\usepackage{booktabs}
\usepackage{amsmath}
\usepackage{algorithm}
\usepackage{algorithmic}
\usepackage{graphicx}
\usepackage{subfigure}

\usepackage{float}

\begin{document}\label{section1}

\title{Small Sample Learning in Big Data Era}

\author{\name Jun Shu \email xjtushujun@gmail.com
       \AND
       \name Zongben Xu \email zbxu@mail.xjtu.edu.cn
       \AND
       \name Deyu Meng  \email  dymeng@mail.xjtu.edu.cn\\
       \addr School of Mathematics and Statistics \\
        Ministry of Education Key Lab of Intelligent Networks and Network Security\\
       Xi'an Jiaotong University, Xi¡¯an, China}


\maketitle

\begin{abstract}
  As a promising area in artificial intelligence, a new learning paradigm, called Small Sample Learning~(SSL), has been attracting prominent research attention in the recent years. In this paper, we aim to present a survey to comprehensively introduce the current techniques proposed on this topic. Specifically, current SSL techniques can be mainly divided into two categories. The first category of SSL approaches can be called ``concept learning", which emphasizes learning new concepts from only few related observations. The purpose is mainly to simulate human learning behaviors like recognition, generation, imagination, synthesis and analysis. The second category is called ``experience learning", which usually co-exists with the large sample learning manner of conventional machine learning. This category mainly focuses on learning with insufficient samples, and can also be called small data learning in some literatures. More extensive surveys on both categories of SSL techniques are introduced and some neuroscience evidences are provided to clarify the rationality of the entire SSL regime, and the relationship with human learning process. Some discussions on the main challenges and possible future research directions along this line are also presented.
  \end{abstract}


\section{Introduction} \label{section1}

Learning could be understood as a behavior that a person/system improves him(her) self/itself by interacting with the outside environment and introspecting in the internal model of the world, so as to improve his(her)/its cognition, adaptation and regulation capability to environment. Machine learning aims to simulate such behavior by computers for executing certain tasks, and generally contain the following implementation elements. Firstly, a performance measure is generally required to be defined, and then useful information are exploited from the pre-collected historical experience (training data) and pre-known prior knowledge under the criterion of maximizing the performance measure to train a good learner to help analyze future data~\citep{jordan2015machine}. Learning is substantiated to be beneficial to tasks like recognition, causality, inference, understanding, etc., and has achieved extraordinary performance among various practical tasks including image classification~\citep{Krizhevsky2012Alex, He2016Res}, speech recognition~\citep{Hinton2012, Mikolov2011,Sainath2013}, sentiment analysis~\citep{Cambria2013}, machine translate~\citep{Sutskever2014}, Atari video games~\citep{Mnih2015}, Go games~\citep{Silver2016,Silver2017}, Texas Hold'em poker~\citep{Moravvcik2017}, skin cancer diagnosis~\citep{Esteva2017}, quantum many-body problem~\citep{Carleo2017}, etc..

\begin{figure}[t]
\setlength{\abovecaptionskip}{0.cm}
\setlength{\belowcaptionskip}{-1.cm}
  \centering
  \subfigure[BPL]{
    \label{fig:subfig:a} 
    \includegraphics[width=2.4in]{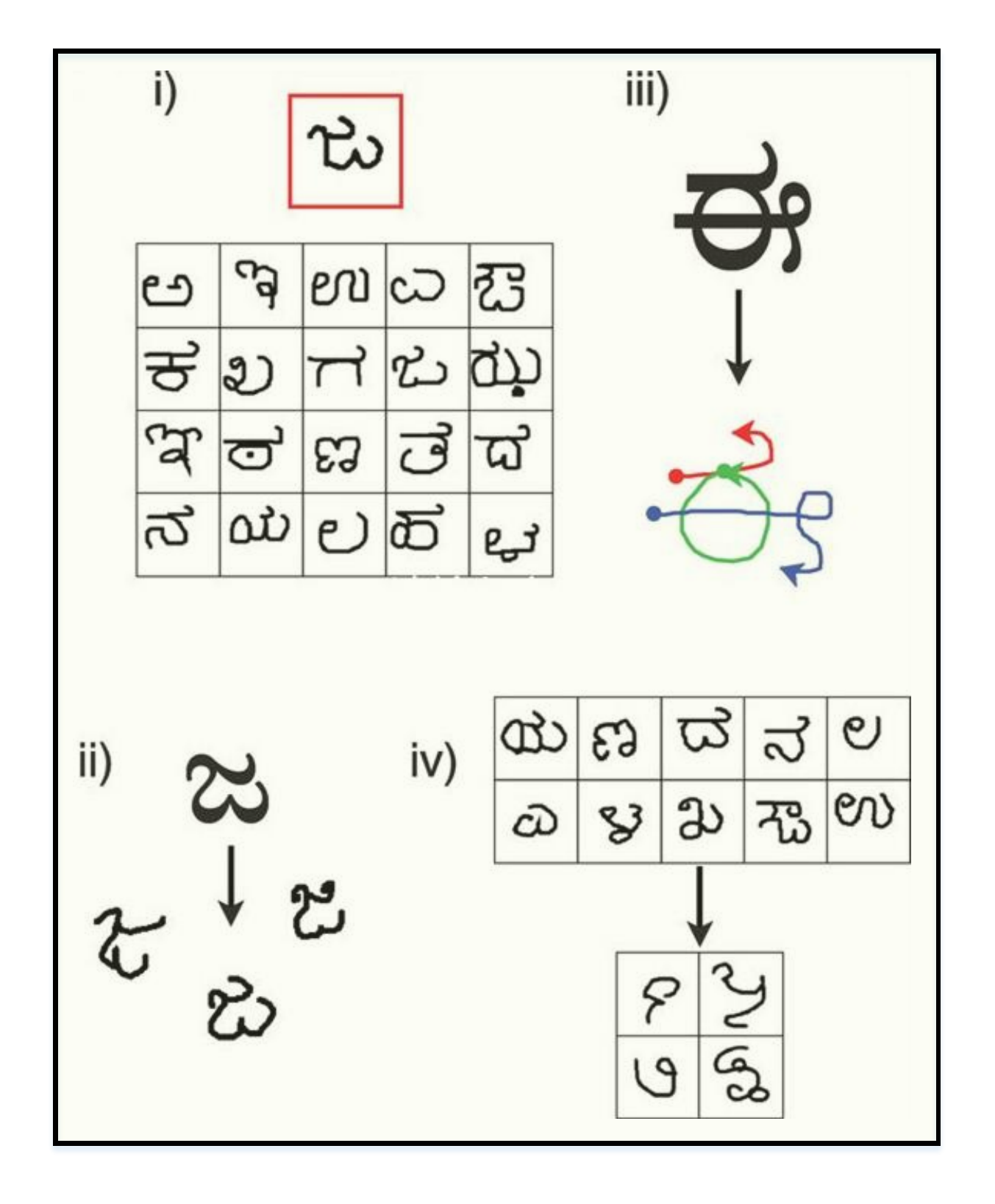}}
  \hspace{1in}
  \subfigure[RCN]{
    \label{fig:subfig:b} 
\includegraphics[width=2.4in]{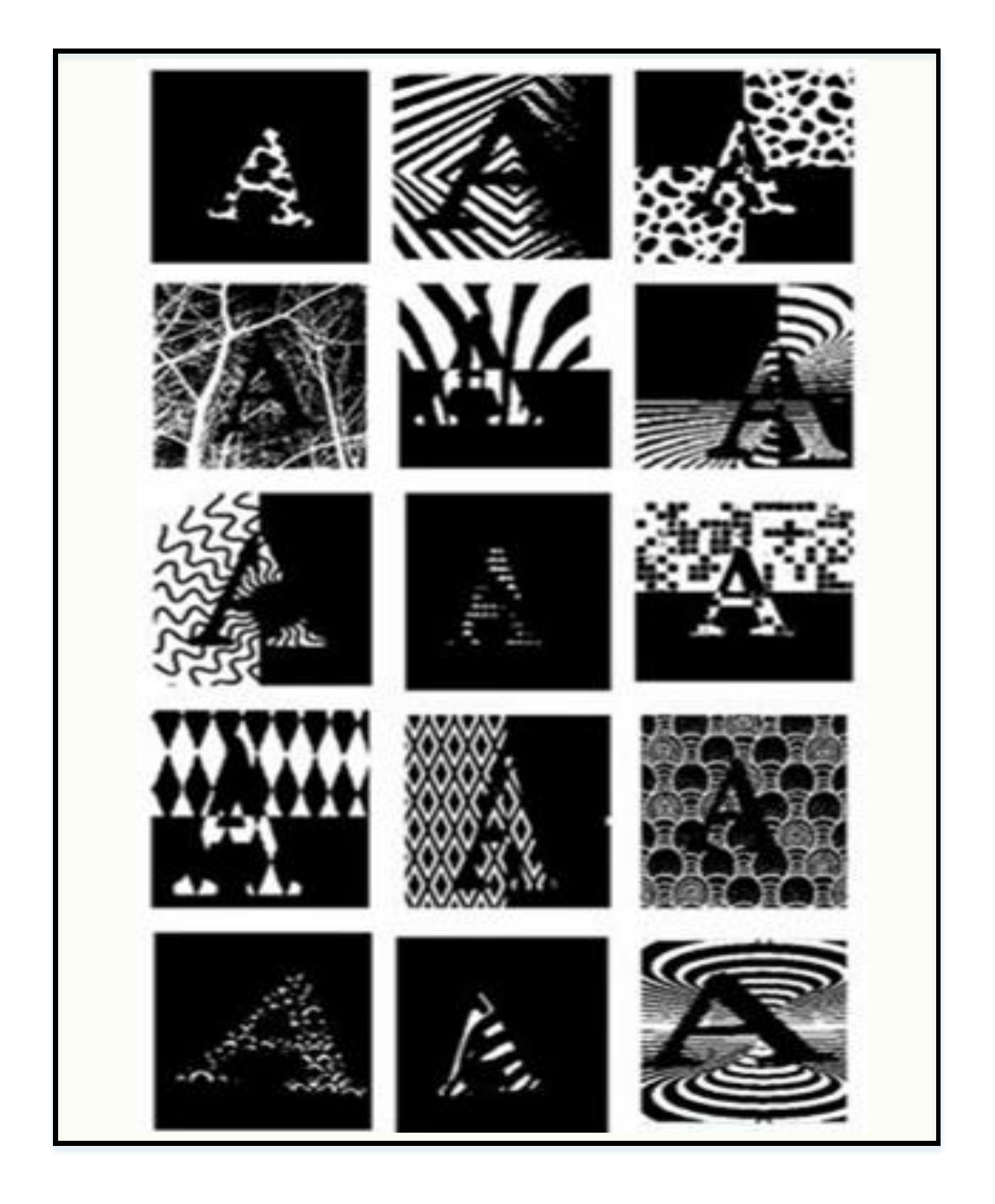}}
  \vspace{-4mm} \caption{\small Examples of Small Sample Learning (SSL). (a) and (b) are reproduced from \citep{Lake2015} and \citep{George2017}, respectively.
  (a) Demonstration of Bayesian program learning(BPL). Provided only a single example (red boxes), BPL~\citep{Lake2015} can rapidly learn the new concept (i.e., the generation procedure of character) with prior knowledge embedded into models to classify new examples, generate new examples, parse an object into parts and generate new concepts from related concepts.
  (b) Demonstration of Recursive Cortical Network (RCN). Given the same character ``A" in a wide variety of appearances shape, RCN~\citep{George2017} can parse ``A" with contours and surfaces, scene context and background with lateral connections, and achieve higher accuracy compared with CNN with fewer samples.}
  \label{f1} 
\end{figure}

In the recent decades, machine learning has made significant progress and obtained impressively good performance on various tasks, which makes this line of approaches become the most highlighted techniques of the entire artificial intelligence field. While such success seems to make people more and more optimistic to the power of current machine learning approaches, many researchers and engineers began to recognize that most latest progresses of machine learning are highly dependent on the premise of large number of input samples (generally with annotations). Such kind of learning manner can be called Large Sample Learning (LSL) for notation convenience. The real cases, however, are always deviated from such ideal circumstances, and generally with the characteristic of Small Sample Learning (SSL). Specifically, there are mainly two categories of SSL scenarios. The first can be called \textbf{\emph{concept learning}}, aiming to recognize and form never-seen new concepts through only few observations on these concepts by associating with previously learned knowledge of other ones. The other category is \textbf{\emph{experience learning}}, sometimes also called small data learning, mainly proposed from the opposite side of LSL, i.e., to carry out machine learning on the condition of lacking sufficient training samples.

As a fundamental and widely existed learning paradigm in real cases, the early attempts of SSL might be originated from the context of multimedia retrieval~\citep{Zhou2001}, and is gradually attracting increasing attention throughout various areas in the recent years
~\citep{Miller2000,bart2005cross,Fei-Fei2003,Fei-Fei2006}. A representative method is proposed by \citep{Lake2015}, achieving human-level performance on one-shot character classification task, and able to generate new examples of a concept trained from only one sample of the class, which are even indistinguishable from human behavior (see Fig.~\ref{fig:subfig:a}). Afterwards, \citet{Lake2016} proposed a ``characters challeng'' advancement. That is, after an AI system views only a single exemplar, the system should be able to distinguish novel instances of an unfamiliar handwritten character from others. Lately, AI startup Vicarious~\citep{George2017} claimed to outperform deep neural networks on challenging text recognition task with less one three-hundredth data and break the defense of modern text-based CAPTCHAs. In Vicarious's official blog (\url{https://www.vicarious.com/2017/10/26/common-sense-cortex-and-captcha/}), they explained that why the central problem in AI is to understand the letter ``A'' (see Fig.~\ref{fig:subfig:b}), and believed that ``for any program to handle letter forms with the flexibility that human beings do, it would have to possess full-scale artificial intelligence''.

Generally speaking, while human can very easily perform these tasks, they are difficult for classical AI systems. Human cognition is distinguished by his/her ability to rapidly constitute a new concept from only a handful example, through latently leveraging his/her possessed prior knowledge to enable flexible inductive inferences~\citep{Hassabis2017}. The concept takes an important role in human cognition~\citep{carey1999knowledge}, where our minds make inferences that appear to go far beyond the data available through learning concepts and grasping causal relations~\citep{tenenbaum2011grow}. For example, as shown in \citep{roymachines}, human can recognize a Segway even if he/she has seen it only once (i.e., \textbf{\emph{one-shot learning}}). This is because our mind can decompose the concept into a collection of known parts such as wheels a steering stick. Many variations of these components are already encoded in our memory. Hence, it is generally not difficult for a child to compile different models of Segways rapidly. Even in the extreme case that one has not seen the sample on the concept before~(i.e., \textbf{\emph{zero-shot learning}}), given a description of some attributes of a Segway, one can still guess how its components should be connected, and through using his/her previous possessed knowledge on these attributes, he/she can still recognize a Segway without seeing it. SSL tries to imitate human cognition intelligence to solve these hard tasks that classical AI system can hardly process with only few exemplars.

In summary, SSL is a fundamental and gradually more widespread new learning paradigm featured by few training examples, and aims to simulate human learning capabilities that rapidly discover and represent new concept with few observations, parse a scene into objects and relations, recombine these elements to synthesis new instances through imagination, and implement other small sample learning tasks. The following issues encountered by current machine learning approaches are promising to be alleviated by the future development of SSL techniques, which makes this research meaningful for exploration:

1) \textbf{Lack of labels due to high cost of human annotations.}
As aforementioned, LSL can achieve excellent performance in various tasks in the presence of large amount of high-quality training samples. For example, to learn a deep learning model with tens or even hundreds of layers and containing a huge number of model parameters, we need to pre-collect large amount of training samples which are labelled with fully ground-truth annotations~\citep{goodfellow2016deep}. Typical datasets so generated including PASCAL VOC~\citep{everingham2010pascal}, ImageNet~\citep{russakovsky2015imagenet, deng2009imagenet}, Microsoft COCO~\citep{lin2014microsoft} and many other known ones. In practice, however, it could be difficult to attain such high-quality annotations for many samples due to the high cost of data labeling process (e.g., small-scale event annotation from surveillance videos with crowded large-scale scenes \citep{jiang2014easy}) or lack of experts' experience (e.g., certain diseases in medical images \citep{fries2017swellshark}). Besides, many datasets are collected by crowdsourcing system or search engines for reducing human labor cost. They, however, inevitably contain large amount of low-quality annotations (i.e., coarse or even inaccurate annotations). This leads to the known issues of weakly supervised learning~\citep{zhou2017brief} or webly-supervised learning~\citep{chen2015webly}, attracting much research attention in recent years (Section \ref{WSL}). For example, the task of semantic segmentation~\citep{hong2017weakly} usually could only be implemented on pre-collected images with image-level labels, rather than expected pixel-level labels. In such case, even under large amount of training samples, the conventional end-to-end training manner for LSL still inclines to fail due to such coarse annotations.

2) \textbf{Long-tail distribution existed extensively in big data.}
The long tail phenomena appear in the dataset where a small number of objects/words/classes are very frequent,
while many more are rare~\citep{bengio2015sharing}. Taking the object recognition problem as an example~\citep{ouyang2016factors}, \cite{rahman2018zero} showed that for the known ILSVRC dataset~\citep{russakovsky2015imagenet}, instance numbers for all classes follows an evident long-tail distribution. In all 200 classes of this dataset, only 11 highly frequent classes cover 50\% amount of samples in the entire dataset, which makes a learner easily dominates its performance on these head classes while degrades its performance on other tail classes. A simple amelioration strategy is re-balancing training data like sampling examples from the rare classes more frequently~\citep{shen2016relay} or reducing the number of examples from the top numbered classes~\citep{he2009learning}. This strategy, however, is generally heuristic and suboptimal. The former manner tends to generate sample redundancy and encounters the problem of over-fitting to the rare classes, whereas the latter inclines to lose critical feature knowledge within the classes with more samples~\citep{wang2017learning}. Thus necessary SSL methods is expected to be helpful to alleviate such long-tail training issue by leveraging more beneficial prior knowledge of small-sample classes.

3) \textbf{Insufficient data for conventional LSL approaches.} Although we are in big data era, there still exist many domains lacking sufficient ideal training samples. For example, in intelligent medical diagnose issue, medical imaging data are much more difficult to be annotated with certain lesions in high-quality by common people without specific expertise compared to general images with trivial categories. Alternatively, new diseases consistently occur with few historical data, and rare diseases frequently occur with few cases, which can only obtain scarce training samples with accurate labels. Another example is intelligent communications, where systems should perform excellent transmission under very few pilot signals. In such cases, conventional LSL approaches can hardly perform and effective SSL techniques are urgently required.

4) \textbf{Arose from cognitive science studies.} Many scholars are attempting to achieve future AI by making machines really mimic humans for thinking and learning~\citep{russell2016artificial}. The main motivation of SSL is a manner to more or less construct such a learning paradigm, i.e., simulating humans to learn new concepts from few observations with strong generalization ability. Inspired from cognitive science studies, some progresses have been made on this point~\citep{Hassabis2017}. Recently, NIPS 2017 Workshop on Cognitively Informed Artificial Intelligence~(\url{https://sites.google.com/view/ciai2017/home}) made efforts to bring together cognitive scientists, neuroscientists, and machine learning researchers to discuss opportunities for improving AI by leveraging scientific understanding of human perception and cognition. Valuable knowledge and experiences from cognitive science are expected to feed AI and SSL, and inspire useful learning regimes with the strength and flexibility of the human cognitive architecture.
\begin{figure}[htb]
\centering
  \subfigure[Machine learning paradigm]{
  \label{fig2a}
  \includegraphics[width=4.5in]{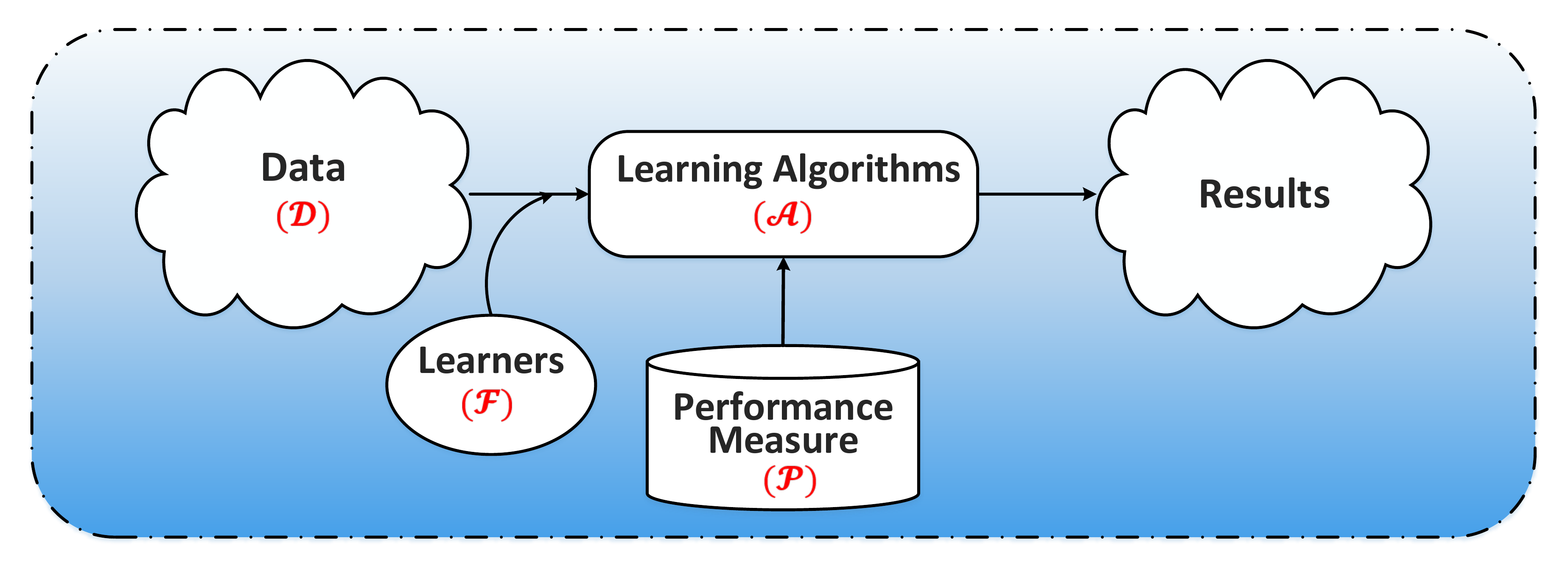}}\\
  \subfigure[Experience learning paradigm]{
  \label{fig2b}
   \includegraphics[width=4.5in]{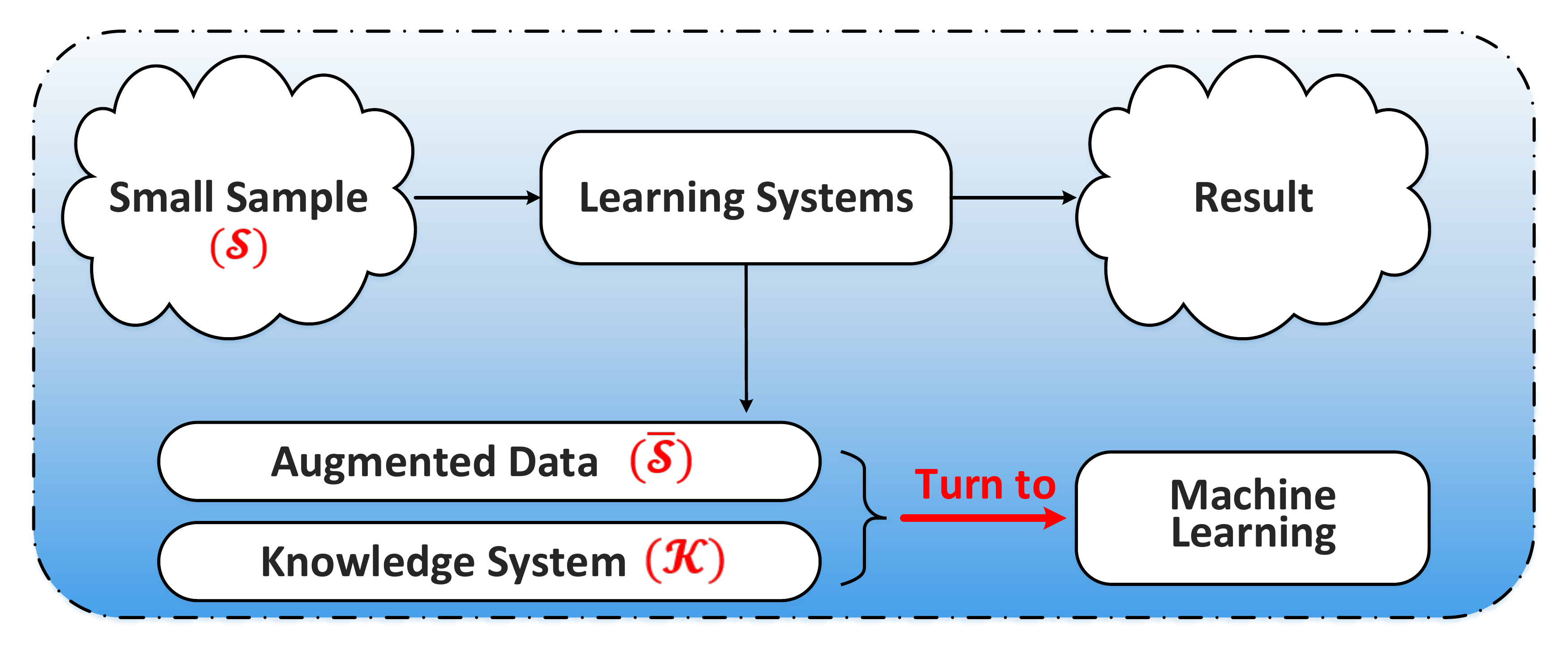}}\\
  \subfigure[Concept learning paradigm]{
  \label{fig2c}
 \includegraphics[width=4.5in]{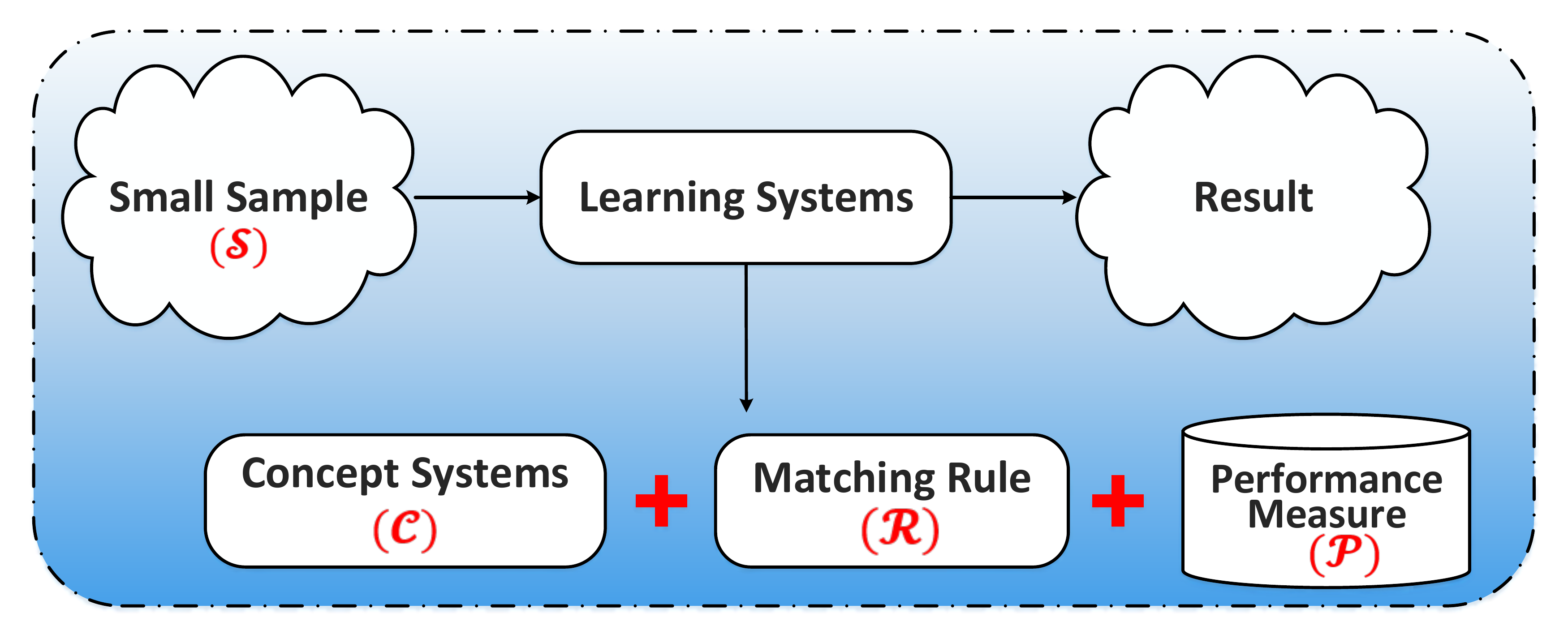}}\\
  \vspace{-4mm}\caption{\small Concept illustrations of (a) machine learning, (b)experience learning and (c) concept learning. Compared with conventional LSL, experience learning tries to turn SSL into LSL through employing augmented data $\overline{\mathcal{S}}$ and knowledge system $\mathcal{K}$ provided small sample set $\mathcal{S}$, and still retain good performance; and concept learning aims to associate concepts in the  concept system $\mathcal{C}$ with small sample set $\mathcal{S}$ through matching rule $\mathcal{R}$ to form a new comcept or complete a recognition task.
  }\label{f2}
\end{figure}

In this paper, we aim to present a comprehensive survey on the current developments of the SSL paradigm, and introduce some closely related research topics. We will try to make definitions on SSL-related concepts to make clear their meanings, so as to help avoid confused and messy utilization of these words in literatures. The relations of SSL to biological plausibility and some discussions on the future directions for SSL worthy to be investigated will also be presented.
The paper is organized as follows. Section \ref{section2} presents a definition for SSL, as well as its two categories of learning paradigms: concept learning and experience learning. Biological plausibility of SSL is also given in this section. Section \ref{section3} summarizes the recent techniques of concept learning, and then Section \ref{section4} surveys those of
experience learning. Section \ref{beyond} introduces some related research directions of SSL. Finally we make some discussions on future directions on SSL in Section \ref{section6} and conclude the full paper in Section \ref{section7}.

\section{Small Sample Learning}\label{section2}
In this section, we first present a formal definition for SSL, and then provide some neuroscience evidence to support the rationality of this learning paradigm.

\subsection{Definition of SSL}\label{section21}
To the best of our knowledge, the current research of SSL focuses on two learning aspects, \textbf{\emph{experience learning}} and \textbf{\emph{concept learning}}, which we will introduce in detail in the following. We thus try to interpret all these related concepts as well as SSL. Besides, some other related concepts, like $k$-shot learning, will also be clarified.

To start with, we try to interpret the task of machine learning based on the description in \citep{jordan2015machine} as follows:
\textbf{\emph{Learning Algorithm $\mathcal{A}$}} helps \textbf{\emph{Learners $\mathcal{F}$}} improve certain \textbf{\emph{performance measure $\mathcal{P}$}} when executing some tasks, through pre-collected experiential \textbf{\emph{Data $\mathcal{D}$}} (see Fig.\ref{fig2a}). Conceptually, machine learning algorithms can be viewed to search, through a large space of candidate learners, guided by training experience, a learner that optimizes the performance metric. This setting usually requires a large amount of labeled data for training a good learner.

\begin{figure}[htb]
\centering
  \subfigure[Representations from different domains]{
  \label{fig3a}
  \includegraphics[width=2.5in]{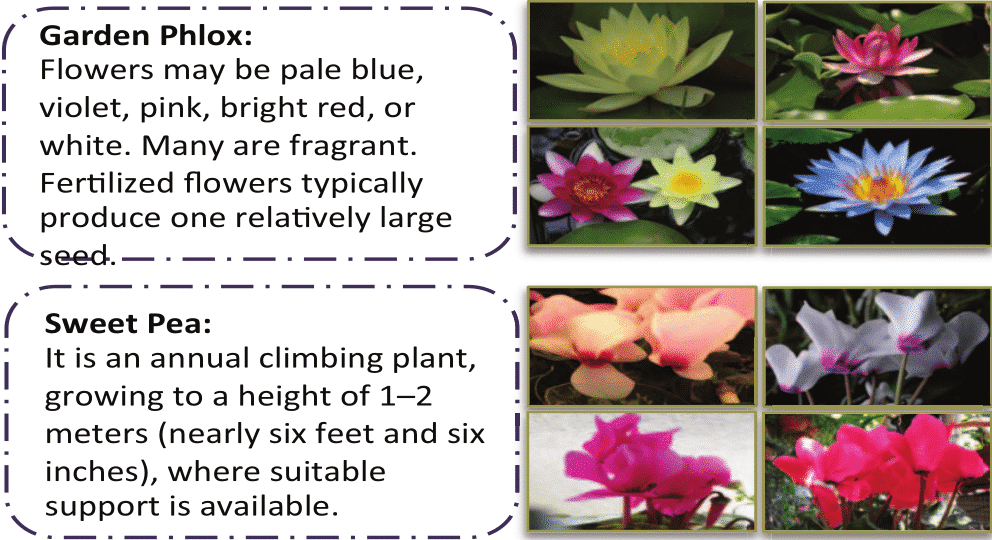}}
  \hspace{0.5in}
  \subfigure[Trained models]{
  \label{fig3b}
   \includegraphics[width=2.5in]{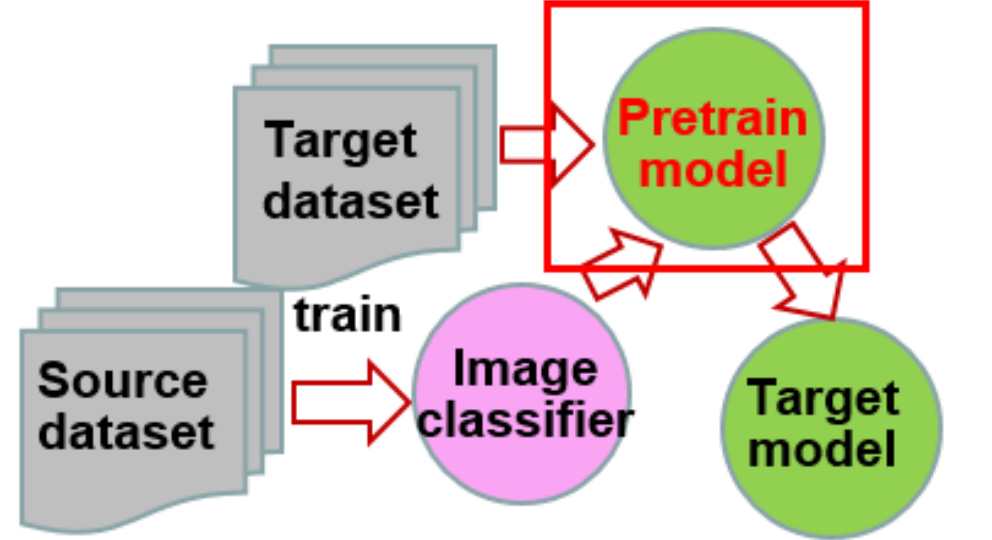}}\\
   \subfigure[Cognition knowledge on concepts]{
  \label{fig3c}
  \includegraphics[width=2.5in]{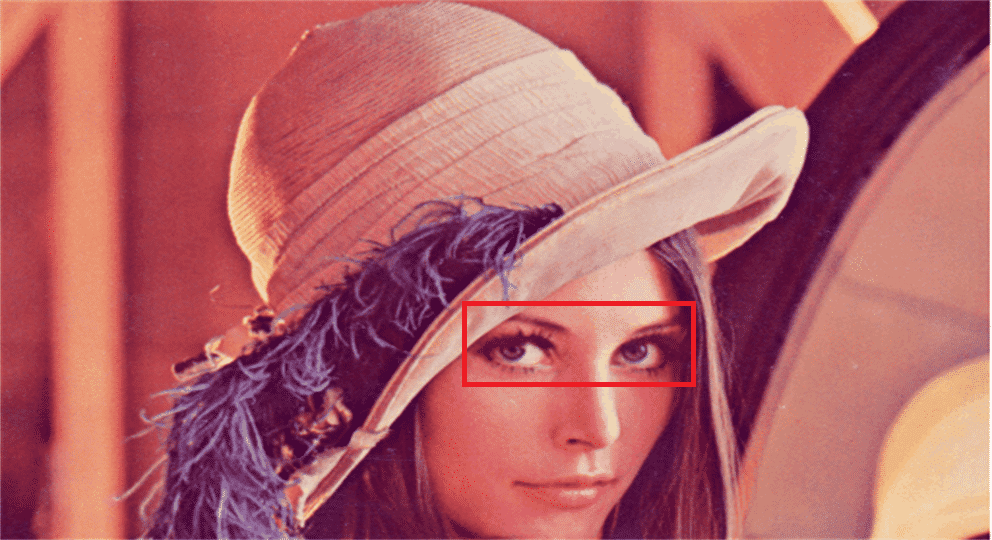}}
  \hspace{0.5in}
  \subfigure[Meta knowledge]{
  \label{fig3d}
   \includegraphics[width=2.5in]{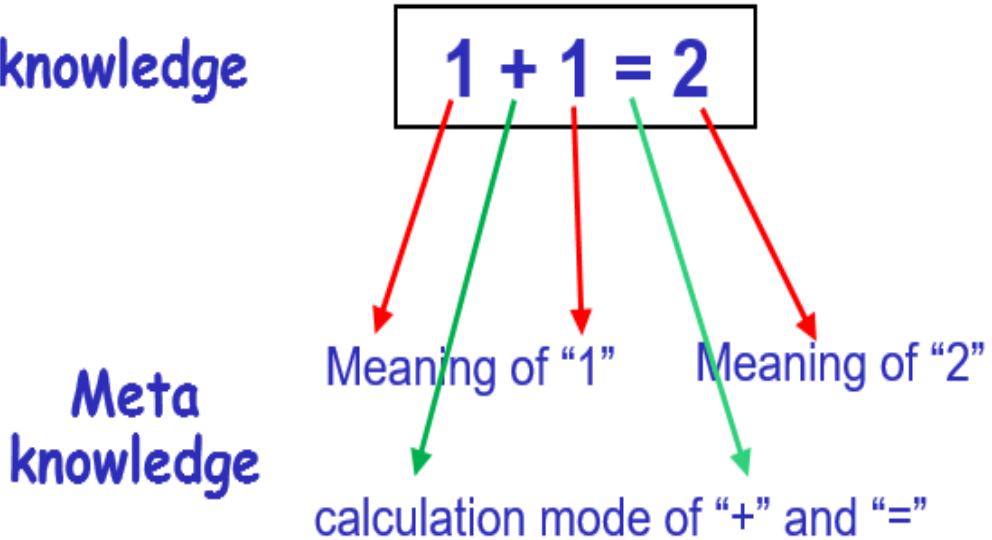}}\\
  \caption{\small Examples of the knowledge system.}\label{f3}
\end{figure}

\subsubsection{Experience Learning \& Concept Learning}
Experience learning is a specific SSL paradigm, in which samples directly related to tasks are highly insufficient. In other words, this kind of SSL regimes co-exists with Large Sample Learning (LSL), and its main goal is to reduce or meet the requirements for the number of sample of LSL methods. Given \textbf{\emph{small sample input $\mathcal{S}$}}, the main strategies employed in this research mainly include two categories of approaches by making use of \textbf{\emph{augmented data $\mathcal{\overline{S}}$}} and \textbf{\emph{knowledge system $\mathcal{K}$}}, respectively (see Fig.\ref{fig2b}), as introduced in the following.
\begin{itemize}
  \item The augmented data approach attempts to compensate the input data with other sources of data highly related to the input small samples, usually yielded through transformation, synthesis, imagination or other ways, so as to make LSL applicable \citep{kulkarni2015deep,long2017zero,antoniou2017data,chen2018semantic}

  \item The knowledge system approach could use following types of knowledge:
   \begin{itemize}
     \item Representations from other domains: Transferring the knowledge of describing the same target while from different domains, e.g., the lack of visual instances can be compensated by semantic descriptions on the same object \citep{srivastava2012multimodal,ramachandram2017deep} (see Fig. \ref{fig3a}).
     \item Trained models: A model trained from other related datasets can be used to fit the small training samples in the given dataset through fine-tuning its parameters. The model can be a neural network, random forest, and others \citep{yosinski2014transferable,hinton2015distilling} (see Fig. \ref{fig3b}).
     \item Cognition knowledge on concepts: Such knowledge include  common sense knowledge, domain knowledge, and other prior knowledge on the learned concept with small training samples \citep{davis2015commonsense,doersch2015unsupervised,stewart2017label}. For example, if we want localize the eyes of Lenna (see Fig. \ref{fig3c}), we can employ the cognition knowledge that the positions of eyes are above that of the mouth.
     \item Meta knowledge: Some high-level knowledge beyond data, and can be helpful to compensate each concept learning with not sufficient samples \citep{lemke2015metalearning,Lake2016}.
   \end{itemize}

\end{itemize}

In some literatures, the setting of experience learning is also called small data learning \citep{vinyals2016matching,santoro2016meta,hariharan2017low,edwards2016towards,altae2017low,munkhdalai2017meta}.
Here we call this learning manner experience learning yet for unified descriptions.

\begin{figure}
  \centering
  \includegraphics[width=4.5in]{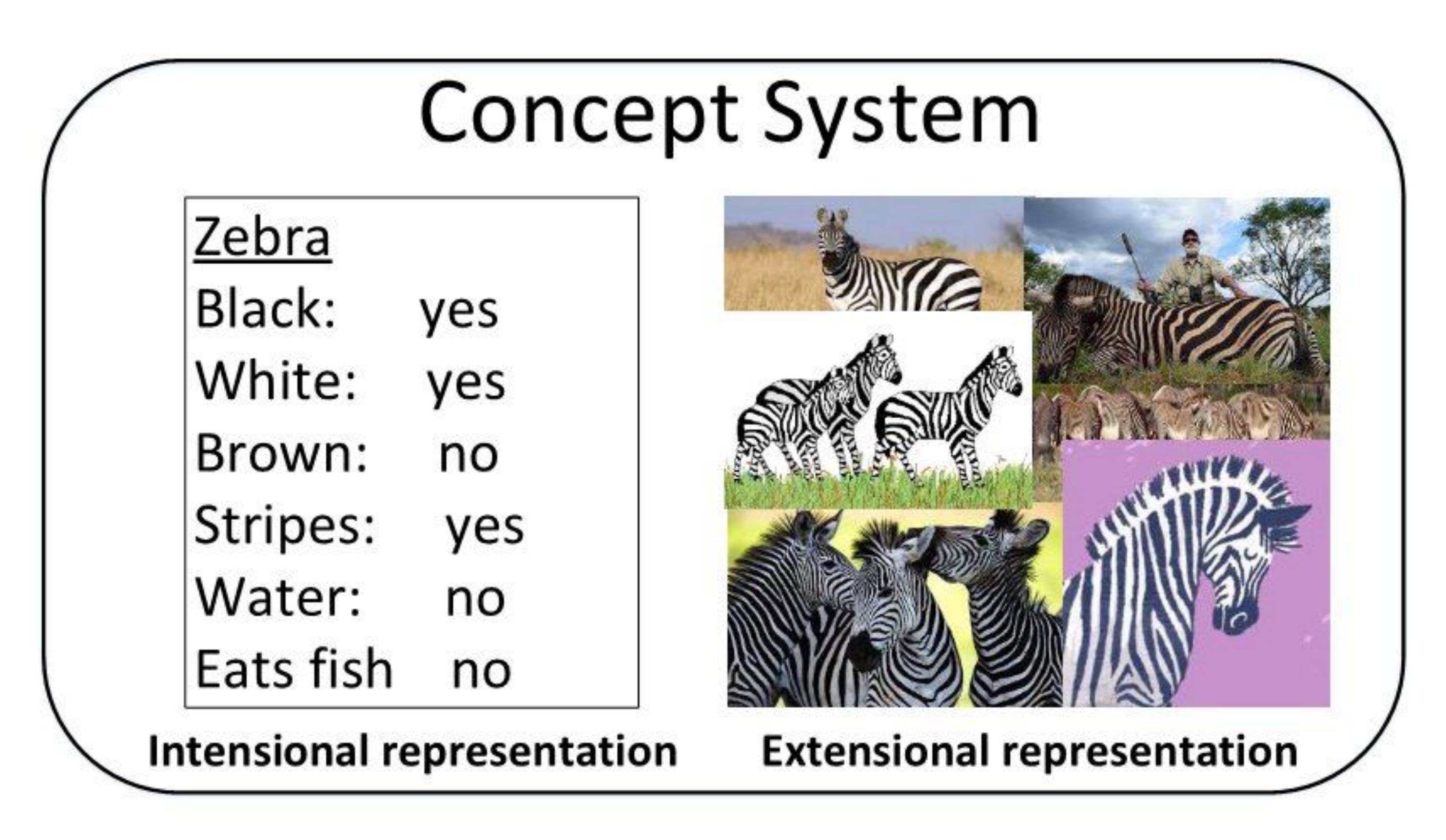}
  \caption{\small Related illustration of the concept system.}\label{f4}
\end{figure}

Compared with experience learning, concept learning represents more primary branch of current SSL research. This learning paradigm aims to
perform recognition or form new concepts (samples) from few observations (samples) through fast processing. Conceptually, concept learning employs \textbf{\emph{matching rule $\mathcal{R}$}} to associate concepts in \textbf{\emph{concept system $\mathcal{C}$}} with input \textbf{\emph{small samples $\mathcal{S}$}}, whose function is mainly performing cognition or completing a recognition task as well as generation, imagination, synthesis and analysis~(see Fig.\ref{fig2c}). The aforementioned concepts are explained as follows:
\begin{itemize}
  \item \textbf{Concept system} includes intensional representations and extension representations of concepts (see Fig.\ref{f4}):
  \begin{itemize}
    \item Intensional representation indicates precise definitions in proposition or semantic form on the learned concept, like its attribute characteristics.
    \item Extensional representation denotes prototypes and instances related to the learned concept.
  \end{itemize}
  \item \textbf{Matching rule} denotes a procedure to associate concepts in concept system $\mathcal{C}$ with small samples $\mathcal{S}$ to implement a cognition or recognition task. The result tries to keep optimal in terms of performance measure $\mathcal{P}$.
\end{itemize}

\subsubsection{$k$-shot Learning}
$k$-shot learning aims to learn information about object categories from $k$ training images, where $k$ is generally a very small number like $0$ or $1$. The mathematical expression to this learning issue can be described as follows: Given dataset $\mathcal{D}_1 = \{\mathcal{X_S,Y_S}\}=\{(x_i^s,y_i^s)\}_{i=1}^{N^s}, \mathcal{D}_2 = \{\mathcal{X_U,Y_U}\}=\{(x_i^u,y_i^u)\}_{i=1}^{N^u}$, each datum $x_i^s$ or $x_i^u\in\mathbb{R}^{d\times 1}$ is a $d$-dimensional feature vector, $y_i^s$ or $y_i^u$ denotes the corresponding label, and usually suppose $\mathcal{Y_S}\cap \mathcal{Y_U}$ $=$ $\emptyset$.
\textbf{\emph{Zero-shot learning}} (ZSL) aims to recognize unseen objects $\mathcal{X_U}$ in $\mathcal{D}_2$ by leveraging the knowledge $\mathcal{D}_1$. That is, ZSL aims to construct a classifier $f:\mathcal{X_U}\rightarrow \mathcal{Y_U}$ by using the knowledge $\mathcal{D}_1$. Particularly, if the training and test classes are not disjoint, i.e., $\mathcal{Y_S}\cap \mathcal{Y_U} \neq \emptyset$, the problem is known as generalized zero-shot learning (GZSL)~\citep{chao2016empirical,xian2017zero,song2018transductive}.
Comparatively, \emph{\textbf{$\mathbf{k}$-shot learning}} aims to construct a classifier $f:\mathcal{X_U}\rightarrow \mathcal{Y_U}$ by means of information in $\mathcal{D}_1\cup \mathcal{D}$, where $\mathcal{D} \subseteq \mathcal{D}_2$, consisting of the unseen objects in which each category contains $k$ known-label objects. Especially, when $k=1$, we call it \textbf{\emph{one-shot learning}}.

Note that experience learning and concept learning are two categories of learning approaches for SSL, while $k$-shot learning just describes a setting manner of the SSL problem, and can be set under both learning manners.
Among current researches, $k$-shot learning is mainly presented in the recognition problem~\citep{fu2018recent}. In the future, more other problems are worthy to be investigated, such as generation~\citep{Lake2015}, synthesis~\citep{long2017zero}, parsing~\citep{zhu2018zero,bansal2018zero,rahman2018zero,shaban2017one}, account for concepts are far more complicated than object categories only.

\subsection{Neuroscience Evidences for SSL}
The motivation of SSL is to mimic the learning capability of humans, who can learn new concepts from small sample with strong generalization ability. Here, we try to list more neuroscience evidences to further support the possible feasibility of the SSL paradigm~\citep{Hassabis2017}.
\begin{figure}
  \centering
  \subfigure[Episodic memory]{
  \label{fig5a}
  \includegraphics[width=2.5in]{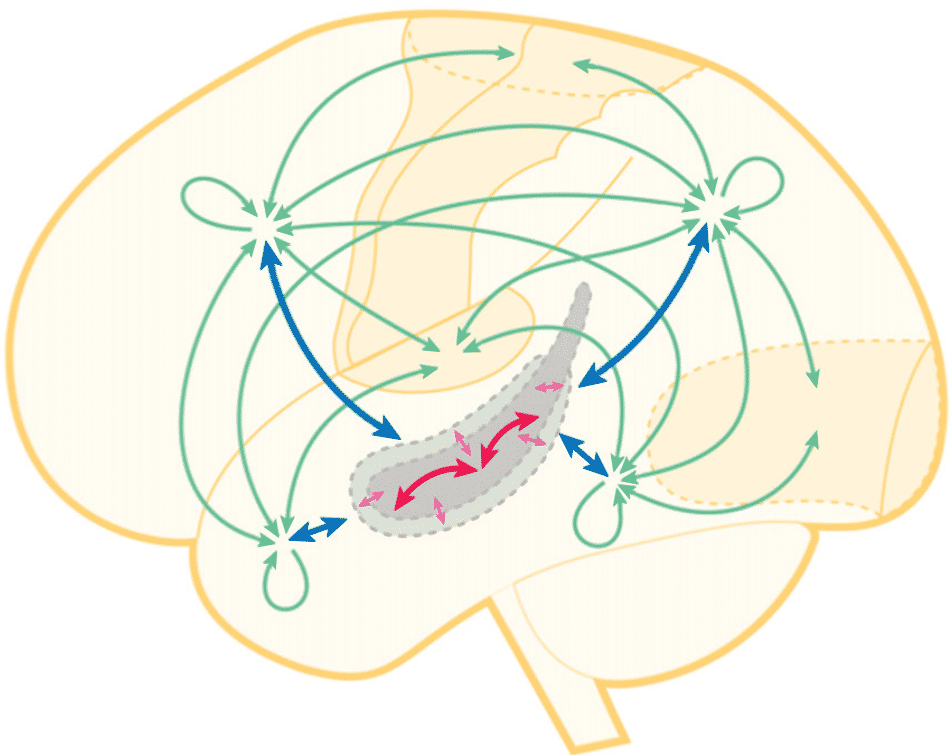}}
  \hspace{0.5in}
  \subfigure[Two-pathway guided search]{
  \label{fig5b}
   \includegraphics[width=2.5in]{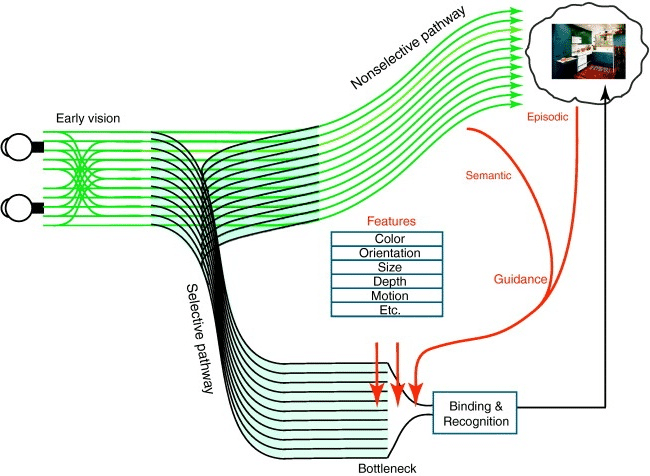}}
   \subfigure[Imagination and planning]{
  \label{fig5c}
   \includegraphics[width=2.5in]{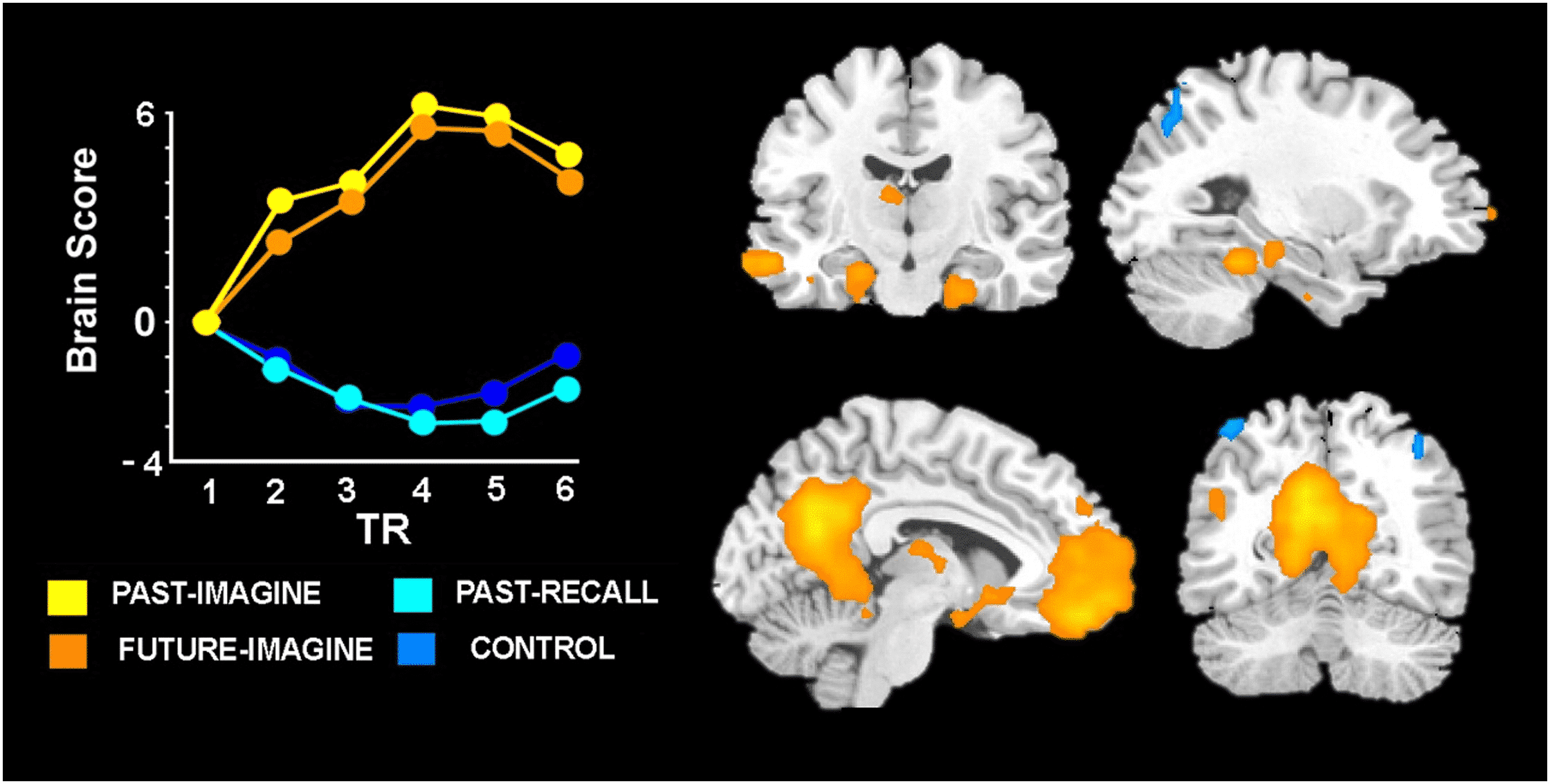}}
    \hspace{0.5in}
    \subfigure[Compositionality and causality]{
  \label{fig5d}
   \includegraphics[width=2.5in]{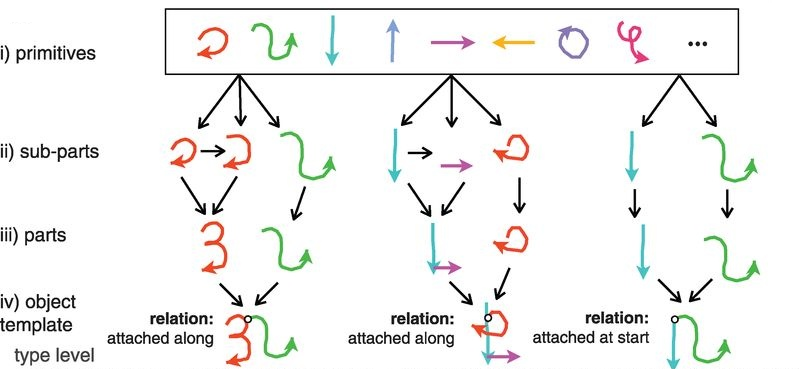}}
  \caption{\small Illustrations for neuroscience evidences for SSL. (a) Episodic Memory (image is reproduced from \citep{kumaran2016learning}). Medial temporal lobe (MTL) surrounded by broken lines, with hippocampus in dark grey and surrounding MTL cortices in light grey. In Complementary Learning Systems (CLS), connections within and among neocortical areas (green) support gradual acquisition of structured knowledge through interleaved learning while rapid learning in connections within hippocampus (red) supports inial learning of arbitrary new information. Bidirectional connections (blue) link neocortical representations to the hippocampus/MTL for storage, retrieval, and replay. (b) Two-pathway guided search (image is reproduced from \citep{wolfe2011visual}). A selective pathway can bind features and recognize objects, but it is severely capacity-limited. While a non-selective pathway can extract statistics from the entire scene, it enables a certain amount of semantic guidance for the selective pathway. (c) Imagination and planning (image is reproduced from \citep{schacter2012future}). In memory system, there is a remarkable resemblance between remembering the past and imagining or simulating the future. Thus a key function of memory is to provide a basis for predicting  the future via imagined scenarios and that the ability to flexibly  recombine elements of past experience into simulations of novel  future events is an adaptive process. (d) Compositionality and causality (image is reproduced from \citep{Lake2015}). To learn a large class of visual concepts, compositionality helps build rich concepts from simpler primitives, and causal structure of the real-world processes handles noise and support creative generalizations.}\label{f5}
\end{figure}

\subsubsection{Episodic Memory}\label{section221}
 Human intelligence is related to multiple memory systems~\citep{tulving1985many}, including procedural, semantic, episodic memory~\citep{tulving2002episodic} and so on. In particular, experience replay~\citep{Mnih2015,schaul2015prioritized}, a theory that can describe how the multiple memory system in the mammalian brain might interact, is critical to maximize data efficiency. In complementary learning systems (CLS) theory~\citep{kumaran2016learning}, mammalians possess two learning systems, including parametric slow-learning neocortical system and non-parametric fast learning hippocampal system (see Fig.\ref{fig5a}). The hippocampus encodes novel information after a single exposure, while this information is gradually consolidated to the neocortex in sleep or resting periods that are interleaved with periods of activity~\citep{Hassabis2017}. In O'Neill's view~\citep{o2010play}, the consolidation is accompanied by replaying in the hippocampus and neocortex, which is observed as a reinstatement of the structured patterns of neural activity that accompanied the learning event. Therefore, experiences stored in a memory buffer can be used to gradually adjust the parameters of learning machine, such as SSL, which supports rapid learn based on an individual experience. This learning procedure is guaranteed by episodic control~\citep{blundell2016model}, that rewarded action sequences can be internally re-enacted from a rapidly updateable memory store~\citep{gershman2017reinforcement}. Recently, episodic-like memory systems have shown considerable promise in allowing new concepts to be learned rapidly based on only a few examples~\citep{vinyals2016matching,santoro2016meta}.

\subsubsection{Two-pathway guided search}\label{section222}
One notable feature of SSL is fast leaning. For example, visual search is necessary for rapid scene analysis
in daily life because information processing in the visual system is limited to one or a few targets or regions at one time. There exists a two-pathway guided search theory~\citep{wolfe2011visual} to support human fast visual search. As show in Fig.\ref{fig5b}, observers extract spatial layout information rapidly from the entire scene via the non-selective pathway. This global information of scene acts as top-down modulation to guide the salient object search in the selective pathway. This two-pathway based
search strategy provides parallel processing of global and local information for rapid visual search.

\subsubsection{Imagination and Planning}\label{section224}
 Humans are experts in simulation-based planning. Specifically, humans can more flexibly select actions based on predictions of long-term outputs, that are generated  from an internal model of the environment learned through experiences \citep{dolan2013goals,pezzulo2014internally}. Human is able to not only remember the past experience, but also image or simulate the future~\citep{schacter2012future}, which includes memory-based simulations and goal-directed simulations. Although imaging is intrinsically subjective and unobservable, we have reasons to believe that it has a conserved role in simulation-based planning across species~\citep{schacter2012future,hassabis2009construction} (see Fig.\ref{fig5c}). In AI system, Some progress has been made in scene understanding~\citep{eslami2016attend}, 3D structure learning~\citep{rezende2016unsupervised}, one-shot generalization of characters~\citep{rezende2016one}, zero-shot recognition~\citep{long2017zero}, and imagination of realistic environment~\citep{chiappa2017recurrent,racaniere2017imagination,gemici2017generative,hamrick2017metacontrol}  based on simulation-based planning and imagination, which leads to data efficiency improvement and novel concept learning.

It should be indicated that there exists two key ideas, compositionality and causality~\citep{Lake2016}, in imagination and planning procedure. Rich concepts can be built compositionally from simpler primitives~\citep{Lake2015}. Compositionality allows for reuse of a finite set of primitives across many various scenarios by recombining them to produce an exponentially large number of novel yet coherent and useful concepts~\citep{Lake2016}. Recent progress has been made by SCAN~\citep{higgins2017scan} in implicit hierarchy of abstract concepts from as few as five symbol-image pairs per concept. The capacity to extract causal knowledge~\citep{sloman2005causal} from the environment allows us to image and plan future events and to use those imagination and planning to decide on a course of simulations and explanation~\citep{tervo2016toward,bramley2017constructing}. Typically, \citet{Lake2015} exploits causal relationship of combining primitive to reduce the dimension of hypothesis, and therefore leads to successes in classifying and generating new examples after seeing just a single example of a new concept (see Fig.\ref{fig5d}).

\begin{figure}
  \centering
  \includegraphics[width=4.5in]{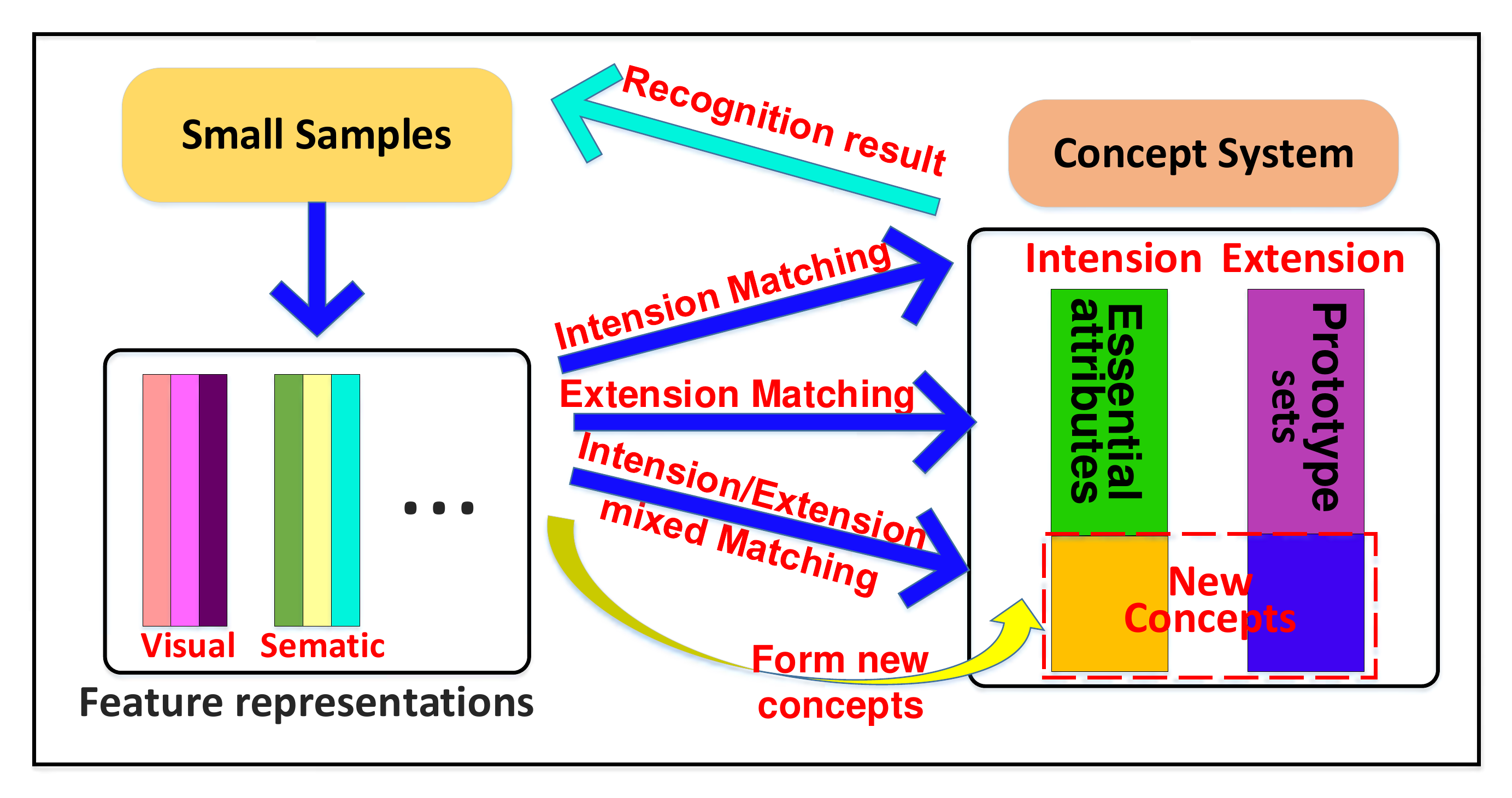}\\
  \caption{\small Illustration of the concept learning framework.}\label{f6}
\end{figure}

\section{Techniques on Concept learning}\label{section3}
In this section, we will give an overview on the current developments on concept learning techniques. We first present a general methodology for concept learning, including intension matching, extension matching, and intension/extension mixed matching. Then we review relevant methods proposed from these three aspects, respectively.

\subsection{A General Methodology for Concept Learning}
Concept learning aims to perform recognition or form new concepts (classes) from few observations (samples) through fast processing. As shown in Fig.\ref{f6}, a concept system generally includes intensional representation and extensional representation of concepts. When small samples come, it is assumed that there exist various domain feature representations, like visual representations and semantic representations. Then concept learning employs matching rule to associate concepts in concept system with feature representations of small samples.

We summarize this procedure in Algorithm \ref{algorithm1}~(steps 1-3 can be executed in random ordering). Particularly, different domain representations can be integrated to match the intensional representation of concepts (intension matching), and virtual instances can be generated as extensional representations to match small samples (extension matching). Moreover, feature representations and concept representations can be aligned at a middle-level feature space, sometimes an embedding space. Through these matching rules, the recognition task can be implemented to return the final result. Under the unexpected situation that no concepts match the small samples, new concept could be formed via synthesis and analysis to undate the concept system \citep{scheirer2013toward,shmelkov2017incremental}.

\begin{algorithm}[tb]
   \caption{A General Methodology for Concept Learning}
   \label{algorithm1}
\begin{algorithmic}
   \STATE {\bfseries Input:} Small sample inputs (might be featured in different description domains, like semantic and pixel-based.).
   \STATE {\bfseries Execute:}
   \STATE \ \ \ \ \ \ \ 1) \textbf{Intension Matching} (via integrating different domain representations);
   \STATE \ \ \ \ \ \ \ 2) \textbf{Extension Matching} (via generating virtual instances to judge);
   \STATE \ \ \ \ \ \ \ 3) \textbf{Intension/Extension Mixed Matching} (via aligning features and concepts at a middle-level feature space);
   \STATE \ \ \ \ \ \ \ 4) \textbf{New Concept Formation} (via synthesis and analysis to form new concept).
   \STATE {\bfseries Output:} Realize recognition task or form a new concept.
\end{algorithmic}
\end{algorithm}
\subsection{Intension Matching} \label{IM}
To match the intensional representation of concepts and different domain representations, it generally needs to learn a mapping $f$ between feature representations $\varphi$ and intensional representations $\Phi$, which can be bidirectional, i.e., from $\varphi$ to $\Phi$ or from $\Phi$ to $\varphi$, and then output the results that maximize the similarity between two representations.

\subsubsection{From visual feature space to semantic space}\label{vtos}
The first category of intension matching approaches learns a mapping function by regression from the visual feature space to the semantic space ($\mathbf{V\rightarrow S}$), which includes attributes~\citep{farhadi2009describing,parikh2011relative}, word vectors~\citep{frome2013devise,socher2013zero}, text descriptions~\citep{elhoseiny2013write,reed2016learning} and so on. In this case, there mainly exist two categories:

\begin{itemize}
  \item \textbf{Semantic Embedding}. This approach directly learns a mapping function between the visual feature space and the semantic embedding space. The function is usually learned from the labelled training visual data consisting of seen classes only. After that, zero-shot classification is performed directly by measuring similarity using nearest neighbour (NN) or its probabilistic variants such as direct attribute prediction~\citep{lampert2009learning,lampert2014attribute}.
  \item \textbf{Semantic Relatedness}. This approach learns an $n$-way discrete classifier for the seen classes in the visual feature space, which is then used to compute the visual similarity between an image of unseen class to those of the seen classes. Specifically, the semantic relationship between the seen and unseen classes is modelled by the distance between their prototypes to combine knowledge of seen classs, or the knowledge graph to distill the relationships~\citep{rohrbach2010helps,fergus2010semantic,deng2014large,marino2017more}.
\end{itemize}

The pioneer work of semantic embedding was conducted by \cite{lampert2009learning,lampert2014attribute}, which used a Bayesian model to build the relationship between visual feature space and the semantic space. They provided two models for zero-shot learning, i.e., direct attribute prediction (DAP) and indirect attribute
prediction (IAP), with the idea that learned the probability of attributes for given visual instance as prior and computes a MAP prediction of the unseen classes. Variants like topic model~\citep{yu2010attribute}, random forests~\citep{jayaraman2014zero}, and Bayesian networks~\citep{wang2013unified} have been explored. Whilst, \cite{palatucci2009zero} presented a semantic output codes classifier, which directly learned a function (e.g.,ar regression) from visual feature space to the semantic feature space by one-to-one attribute encoding according to its knowledge base. Except for linear embedding, benefited from deep learning~\citep{goodfellow2016deep}, some nonlinear embeddings have also been developed. For example, \cite{socher2013zero} learned a deep learning model to map image close to semantic word vectors corresponding to their classes, and this embedding manner could be used to distinguish whether an image is of a seen or unseen class. Also, \cite{frome2013devise} presented a deep visual-semantic embedding model (DeViSE) trained to bridge visual objects using both labeled image data as well as semantic information gleaned from unannotated texts. Besides, they tried to alleviate the limitation of ability that scales to large number of object categories by introducing unannotated text beyond annotated attributes, achieving good performance on the 1000-class ImageNet object recognition task for the first time. Afterwards, \cite{Zhang_2018_CVPR} firstly investigated Polynomial and the RBF family of kernels to obtain a non-linear embedding.
Recently, some works were investigated by additional constraints to learn the mapping. For example,\cite{deutsch2017zero} cast the problem of ZSL as fitting a smooth function defined on a smooth manifold (visual domain) to sample data.
To enhance the capability of discrimination, \cite{morgado2017semantically} introduced two forms of semantic constraints to the CNN architecture. The model was encouraged to learn an hidden semantic layer together with a semantic code for classification.  \citep{yu2018stacked} further applied attention mechanism to generating an attention map for weighting the importance of different local regions and then integrated both the local and global features to obtain more discriminative representations for fine-grained ZSL.
Recently, \cite{chen2018zero} tried to introduce adversarial learning enables semantics transfer across classes to improve classification.
Another attempt presenting an unsupervised-data adaptation inference framework into few/zero-shot learning was made by \citep{tsai2017learning} to learn robust visual-semantic embeddings. Specifically, they combined auto-encoders representation learning models together with cross-domain learning criteria (i.e., Maximum Mean Discrepancy loss) to learn joint embeddings for semantic and visual features in an end-to-end learning framework.

Based on semantic relatedness, \cite{norouzi2013zero}  mapped images into the semantic embedding space via convex combination of the seen class label embedding vectors, i.e., using the probabilities of a softmax output layer to weight the semantic vectors of all the classes, that can perform zero-shot learning task on the large-scale ImageNet dataset. Likewise, \cite{mensink2014costa} used co-occurrence statistics learning concept-concept relationships from texts (between the new label and existing ones), as a weight to combine seen classes classifiers. While \cite{changpinyo2016synthesized} directly applied the convex combination scheme to synthesizing classifiers for the unseen classes. Similar work like \citep{misra2017red} was also proposed, and  applied a simple composition rule to generating classifiers for new
complex concepts. On the other hand, \cite{salakhutdinov2011learning} used knowledge graph like WordNet early to build a hierarchical classification model that allowed rare objects to borrow statistical strength from related objects that may have many training instances. After that, \cite{deng2014large} introduced hierarchy and exclusion (HEX) graphs to train object classifiers by leveraging mutual exclusion among different classes.
In order to define a proper similarity distance metric between a test image and the unseen class prototypes for ZSL, \cite{fu2015zero} further explored rich intrinsic semantic manifold structure using a semantic graph in which each class is a
node and the connectivity on the graph is determined by the semantic relatedness between classes.
Recently, semantic relatedness provides supervision information for transferring knowledge from seen classes to unseen classes. For example, \cite{guo2017zero} proposed a sample transfer strategy that transferred samples based on their transferability and diversity
from seen classes to unseen classes via the class similarity, and assigned pseudo labels for them to train classifiers.
Furthermore, \cite{li2017zero} exploited the intrinsic relationship between the semantic space manifold and the transfer ability of visual-semantic mapping to generate more consistent semantic space with the image feature space.
The graph convolutional network (GCN) technique was introduced  into \citep{wang2018zero} to transfer information (message-passing) between
different categories. They tried to distill information via both semantic embeddings and knowledge graphs, in which a knowledge graph provided supervision to learn meaningful classifiers on top of semantic embeddings.  Another work employed information propagation mechanism to reason the unseen labels is proposed by \citep{lee2017multi}. They designed multi-label ZSL by incorporating knowledge graphs for describing the relationships between multiple labels.

Since visual domain and semantic domain have different tasks and non-overlapping label spaces, the aforementioned existing methods are prone to the projection domain shift problem~\citep{fu2014transductive}. To alleviate this issue, some works focus on manifold assumption. \cite{fu2015transductive} firstly proposed a method to preserve the coherent of manifold structures of different representation spaces. And then \cite{li2017zerostructure} incorporated a graph Laplacian regularization to preserve the geometric properties of target data in the label space as visual feature space, and the similar work focusing on preserving the locally visual structure was conducted by \cite{ji2017manifold}. Moreover, \cite{xu2017matrix} used matrix tri-factorization framework with manifold regularization on visual feature and semantic embedding spaces.
As well, \cite{xu2017transductive} investigated manifold regularization regression for zero-shot action recognition.  However, some works investigated the joint structure learning between seen and unseen class. For example,
\cite{kodirov2015unsupervised} cast the mapping function learning problem as a sparse coding problem, joint learning seen and unseen semantic embedding. Specifically, each dimension of the semantic embedding space corresponds to a dictionary basis vector and the coefficients/sparse code of each visual feature vector is its projection in the semantic embedding space, enforcing visual projection in the semantic embedding space to be near to the unseen class prototypes. \cite{zhang2016zero} further proposed a joint structured prediction scheme to seek a globally well-matched assignment structure between visual clusters and unseen classes in test time.
Another attempts borrow the idea from self-paced learning \citep{kumar2010self,jiang2014easy,jiang2014self} and  were made by \cite{yu2017transductiveadaptive} and \cite{niu2017zero}. In a nutshell, their method iteratively selected the unseen instances from reliable to less reliable to gradually refine the predicted test labels and update the visual classifiers for unseen categories alternatively.
Along this, similar works were developed by \citep{ye2018self} and \citep{luo2018zero}.
Comparatively, in each iteration, \cite{ye2018self} selected the most confidently predicted unlabeled instances to refine the ensemble network parameters, and \cite{luo2018zero} refined the class prototypes instead of labels. Different with other methods, \cite{kodirov2017semantic} took the encoder-decoder paradigm, and they insist that it is very effective in mitigating the domain shift problem with additional reconstruction constraint. Likewise, \cite{fu2018zero} extended \citep{fu2015zero} by introducing a ranking loss. Specifically, the ranking loss objective was regularised by unseen class prototypes to prevent the projected object features from being biased towards the seen prototypes.

Another important factor degrading the performance of recognition is that the textual representation is usually too noisy.
Against this issue, \cite{qiao2016less} proposed an $l_{2,1}$-norm based objective function which can simultaneously suppressed the noisy signal in the text and learned a function to match the text document and visual features. Afterwards, \cite{al2017automatic} used a linguistic prior in a joint deep model to optimize the class-attribute associations to address noise and missing data in the text corpora. Besides, \cite{elhoseiny2017link} proposed a learning framework that was able to connect text terms to its relevant parts of objects and suppress connections to non-visual text terms without any part-text annotations.
More recently, \cite{Zhu_2018_CVPR} simply passed textual features through additional fully connected layer before feeding it into the generator, and they argued that the  modification achieved the comparable performance of noise suppression.

\subsubsection{From semantic space to visual feature space} \label{stov}

The second category of approaches along this research line learns a mapping function from the semantic space to the visual feature space ($\mathbf{S\rightarrow V}$).
The motivation to learn the mapping is to solve the hubness problem for the first time, i.e., the neighbourhoods surrounding mapped vectors contain many items that are ``universal'' neighbours. \cite{radovanovic2010hubs} and \cite{dinu2014improving} firstly noticed this problem in zero-shot learning. \cite{shigeto2015ridge} argued that least square regularised projection functions make the hubness problem worse and firstly proposed to perform reverse regression, i.e., embedding class prototypes into the visual feature space.
Transductive setting assumption was used in \citep{shojaee2016semi}, and they used both labeled samples of seen classes and unlabeled instances of unseen classes to learn a proper representation of labels in the space of deep visual features in which samples of each class are usually condensed in a cluster. After that,
\cite{changpinyo2017predicting} learned a mapping function such that the semantic representation of class can predict well its class exemplar (center) that characterized the clustering structure, and then the function was used to construct nearest-neighbor style classifiers.
Different from the aforementioned, \cite{zhang2017learning} learned an end-to-end deep learning model that maps semantic space to visual feature space, and dealt with the hubness problem efficiently. Recently, \cite{annadani2018preserving} learned an a multilayer perceptron based encoder-decoder, and preserved the structure of the semantic space in the embedding space (visual feature space) by utilizing semantic relations between categories while ensured discriminative capability.

\subsection{Extension Matching} \label{EM}
This category of approaches are constructed through using the input feature or generating a series of virtual instances according to the feature to make it possible to compare with the instances in the extension representation, and help find the one that maximizes the similarities between extensional representation and the feature/virtual instances. This is motivated by the fact that human can associate familiar visual elements and then imagine an approximate scene given a conceptual description. Note that extension matching is different from learning a mapping function from the semantic space $\mathbf{S}$ to the visual feature space $\mathbf{V}$. Intuitively, the latter can be regarded as learning how to recognize the characteristics of an image and match it to a class. On the contrary, extension matching can be described as learning what a class visually looks like. Sometimes, extension matching has two explicit advantages over the $\mathbf{S\rightarrow V}$ learning manner as introduced in the previous section:
\begin{itemize}
  \item $\mathbf{S\rightarrow V}$ framework inclines to bring in information loss to the system, so as to degrade the overall performance. Comparatively, extension matching recognizes a new instance in the original space, which can help alleviate this problem.
  \item  Through synthesizing a series of virtual instances, we can always turn the SSL problem into a conventional supervised learning problem (LSL problem) such that we can take advantage of the power of LSL techniques in the SSL task, or directly use nearest neighbour (NN) algorithm.
\end{itemize}

The early work of extension matching was conducted by \citep{yu2010attribute}, through synthesizing data for ZSL using the Author-Topic (AT) model. Yet the drawback of the method is that it only deals with discrete attributes and discrete visual features like bag-of-visual-word feature accounting for the attributes, and yet the visual features usually have continuous values in real world. There are more methods proposed in recent years, which can be roughly categorized into the following three categories:

1) Learning an embedding function from $\mathbf{S}$ to $\mathbf{V}$.
\cite{long2017zerocvpr,long2017zero} provided a framework to synthesize unseen visual (prototype) features by given semantic attributes. As aforementioned, $\mathbf{S\rightarrow V}$ framework may lead to inferior performance owing to three main problems, in terms of structural difference, training bias, and variance decay, respectively. In correspondence, a latent structure-preserving space via dual-graph approach with the diffusion regularisation is proposed in their work.

2) Learning a probabilistic distribution for each seen class and extrapolating to unseen class distributions using the class-attribute information.
Assume that data of each class in the image feature space approximately followed a Gaussian distribution, \cite{guo2017synthesizing} synthesized samples by random sampling with the distribution for each target class. Technically, the conditional probabilistic distribution for each target class was estimated by linear reconstruction based on the structure of the class attributes. While \cite{zhao2017zero} posed ZSL as the missing data problem, estimating data distribution of unseen classes in the image feature space by transferring the manifold structure in the label embedding space to the image feature space. More generally, \cite{verma2017simple} modeled each class-conditional distribution as an exponential family distribution and the parameters of the distribution of each seen/unseen class are defined as functions of the respective seen class attributes. Besides, these functions can be learned using only the seen class data and can be used to predict the parameters of the class-conditional distribution of each unseen class. Another attempt was to develop a joint attribute feature extractor in \citep{lu2017zero}.  In their method, each fundamental unit was put in charge of the extraction of one attribute feature vector, and then based on the attribute descriptions of unseen classes, a probability-based sampling strategy was exploited to select some attribute feature vectors to synthesize combined feature representations for unseen class.

3) Using the generative model like generative adversarial network (GAN) \citep{goodfellow2014generative} or variational autoencoder (VAE)~\citep{kingma2013auto} to model the unseen classes' distributions with the semantic descriptions and the visual distribution of the seen classes. Especially, using generated examples of unseen classes and given examples of seen classes to train a classification model provides an easy manner to handel the GZSL problem. For example, \cite{bucher2017generating} learned a conditional generator (e.g., conditional GAN~\citep{odena2017conditional}, denoising auto-encoder~\citep{bengio2013generalized}, and so on) for generating artificial training examples to address ZSL and GZSL problems. Furthermore, \cite{xian2017feature} proposed a conditional Wasserstein GAN~\citep{gulrajani2017improved} with a classification loss, f-CLSWGAN, generating sufficiently discriminative CNN features from different sources of class embeddings.
Similarly, by leveraging GANs, \cite{zhang2018visual} realized zero-shot video classification.
Other works focus on VAE~\citep{kingma2013auto}. For example, \cite{wang2018zero} represented each seen/unseen class using a class-specific latent-space distribution, and used VAE to learn highly discriminative feature representations for the inputs. At test time, the label for an unseen-class test input is the class that maximizes the VAE lower bound. Afterwards, \cite{mishra2017generative}  trained a conditional VAE~\citep{sohn2015learning} to learn the underlying probability distribution of the image features conditioned on the class embedding vector. Similarly, \cite{arora2018generalized} proposed a method able to generate semantically rich CNN feature distributions via the conditional VAE with discriminator-driven feedback mechanism improving the reconstruction capability.

\subsection{Intension/Extension Mixed Matching}
This category of method aims to map both feature and concept representations into a middle-level representation space, and then predict the class label of an unseen instance by ranking the similarity scores between semantic features of all unseen classes and the visual feature of the instance in the middle-level representation space. The middle-level representation may be the result after a  mathematical transformation (say, Fourier transformation). This strategy is different from semantic relatedness strategies as introduced in Section \ref{vtos}, which accounts for semantic relatedness provides supervision information for combining or transferring knowledge from seen classes to unseen classes. Comparatively, intension/extension mixed matching implicitly/explicitly learns a middle-level representation space, in which the similarity between visual and semantic space can be easily determined. There are mainly three categories of typical approaches for learning the middle-level representation space, as summarized in the following.

1) Learning an implicit middle-level representation space through learning consistency functions like compatibility functions, canonical correlation analysis (CCA), or other strategies.
E.g., \cite{akata2013label,akata2016label} proposed a model that implicitly learned the instances and the attributes embeddings onto a common space where the compatibility between any pair of them can be measured. When given an unseen image, the correct class can be obtained through the rank higher than the incorrect ones.  This consistency function has the form of a bilinear relation $\mathbf{W}$ associating the image embedding $\theta(x)$  and the label representation $\Phi(y)$ as $S(x,y;\mathbf{W})=\theta(x)^T\mathbf{W}\Phi(y)$.
To make the whole process simpler and efficient, \cite{romera2015embarrassingly} proposed a different loss function and regularizer based on the same principle as \citep{akata2013label,akata2016label} with a closed form solution to $\mathbf{W}$. Moreover, \cite{akata2015evaluation} learned a joint embedding semantic space between attributes, text, and
hierarchical relationships while \cite{akata2013label} considered attributes as output embeddings.
To learn more powerful mapping, some works focus on nonlinear styles.
\cite{xian2016latent} learned a nonlinear (piecewise linear) compatibility function, incorporating multiple linear compatibility
units and allowed each image to choose one of them and achieve factorization over such (possibly complex combinations of) variations in pose, appearance and other factors. Readers can refer to \citep{xian2017zero}, in which both the evaluation protocols and data splits are evaluated among the linear compatibility functions and nonlinear compatibility functions.
Another attempt was tried to learn visual features rather than fixed visual features \citep{akata2013label,akata2016label,akata2015evaluation,romera2015embarrassingly}.
Inspired by \citep{elhoseiny2013write,elhoseiny2017write} to learn pseudo-concepts to associate novel classes using Wikipedia
articles, \citep{ba2015predicting} used text features to predict the output weights of both the convolutional and the fully connected layers in a deep CNN as visual features.
Also, existing methods may rely on provided fixed label embeddings. However, a representative work is like \cite{jiang2017learning}, which learned label embeddings with or without side information (encode prior label representation) and integrated label embedding learning with classifier training. This is expected to produce adaptive label embeddings that are more informative for the target classification task.
To overcome a large performance gap in zero-shot classification between attributes and unsupervised word embeddings, \cite{reed2016learning} extended \citep{akata2015evaluation}'s work to train an end-to-end deep neural language models from texts, and used the inner product of features generated by deep neural encoders instead of bilinear compatibility function, achieving a competitive recognition accuracy compared to attributes.

On the other hand, the early work of CCA is proposed by \citep{hardoon2004canonical}, using kernel CCA to learn a semantic representation to web images and their associated text. Recently, \cite{gong2014multi} investigated a three-view CCA framework that incorporates the dependence of visual features and text on the underlying image semantics for retrieval tasks. \cite{fu2015transductive} further proposed  transductive multi-view CCA to learn a common latent embedding space aligning different semantic views and the low-level feature view, which alleviated the bias/projection domain shift. Afterwards, \cite{cao2017generalized} proposed a unified multi-view subspace learning method for CCA using the graph embedding framework for visual recognition and cross-modal retrieval. Also, there exist other CCA variants for the task, like \citep{qi2017joint,mukherjee2017deep}. For example, \cite{qi2017joint} proposed an embedding model jointly transferring inter-model and intra-model labels for an effective image classification model. The inter-modal label transfer is
generalized to zero-shot recognition. \cite{mukherjee2017deep} introduced deep matching autoencoders~(DMAE) which learned a common latent space and pairing from unpaired multi-modal data. Specifically, DMAE is a general cross-modal learner that can be learned in an entirely unsupervised way, and ZSL is the special case.

2) Learning an implicit middle-level representation space through dictionary learning.
\cite{zhang2016zero} firstly learned an intermediate latent embedding based on dictionary learning to jointly learn the parameters of model for both domains that can not only accurately represent the observed data in each domain but also infer cross-domain statistical relationships when one exists. Similar works were also proposed, like \citep{peng2016joint,ding2017low,jiang2017learning,ye2017zero,yu2017transductive,kolouri2017joint}.
For example, to mitigate the distribution divergence across seen and unseen classes, \cite{ding2017low} learned a semantic dictionary to link visual features with their semantic representations based on a low-rank embedding space assumption, in which the latent semantic dictionary for unseen data should share its majority with semantic dictionary for the seen data.
\cite{jiang2017learning} further proposed a method to learn a latent attribute space with a dictionary learning framework to tackle the problems of attribute-based approaches simultaneously, i.e., discriminative~\citep{yu2013designing}, interdependent~\citep{jayaraman2014decorrelating}, large variations~\citep{kodirov2015unsupervised} within each attribute.
Analogously, \cite{yu2017transductive} formulated a dictionary framework to learn a  bidirectional mapping based semantic relationship modeling scheme that sought for cross-modal knowledge transfer by simultaneously projecting the image features and label embeddings into a common latent space.
Latest work in \citep{kolouri2017joint} modeled the relationship between visual features and semantic attributes via joint sparse dictionaries, demonstrating an entropy regularization scheme can help address the domain shift problem, and a transductive learning
scheme can help reduce the hubness phenomenon.

3) Learning an explicit middle-level representation space.
\cite{zhang2015zero,zhang2016zero} advocated the benefits of using attribute-attribute relationships, termed semantic similarity, as the intermediate semantic representation and learned a function to match the image features with the semantic similarity. As an extension of \cite{zhang2015zero,zhang2016zero}, \cite{long2017zero123} aggregated visual representation to a discriminative representation which simplifies $n$ images to one template correspond to one class to achieve high inter-class variation and low intra-class variation, more powerful than large margin mechanism \citep{zhang2015zero,zhang2016zero}, and then mapped semantic embeddings to the discriminative representation space.
Another work was made by \citep{yu2017zero} using matrix decomposition to expand a latent space from the input modality under an implicit process. The intuition they considered is that learning an explicit encoding function between different modalities may be easily spoiled.  Specifically, they learned the optimal intrinsic semantic information of different modalities via decomposing the input features based on an encoder-decoder framework, explicitly learning a feature-aware latent space via jointly maximizing the recoverability of the original space from the latent space and the predictability of the latent space from the original space. To eliminate the limitation of the existing attribute-based methods, i.e., the dependency on the attribute signatures of the unseen classes, sometimes laborious,
\cite{demirel2017attributes2classname} learned a discriminative word representation such that the similarities between class and attribute names follow the visual similarity, and used this learned representation to transfer knowledge from seen to unseen classes.
Similar as \citep{jiang2017learning} for learning latent attributes, \cite{li2018discriminative} proposed to learn the latent discriminative features for ZSL in both visual and semantic space, as well learning features from a region with object instead of pre-trained CNN features. Especially, a category-ranking problem was modeled to learn latent attributes to ensure the learned attributes are discriminative.

\section{Techniques on Experience Learning}\label{section4}
In this section, we will give an overview on the techniques on experience learning. Firstly, we present a mathematical expression for general methodology of experience learning, especially including two categories of approaches along this research line, and then review relevant main techniques.

\subsection{General Methodology for Experience Learning}
Experience learning denotes the machine learning paradigm designed under the circumstance of insufficient samples, also called small data learning. A natural strategy of solving the experience learning is to borrow ideas from LSL techniques, which constitutes the main idea to construct a rational experience learning method. Specifically, a experience learning task can be implemented using the following approaches:
\begin{itemize}
  \item Approach 1: Increase samples and then directly employ conventional  LSL methods;
  \item Approach 2: Utilize small samples to rectify the known models/knowledge learned /obtained from other data sources;
  \item Approach 3: Reduce the dependency of LSL upon the amount of samples to make the method feasible to small samples;
  \item Approach 4: Meta learning.
\end{itemize}
We call the above first strategy as data augmentation (DA) strategy, and the other three as LSL model modification (MD) strategy for convenience. Note that we can perform above strategies simultaneously, instead of using only one in implementing a SSL task. In the following, we will review relevant main techniques of experience learning from these four aspects, respectively.

\subsection{Approach 1: Augmented Data}
A direct manner for experience learning is to generate more data from small training samples to compensate the issue of insufficient data. The LSL methods can then be directly employed for solving the problem. In the following we summarize five kinds of techniques designed in this manner, and in practice possibly more imaginative strategies could be further designed or have been used in practice.

\subsubsection{Deformations}
By the imagination mechanism of human being, more hallucination samples can be formed as the augmented data. This can be realized through using various transformations on original samples, e.g., adding noise, mirroring, scaling, pose and lighting~\citep{kulkarni2015deep}, rotation~\citep{okafor2017operational}, polar harmonic transform~\citep{yap2010two}, radial transform~\citep{salehinejad2017training}, and so on. For example, for audio data, \citet{salamon2017deep} applied four different deformations, namely, time stretching, pitch shifting, dynamic range compression, and background noise to overcome the problem of data scarcity. For domain-specific applications, there exists a requiring for preserving class labels by leveraging task-specific data transformations. \cite{ratner2017learning} directly leveraged user domain knowledge in the form of transformation operations, able to generate realistic transformed data points which were useful
for data augmentation.
Recently, \cite{cubuk2018autoaugment} introduced an automated approach to find data augmentation policies from data, that is, to use a search algorithm in the search space of data augmentation policies like translation, rotation, or shearing to find the best policy such that the neural network yields the highest validation accuracy on a target dataset.

\subsubsection{Generative model}\label{GM}
This strategy attempts to find the generalization model underlying the given small samples, and then use this model to further generate more samples for training.
The early work along this line focused on semi-supervised learning with generative model, aroused the attention of many works~\citep{chen2016infogan,chongxuan2017triple,gan2017triangle,deng2017structured}. Typically,  \citep{chongxuan2017triple} achieved 5\% error rate on SVHN dataset~\citep{netzer2011reading} using 1000 examples~(fewer than 1\% of the whole samples).
Recently, \cite{choe2017face} attempted to generate face images with several attributes and poses using GAN, enlarging the novel set to achieve increased performance on low-shot face recognition task.
Analogously, \cite{shrivastava2017learning} further developed a model called SimGAN that improved the realism of synthetic images from a simulator using unlabeled real data, while preserving the annotation information. Results show that there exists a significant improvement on gaze estimation and hand pose estimation using synthetic images. As an extension of \citep{shrivastava2017learning}, \cite{Lee2018Simulated} leveraged the flexibility of data simulation process and the efficacy of bidirectional mappings between synthetic data and real data.  To enhance few-shot learning systems more efficiently,
\cite{antoniou2017data} proposeed data augmentation GAN (DAGAN) to automatic learn to augment data.
There also exist some works using novel generative model. E.g., \cite{hariharan2017low} learned to hallucinate additional examples for novel classes by transferring modes of variation from the base classes with reconstruction and classification loss to address low-shot learning.
Inspired by the fact that human can easily visualize or imagine what novel objects look like from different views, \cite{wang2018low} trained a hallucinator via meta learning to generate additional examples and provide significant gains for low-shot learning.
Likewise, \cite{vedantam2017generative} firstly tried to define a visually grounded imagination with evaluation metrics of 3C's, i.e. correctness, coverage, and compositionality, and further propose how to create generative models which can imagine compositionally novel concrete and abstract visual concepts via modified VAE. Also, to learn compositional and hierarchical representations of visual concepts, \cite{higgins2017scan} further described symbol-concept association network~(SCAN). Crucially, SCAN can imagine and learn novel concepts that have never been experienced during training with compositional abstract hierarchical representations.

\subsubsection{Pseudo-label method}
This strategy is specifically imposed on small labeled sample set while sufficient unlabeled sample cases (i.e., semi-supervised data).
The augmented data can be obtained through generating confident pseudo-labels by a self ameliorable model, such as curriculum/self-paced learning~\citep{bengio2009curriculum,kumar2010self,jiang2014easy,jiang2014self}, dual learning~\citep{he2016dual}, and data programming~\citep{ratner2016data}.

Curriculum Learning and self-paced learning are learning regimes inspired by the learning process of humans and animals that implements learning through gradually including samples into training process from easy to complex so as to increase the entropy of training samples~\citep{bengio2009curriculum,kumar2010self,jiang2014easy,jiang2014self}. This regime provides a good way to label the grade $t+1$ samples by the learning results at grade $t$ to boost the performance of SSL. For example, \cite{lin2018active} developed a novel cost-effective framework for face identification, that is capable of automatically annotating new instances and incorporating them into training under weak expert recertification. Experiments demonstrated the effectiveness in terms of accuracy and robustness against noisy data under only small fraction of sample annotations.
In object detection, a self-paced learning (SPL) framework was embedded in its optimization process, the selected training images going from ``easy" to ``hard" to gradually improve object detector~\citep{dong2017few}. This method specifically trained the detector using the few annotated images per
category (few-example), and then the detector generated reliable pseudo box-level labels and got improved with these pseudo-labeled bounding boxes. The method can achieve competitive performance compared to state-of-the-art weakly supervised object detection approaches~\citep{diba2017weakly}, with only requirement of about 1\% of the images in the the entire dataset to be annotated.
For salient object detector, \cite{zhang2017supervision} alternately completed the learning procedure without using any pixel-level human annotation: firstly generate reliable supervisory signals from the fusion process of weak saliency models to train the deep salient object detector in iterative learning stages, and then the obtained deep salient object detector is used to update the weak saliency map collection for the next learning stage.
In medical imaging analysis, \cite{li2017self} proposed a self-paced convolutional neural network framework to augment the size of training samples by refining the unlabeled instances, achieving classify computed tomography (CT) image patches with very scarce manual labels. Recently, \cite{meng2015objective} demonstrated an insightful understanding that self-paced learning is robustness to outliers/heavy noises account for learn with a latent non-convex regularized penalty. Therefore, SPL has arose the attention on weakly-supervised learning and webly-supervised learning recently (see Section \ref{WSL}).

Dual learning is firstly propose by \citep{he2016dual} in neural machine translation. Specifically, the method defines primal and  dual tasks, e.g., English-to-French translation versus French-to-English translation, and then can form a closed loop between the primal and dual tasks. In the closed loop, primal translation model can translate unlabelled English to pseudo-French, and dual translation model can translate pseudo-French to pseudo-English. Then difference of groud-truth English and pseudo-English will return rewards using reinforcement learning algorithms to learners. After many iterations, confident pseudo-labelled data can benefit the models. In the recent years, the dual learning framework is successfully applied to visual question answering(VQA)~\citep{li2018visual}, semantic image segmentation~\citep{luo2017deep}, image-to-image translation~\citep{zhu2017unpaired,yi2017dualgan}, zero-shot visual recognition~\citep{chen2018zero}, and neural machine translation~\citep{he2016dual,he2017decoding}. Some improvement of dual learning framework has been further development, such as~\citep{he2017decoding,wu2017sequence,xia2017dual}.

Data programming~\citep{ratner2016data} denotes a technique aiming at learning with labeling functions to generate labeled training sets programmatically. After the raising of this idea, \cite{wu2018fonduer} proposed Fonduer, together with data programming, to provide weak supervision of domain expertise to guide extract information from richly formatted data.
Additionally, there exists other approaches to generate labeled training sets quickly. Typical works include Snorkel~\citep{bach2017snorkel}, SwellShark~\citep{fries2017swellshark}, EZLearn~\citep{grechkin2017ezlearn}, and Flipper~\citep{varma2017flipper}.

\subsubsection{Cross-domain synthesis}\label{cds}
The motivation of cross-domain synthesis approach is to compensate the domain with few samples by knowledge/data from other related domains~\citep{srivastava2012multimodal}. Mathematically, the small samples $\mathcal{S}$ are assumed to be generated from the domain $\mathcal{D}$, which can also be alternatively expressed with $\mathcal{S}_i$ obtained from other related domains $\mathcal{D}_i(i=1,2,\cdots,l)$. By establishing the mapping $P_i :\mathcal{D}_i\rightarrow \mathcal{D}(i=1,2,\cdots,l)$, the augmented data can then be formed as $\mathcal{S}\cup P_1(\mathcal{S}_1) \cup \cdots \cup P_l(\mathcal{S}_l)$. Recent researches have made excellent progress on different fashions of such cross-domain synthesis, benefitting from various generative models (as introduced in Section \ref{GM}), like CNN~\citep{lecun1998gradient}, RNN with the Long Short-Term Memory (LSTM)~\citep{hochreiter1997long}, GAN, VAE, and other techniques including text to image synthesis~\citep{mansimov2015generating,reed2016generative}, attribute to image~\citep{yan2016attribute2image,lample2017fader,dixit2017aga,chen2018semantic},  image to text synthesis~\citep{vinyals2015show,karpathy2015deep,xu2015show,ren2015exploring}, image to image synthesis (translation, style transfer)~\citep{isola2017image,zhu2017unpaired,gatys2016image,johnson2016perceptual}, text to speech~\citep{van2016wavenet,anderson2013expressive,gibiansky2017deep}, video to speech~\citep{owens2016visually}, speech to video~\citep{deena2009speech}, and so on. Crucially, cross-domain synthesis like text to image or attribute to image takes an important role in many techniques on concept learning (see Section \ref{EM}).

A representative case necessary to use this technique is medical imaging analysis. In this practical domain, high-quality supervised samples (e.g., labeled with certain diseases) are generally scarce, expensive, and fraught with legal concerns regarding patient privacy. For these issues, cross-domain image synthesis has recently gained significant interest. The main research contents focus on image to image synthesis, including cross MRI image synthesis~\citep{ye2013modality,van2015cross,vemulapalli2015unsupervised,joyce2017robust,chartsias2018multimodal}, MR image to CT image synthesis~\citep{roy2014mr,huynh2016estimating,torrado2016fast,cao2017dual}, and label map to MRI~\citep{cordier2016extended}. For example, \cite{van2015cross} proposed location sensitive deep network(LSDN), improving MRI-T1 image to MRI-T2 image results by conditioning the synthesis on the position in the volume from which the patch comes. To be more general, \cite{joyce2017robust} tried to synthesize MRI FLAIR with multi-input, i.e., MRI T1, T2, and DWI, robust to missing data and misaligned inputs via learning
a modality-invariant latent representation. Afterwards, \cite{chartsias2018multimodal} extended \cite{joyce2017robust}'s work, easily predicting new output modalities through the addition of decoders which can be trained in isolation.

\subsubsection{Domain adaptation / data transportation} \label{dt}
\begin{figure}
  \centering
  \includegraphics[width=6in]{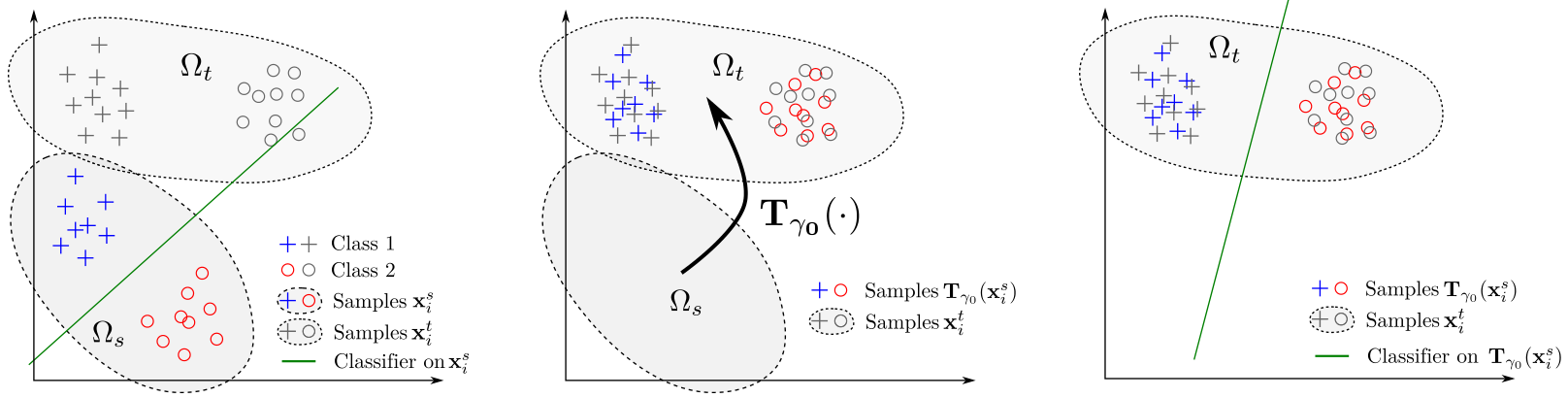}\\
  \caption{Illustration of the approach of data transportation for augmented data (image is reproduced from \citep{courty2017optimal}).}\label{da}
\end{figure}

An SSL problem can be handeled through borrowing the solution of the same type of other learning problems, which refers to the domain adaptation problem~\citep{pan2010survey}. Readers can refer to \citep{csurka2017domain,venkateswara2017deep} for more technical details. The learning manner is to transform the data from source domain, with sufficient annotated ones, by a differential homomorphism $\mathbf{T}: \Omega_s\rightarrow\Omega_t$ with certain constrains, to help form an augmented data set to compensate the small sample set collected from the target domain, with only few or no annotated data to obtain more rational solution (see Fig.\ref{da}).

The early works proposed in this manner were constructed based on instance re-weighting \citep{zadrozny2004learning,sugiyama2008direct,kanamori2009efficient,huang2007correcting,yan2017mind}, which estimated the ratio between the likelihoods of being a source or target example or use maximum mean discrepancy (MMD) measure~\citep{borgwardt2006integrating} to weight data instances. 
A second- or higher-order knowledge transfer\citep{koniusz2017domain} was used to bring closer the within-class scatters while maintaining good separation of the between-class scatters, and was further investigated on Action Recognition datasets\citep{tas2018cnn}.
Another research direction focused on transformation, which matched both source and target domain under some constraints. For example, \citep{saenko2010adapting} learned a liner transformation between two domains by minimizing the effect of
domain-induced changes in the feature distribution. And the non-linear transformation between the two domains was investigated by
\citep{long2014transfer} through minimizing the distance between the empirical expectations of source and target data distributions integrated within a kernel embedding.
A typical example was to learn a transportation plan with the optimal transport theory~\citep{courty2017optimal}, which constrained labeled samples of the same class in the source domain to remain close during transport. \cite{courty2017joint} went a step further to implicitly learn a non-linear transformation that minimized the optimal transport loss between the joint source distribution and an estimated target joint distribution, corresponding to the minimization of a bound on the target error. To compensate for the lack of target structure
constraint, \cite{liang2018aggregating} added a novel relaxed domain-irrelevant clustering-promoting term that jointly bridged the cross-domain semantic gap and increased the intra-class compactness in both domains.
A local sample-to-sample matching method was developed by\citep{das2018sample} recently, in which the source and target samples are treated as graphs.
Another strategy was proposed in \citep{bousmalis2017unsupervised} to propose a pixel-level domain adaptation method (PixelDA), which used GAN to learn a transformation from source domain to target domain. This mechanism is same as cross-domain synthesis using GAN (Section \ref{cds}). Along this line, some works were released like \citep{taigman2016unsupervised,tzeng2017adversarial,murez2017image,volpi2017adversarial}, and boosted the performance in object recognition~\citep{hu2018duplex}, person re-identification~\citep{deng2017image}, brain lesion segmentation~\citep{kamnitsas2017unsupervised}, and semantic segmentation~\citep{hong2018conditional}.
The latest work focused on the large domain shifts between source and target datasets in \citep{koniusz2018museum}, and a new dataset called open museum identification challenge (Open MIC) was released.

\subsection{Approach 2: Rectify the Known Models/knowledge with Small Samples}
Under the assumption that the knowledge in knowledge system learned before can share with future learning, the SSL strategy can be rationally constructed by rectifying
the known models/knowledge in knowledge system to adapt the new observations. We will present several techniques of rectification in this section.

\subsubsection{Fine-tuning}\label{section311}
This approach aims to achieve the better cognition level through updating or fine-tuning the current knowledge (a trained model) by small training samples. In this scheme, the small sample set is used to update or finetune the existing model (e.g., a trained deep network)~\citep{yosinski2014transferable,oquab2014learning,hinton2006reducing,Krizhevsky2012Alex}. In practice, it always pretrains a basic model on a source domain (where data are often abundant), and then fine-tunes the trained model~(sometimes change several output layers' topology structure but fixes other layers) on a target domain (where data are insufficient). The motivation is that there exists common representation among various objects in nature, and novel samples can adaptively fit the new similar tasks after the basic model extracts common representation knowledge among many objects.
\begin{figure}
  \centering
  \includegraphics[width=6in]{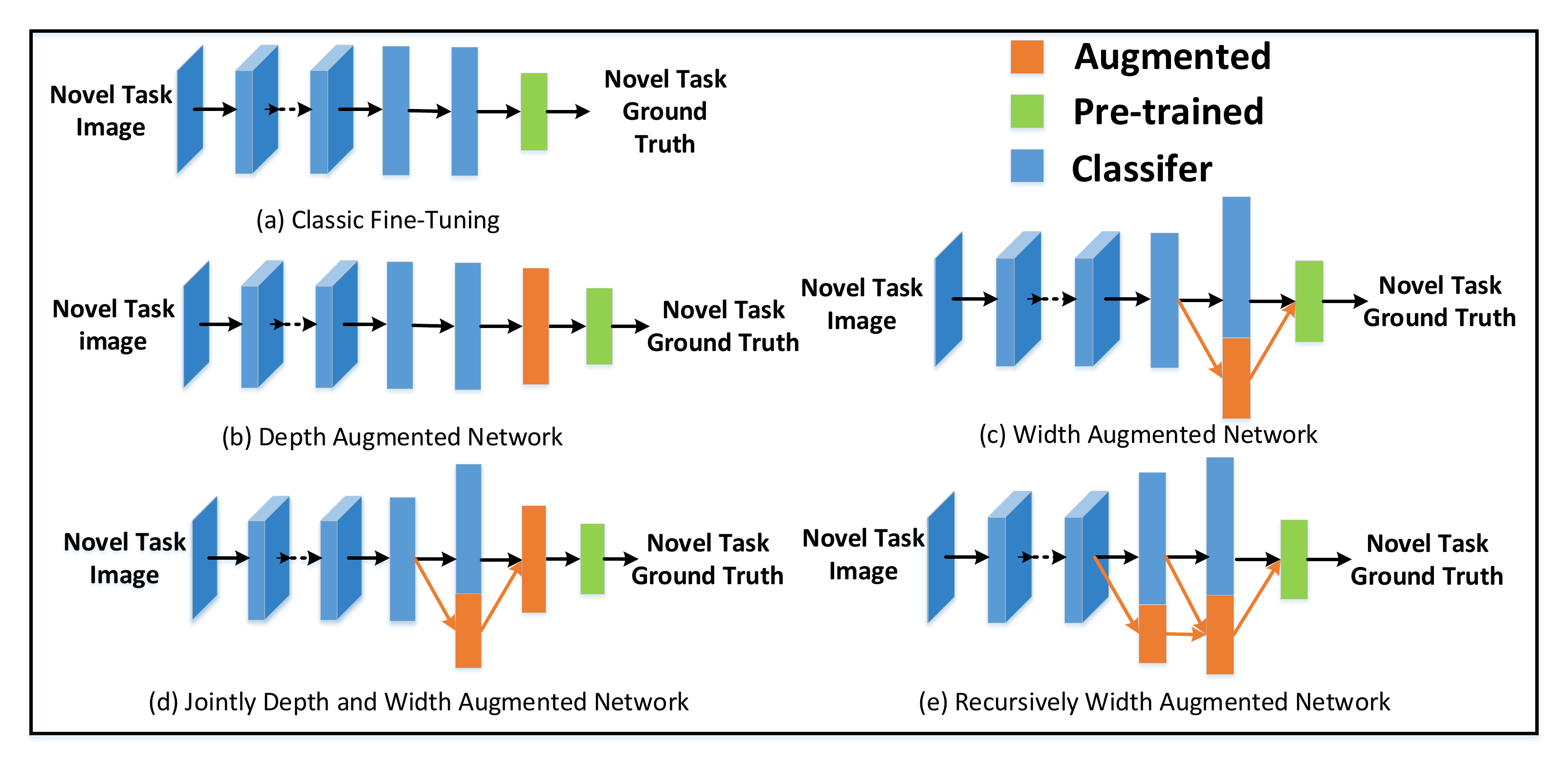}\vspace{-5mm}
  \caption{Illustrations of several fine-tuning networks (image is reproduced from \citep{wang2017growing}). (a) lassic fine-tuning, (b-e) variations of developmental networks~\citep{wang2017growing} with augmented model capacity.}\label{f5}
\end{figure}
For example, in object detection~\citep{girshick2014rich,girshick2015fast,ren2015faster}, the CNN on the ImageNet ILSVRC2012 classification task is pre-trained and fine-tuned to fit the detection task. In SSL, fine-tuning has been successfully applied to visual classifiers in new domains~\citep{chu2016best}, object detection with long-tail distribution~\citep{ouyang2016factors}, neural machine translation~\citep{chu2017empirical}, remote sensing scene classification~\citep{fang2016using}, museum artwork identification\citep{zhang2018artwork} and medical image analysis~\citep{tajbakhsh2016convolutional,shin2016deep}.

Recently, some progresses have been made to improve the fashions of fine-tuning. For example, progressive networks~\citep{rusu2016progressive} solved multiple independent tasks at the end of training without assumptions about the relationship between tasks, and modifies or ignores previously learned task features via the lateral connections. Yet previous tasks were not affected by the newly learned features in the forward pass. Progressive networks were originally proposed for reinforcement learning to transfer knowledge for a simulated robotic environment to a real robot arm, massively reducing the training time required on the real world~\citep{rusu2017sim}. The block-modular architecture~\citep{terekhov2015knowledge} is similar work while more focused on a visual discrimination task. Developmental Networks~\citep{wang2017growing} explored several routes for increasing model capacity during fine-tuning (see Fig.\ref{f5}), both in terms of going deeper (more layers) and wider (more channels per layer). Such strategy achieved good performance beyond classic fine-tuning approaches on certain tasks.

\subsubsection{Distillation}\label{section314}
\begin{figure}
  \centering
    \subfigure[Teacher and Student Networks]{
    \label{ST:a} 
     \includegraphics[width=2.7in]{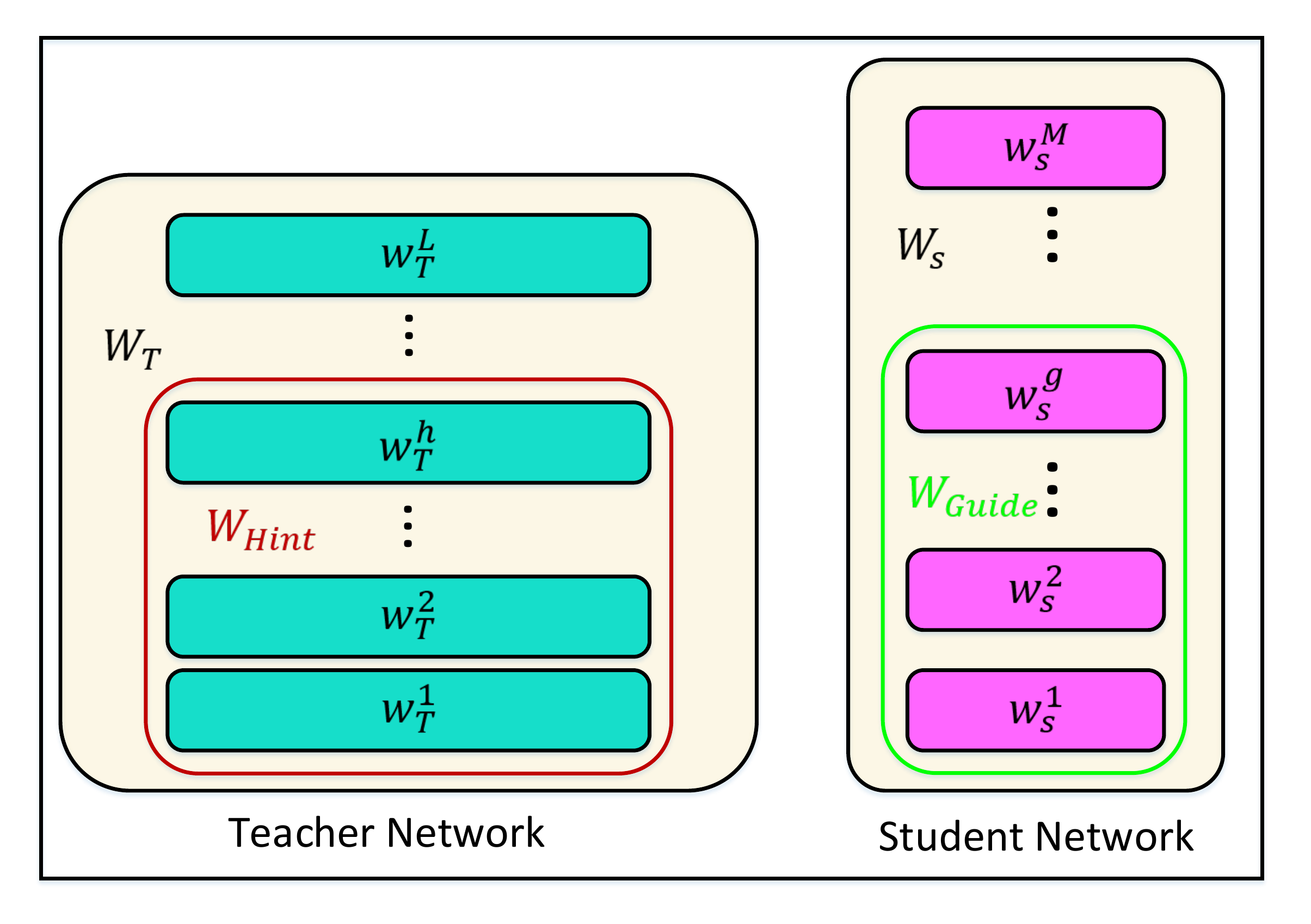}}
  \subfigure[Two examples of TSN]{
    \label{dk1:b} 
    \includegraphics[width=2.4in]{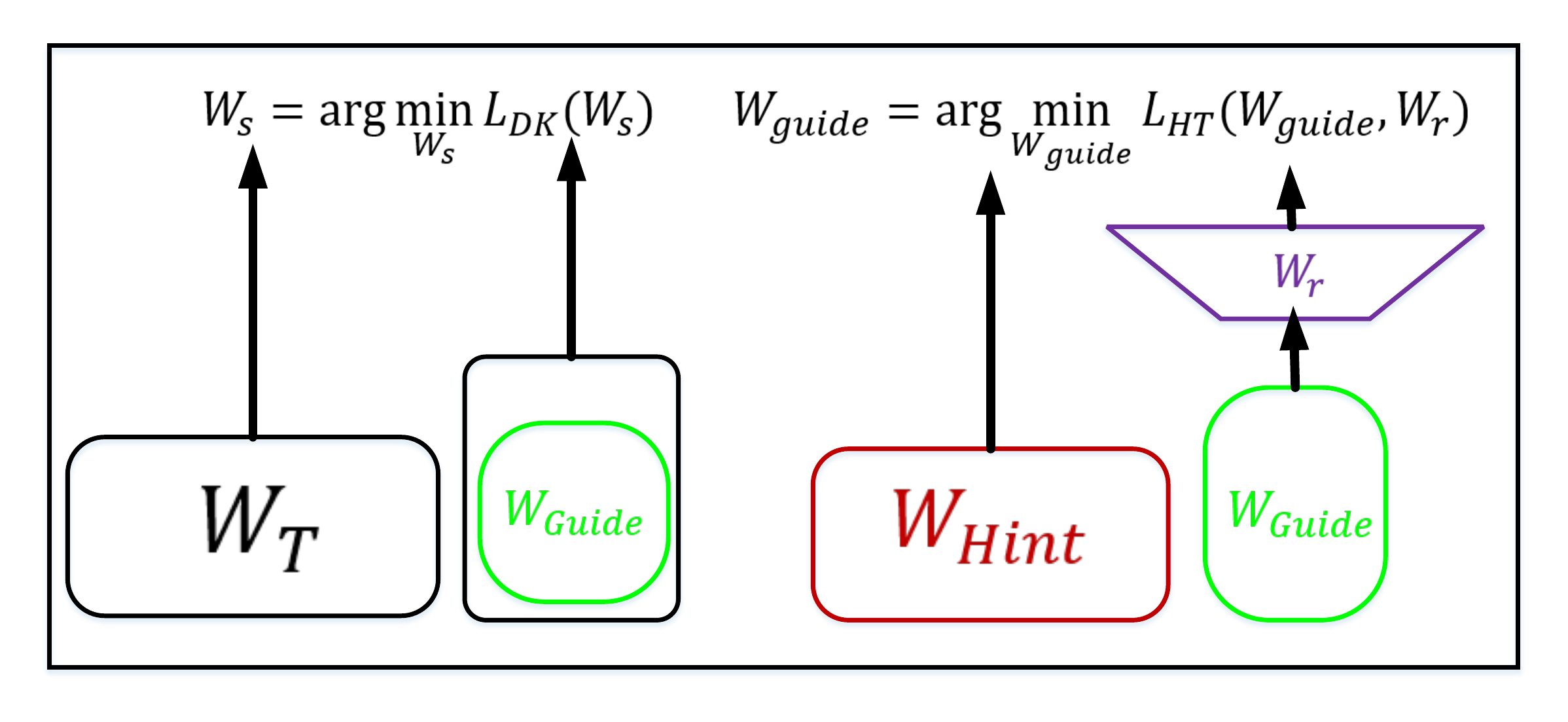}}
  \caption{Illustration of the Teacher and Student Network approach (images are reproduced from \citep{romero2014fitnets}).}\label{figST}
\end{figure}

Knowledge distillation~\citep{hinton2015distilling} is certain form of Knowledge Transfer (KT) approach, and the motivation is transferring knowledge from an ensemble or from a large highly regularized model into a smaller, distilled model, as well as capturing the
information provided by the true labels on small sample dataset. This learning form is sometimes called teacher-student networks(TSN) (see Fig.\ref{ST:a}), where the student $f^{s}$ is penalized according to a softened version of the teacher's $f^t$ output or ensemble of teacher networks. Formally, The parameters of the student network model $W_{s}$ are learned by minimizing a loss with the form
 \begin{align}\label{eqs}
   W_s = \arg\min_{W_s} L_{KD}(W_s) = \mathcal{H}(y,f^s)+\lambda\mathcal{H}(f^t,f^s),
 \end{align}
where $\mathcal{H}$ refers to the cross-entropy and $\lambda$ is a tunable parameter to balance both cross-entropies. The first term in Eq.(\ref{eqs}) corresponds to the traditional cross-entropy between the output of a (student) network and labels, whereas the second term enforces the student network to be learned from the softened output of the teacher network(see Fig.\ref{dk1:b}).
Alternatively, inspired by curriculum learning strategies~\citep{bengio2009curriculum}, which organized the training examples in a gradually more complex manner, such that the learner network gradually received examples of increasing difficulty w.r.t. the already learned concepts, \cite{romero2014fitnets} introduced a hint-based learning concept to train the student network. Particularly, they utilized not only the outputs but also the intermediate representations learned by the teacher as hints to improve the training process and final performance of the student. Mathematically, they trained the student network parameters from the first layer up to the guided layer as well the regressor parameters by minimizing the following loss function (see Fig.\ref{dk1:b}):
\begin{align}
\begin{split}
  W_{Guide}= &\arg\min_{W_{Guide}}L_{HT}(W_{Guide},W_r)\\
  =&\frac{1}{2}\|u_h(x;W_{Hint})-r(v_g(x;W_{Guide});W_r)\|_2^2,
\end{split}
\end{align}
where $u_h$ and $v_g$ are the teacher/student functions up to their respective hint/guided layers with parameters $W_{Hint}$ and $W_{Guide}$, and $r$ is the regressor function on top of the guided layer with parameters $W_r$. Here, the outputs of $u_h$ and $r$ are expected to have the similar structure. Recently, \cite{yim2017gift} proposed to define distilled knowledge in terms of flow between layers, which was calculated by computing the Gram matrix of features from two different layers. They demonstrated that novel technique optimized fast, and the student network outperformed the original network even trained at a different task. Another attempt was made by \citep{huang2017like} treating KT as a distribution matching problem, that is, matching the distributions of neuron selectivity patterns between teacher and student networks by minimizing the MMD metric. Different from the above methods via model distillation, \cite{radosavovic2017data} investigated omni-supervised learning, a data distillation method, ensembling predictions from multiple transformations of unlabeled data, using a single model, to automatically generate new training annotations.

The KT approach is promising to facilitate the SSL task, and has made some processes recently. For example, \cite{luo2017graph} proposed a graph-based distillation method to distill rich privileged information from a large multi-modal dataset to teach student tasks models, which tackled the problem of action detection on limited data and partially observed modalities. To address the issue of detecting new classes objects,
\cite{shmelkov2017incremental} proposed a method for not only adapting the old network to the new classes with cross-entropy loss,  but also ensuring performance on the old classes does not catastrophic forget with a new distillation
loss which minimized the discrepancy between responses for old classes from the original and the new networks. Experiments demonstrated that the method can perform well even  in the extreme case of adding new classes one by one. Likewise, \cite{chen2018lstd} investigated low-shot object detection, and KT helped transfer object-label knowledge for each target-domain proposal to generalize low-shot learning setting.
In biological domain, \cite{christodoulidis2017multisource} firstly trained a teacher network from six general texture databases of similar domain, and then the model was fine-tuned on the limited number of lung tissue data, and finally transferred knowledge in an ensemble manner. Their fused knowledge was distilled to a network with the original architecture, with 2\% increasing in the performance.

\subsubsection{Domain adaptation/model adaptation}
\begin{figure}
  \centering
    \subfigure[Adapting existing model]{
    \label{adapation:a} 
     \includegraphics[width=2.6in]{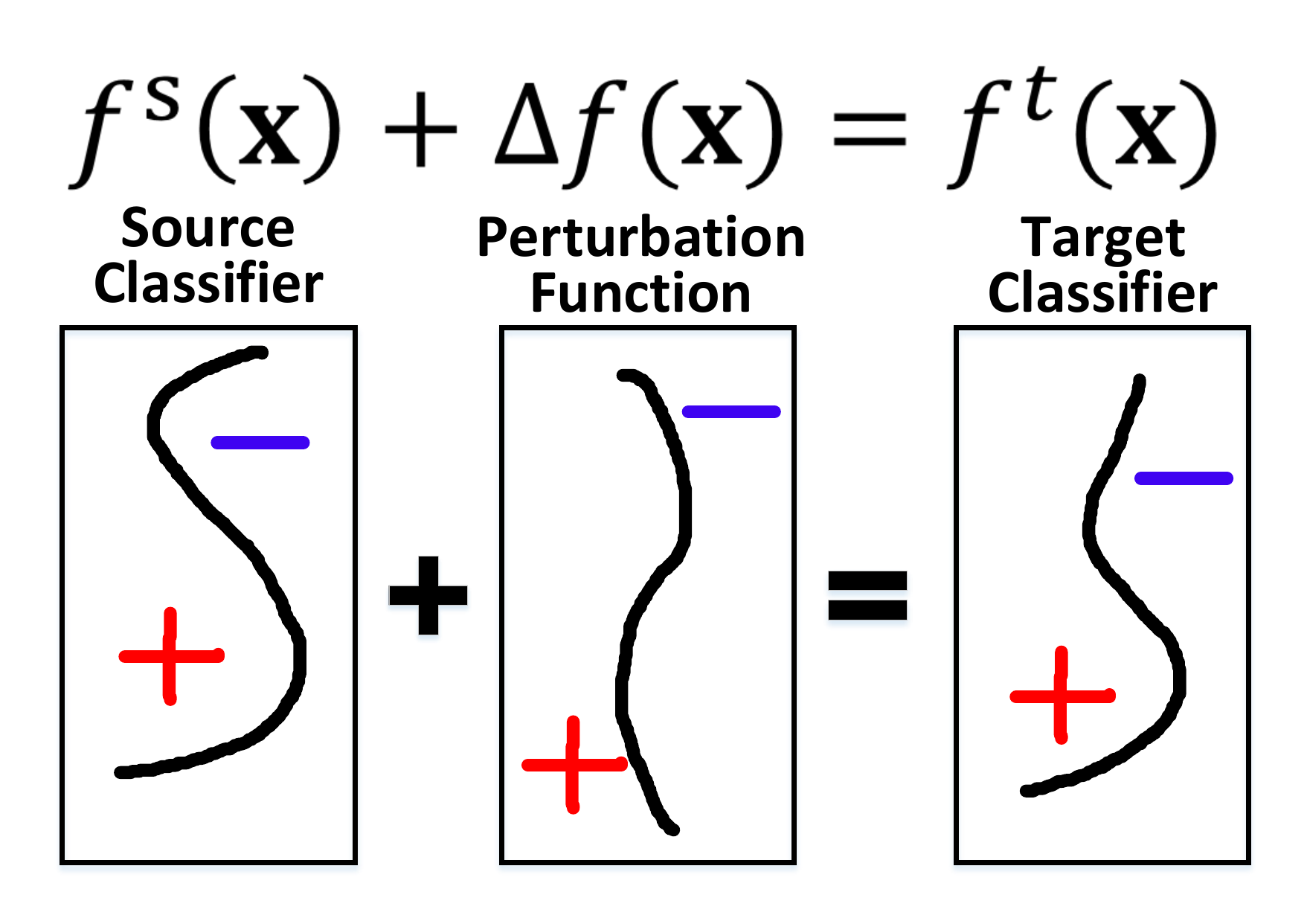}}
  \subfigure[Diagram of model adaptation]{
    \label{example:b} 
    \includegraphics[width=2.6in]{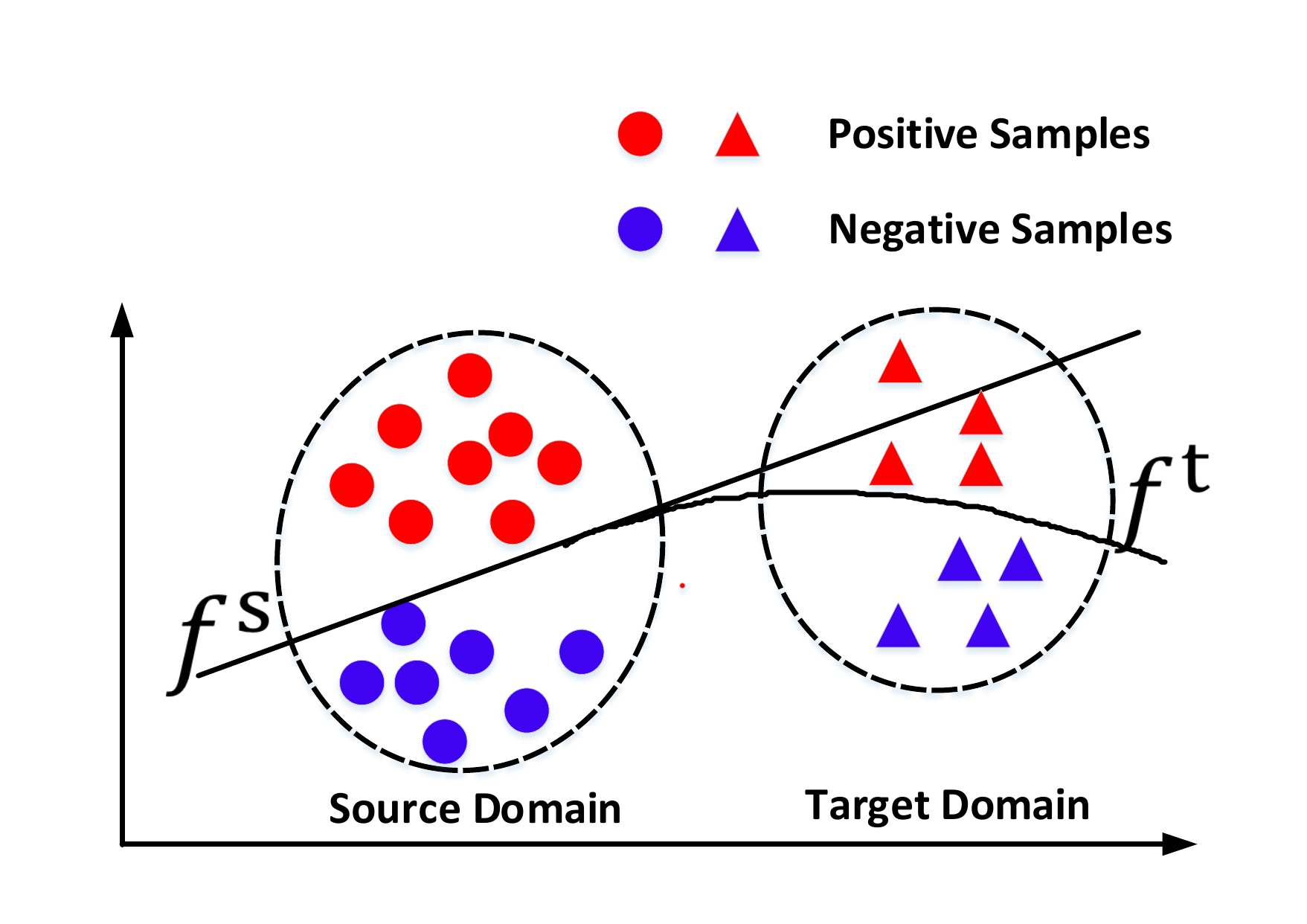}}
  \caption{Illustration of the model adaptation approach (image is reproduced from \citep{csurka2017domain}).}\label{figST}
\end{figure}

Except from data transportation in domain adaptation (Section \ref{dt}), there exists another learning fashion: model adaptation. The insight is to adapt one or more existing models in knowledge system to the small sample dataset. The early work assumes that target model (classifier) consists of the source models (existing) and perturbation functions (see Fig.\ref{adapation:a}). \cite{yang2007cross} proposed adaptive support vector machines (A-SVMs), where a set of so called perturbation functions were added to the source classifier to progressively adjust the decision boundaries of target classifier in the target domain. The diagram of source classifier and target classifier can be understood by seeing Fig.\ref{example:b}. Along this research direction, cross-domain SVM~\citep{jiang2008cross}, domain transfer SVM~\citep{duan2009domain}, domain adaptation SVM~\citep{bruzzone2010domain}, adaptive multiple kernel learning (A-MKL)~\citep{duan2012exploiting}, and residual transfer
network~(RTN)~\citep{long2016unsupervised} have been progressively proposed. Particularly, \cite{long2016unsupervised} extended this idea to deep neural networks. Similarly, \cite{rozantsev2017residual} introduced a residual transformation network to relate the parameters of two domain network architecture. Considering that model adaptation and data transportation (Section \ref{dt}) are not independent, optimizing both the transformation and classifier parameters jointly was also developed in \citep{shi2012information,hoffman2013efficient,saito2017maximum}.

With the recent booming of deep neural network techniques, a naive regime to be easily formulated is the fine-tuning strategy (Section \ref{section311}). Through fine-tuning the pretrained networks with target data, an efficient adaptation can be naturally guided. When domain discrepancy between the source and the target is very large, this adaptation manner might not work. This inspires the idea of minimizing the difference in learned feature covariances across domains, which guarantees that fine-tuning can ameliorate the performance. In this learning manner, \cite{tzeng2015simultaneous} combined domain confusion and softmax cross-entropy losses to train the network with the target data, where domain confusion loss tried to learn domain-invariant representations, while softmax cross-entropy loss ensured the output feature representations of the source and target data were distinct. \cite{long2015learning} extended \citep{tzeng2015simultaneous}'s domain confusion loss by incorporating an MMD loss for all of the fully connected layers ($fc6,fc7$ and $fc8$) of the AlexNet. Furthermore, they combined target classifier adaptation with residual learning~\citep{yang2007cross} and feature adaptation with MMD loss in \citep{long2016unsupervised}.
Another attempt using adversarial loss was made by \citep{ganin2015unsupervised}. Specifically, they achieved the adaptation by augmenting a gradient reversal layer connecting the bottom feature extraction layers and the domain classifier, whose function was similar to the discriminator in GAN. In this way, the feature extractor was trained to extract domain invariant features. Against possible overfitting issue during the fine-tuning stage, \cite{sener2016learning} presented an end-to-end deep learning framework to learn domain transformation, through jointly optimizing the optimal deep feature representation and target label inference.

\subsection{Approach 3: Reduce the Dependency of LSL upon the Amount of Samples}
One of the significant issues of LSL is that its effectiveness is generally dependent on large amount of training dataset, that is, the regime is constructed in a data-driven rather than a model-driven manner. A rational paradigm to reduce the dependency of LSL upon the amount of samples is to strengthen the power of a machine learning model to reflect more insights underlying sample domains. We first revisit a general machine learning model as follows:
\begin{align}
  \min_{f\in \Phi}\mathcal{L}(D,f(w))+p(w), \label{equation2}
\end{align}
where $\Phi$ is hypothesis, $f\in \Phi$ is the learner, $\mathcal{L}(D,f(w))$ is loss/cost function measuring discrepancy between predicting output $f(w)$ and ground-truth input, and $p(w)$ is the regularizer. We can then introduce the following manners for this model-strengthen task.

\subsubsection{Model-driven Small Sample Learning}
Using proper models to confine hypothesis space in machine learning (or topology of a neural network) tends to relax the dependence of a learning algorithm on the amount of samples. Following this idea, several promising regimes have been raised recently. we will introduce them in the remainder of this section.

\paragraph{White Box Model:}
The white box model denotes a popular strategy presented recently for improving interpretability against deep learning, which is known to have issues of unclear working mechanism, namely, black box models. The direct work is to whiten deep neural network like CNN. Based on the idea of encoding objects in terms of visual concepts (VCs), \cite{deng2017unleashing} developed an interpretable CNN model for few-shot learning, where the VCs were extracted from a small set of images of novel object categories using features from CNNs trained on other object categories. Recently, \cite{tang2017towards} proposed a composition network (CompNet) through combining And-Or graphs (AOGs) with CNN models, where the learned compositionality is fully interpretable. Alternatively, \cite{garcia2017few} defined a graph neural representations, which cast few-shot learning as a supervised message passing task.

The unfolded approach was pioneered in \citep{gregor2010learning}, where the authors unrolled the ISTA algorithm for sparse coding into a neural network. In this scheme, filters that are normally fixed in the iterative minimization are instead learned. Recently, \cite{yang2016deep} unrolled the ADMM algorithm to design a CNN for MRI reconstruction, demonstrating performance equivalent to the state-of-the-art with advantages in running time but with few training images. This reflects the main idea of the model-driven deep-learning~\citep{xu2017model}, combining the model-based and deep-learning-based approaches. This paradigm can incorporate domain knowledge into model family, and then establish the algorithm family to solve the model family, and the algorithm family could be unfolded to a deep network to learn the unknown parameters in the algorithm family.
Along this line of research, various unfolded technique have been successfully applied into dynamic MR image reconstruction~\citep{schlemper2018deep,qin2017convolutional}, sparse view/data Computed Tomography(CT) reconstruction~\citep{gupta2018cnn,chen2017learned,adler2018learned}, compressive image reconstruction~\citep{metzler2017learned,zhang2018ista}. Especially, \cite{diamond2017unrolled} presented a framework infusing knowledge of the image formation into deep networks that solved inverse problems in imaging by leveraging unrolled optimization with deep priors, outperforming the state-of-the-art results for a wide variety of imaging problems, such as denoising, deblurring, and compressed sensing magnetic resonance imaging(MRI).

\paragraph{Memory Neural Networks:}
Inspired by episodic memory (see Section \ref{section221}), researchers try to endow neural networks with memory~\citep{sukhbaatar2015end,graves2014neural}. \cite{santoro2016meta} firstly applied this mechanism into SSL. Specifically, they proposed memory-augmented neural networks (MANN) though combining with more flexible storage capabilities and more generalized deep architectures, namely, the ability to rapidly bind never-seen information after a single presentation and the ability to slowly learn an abstract method for obtaining useful representations of raw data. MANN was thus expected to achieve efficiently inductive transferring knowledge, which means that new information can be flexibly stored and precisely inferred based on novel data and long experience. Subsequently, some works tried to follow~\citep{santoro2016meta} and further enhance its capability. For easy reference, a list of MANN's variations is displayed in Table \ref{table2}.

\begin{table}[t]
\begin{center}
\begin{tiny}
\begin{tabular}{lcccr}\toprule[1pt]
\textbf{Methods} & \textbf{Setting}& \textbf{Improvements}&\textbf{Datasets} & \textbf{References} \\ \hline
MANN & one-shot learning & --- &Omniglot &  \citet{santoro2016meta}\\
Scaling MANN & one-shot learning &scale to space and time& Omniglot & \cite{rae2016scaling} \\
MANN with Gaussian & one-shot learning & structured generative model &Omniglot&\citet{Harada2017}\\
embeddings &&&&\\
LMN & few-shot learning &online adapting&Omniglot&\citet{Shankar2018}\\
FLMN & one-shot/zero-shot learning & memory interference &Omniglot,&  \citet{mureja2017meta}\\
&&&miniMNIST&\\
Life-long Memory &life-long one-shot learning &scale to large memory&Omniglot& \cite{kaiser2017learning}\\
Module MAVOT & one-shot learning  &  long-term memory &  ImageNet &\citet{liu2017mavot}\\
&for video object tracking&&ILSVRC2015&\\
Augmented LSTM & few-shot learning & long-term memory of  & VQA benchmark & \citet{Ma2018CVPR_a}\\
&&scarce training exemplars& Visual 7W Telling &\\
Memory-Augmented  & one-shot learning&rapidly adapting &  pulmonary & \citet{mobiny2017lung}\\
Recurrent Networks&&&lung nodules&\\
\bottomrule[1pt]
\end{tabular}
\end{tiny}
\end{center}
\vskip -0.1in
\caption{\small Memory-Augmented Neural Networks(MANN) and its variations.}\label{table2}
\end{table}

In details, \cite{rae2016scaling} incorporated sparse access memory (SAM) in MANN to help scale in both space and time as the amount of memory grows, facilitating the method capable of making use of efficient data structures within the network, and obtaining significant speedups during training. Alteratively, \cite{kaiser2017learning} tried to enhance large-scale memory using fast nearest-neighbor algorithms.
Another work was investigated in \citep{Harada2017}, and they constructed a memory augmented network with Gaussian embeddings capturing latent structure  based on the disentanglement of content and style instead of pointwise embeddings. To establish an online model adaptation, \cite{Shankar2018} proposed labeled memory network (LMN) with a label addressable memory module and an adaptive weighting mechanism. As an extension of \citep{Shankar2018}, \cite{mureja2017meta} further proposed feature-label memory network (FLMN) explicitly splitting the external memory into feature and label memories, outperforming MANN~\citep{santoro2016meta} by a large margin in supervised one-shot classification tasks.
In terms of  one-shot learning for video object tracking, \cite{liu2017mavot} employed an external memory to store and remember the evolving features of the foreground object as well as backgrounds over time during tracking, making it possible to maintain long-term memory of the object. To solve the long-tailed distribution of the question-answer pairs on the VQA benchmark dataset~\citep{antol2015vqa},  \cite{Ma2018CVPR_a} developed MANN to increase capacity to remember uncommon question and answer pairs.
In biological domain, \cite{mobiny2017lung} extended MANN to CT lung nodule classification, adapting to the new CT image data received from a never-before seen distribution. \cite{chen2018sequential} further introduced the memory mechanism to recommender
systems. They developed MANN integrated with collaborative filtering to help recommendation in a more explicit, dynamic, and effective manner.

\begin{figure}
  \centering
  \subfigure[Diagram of NMN]{
  \label{fig12a}
  \includegraphics[width=2.9in]{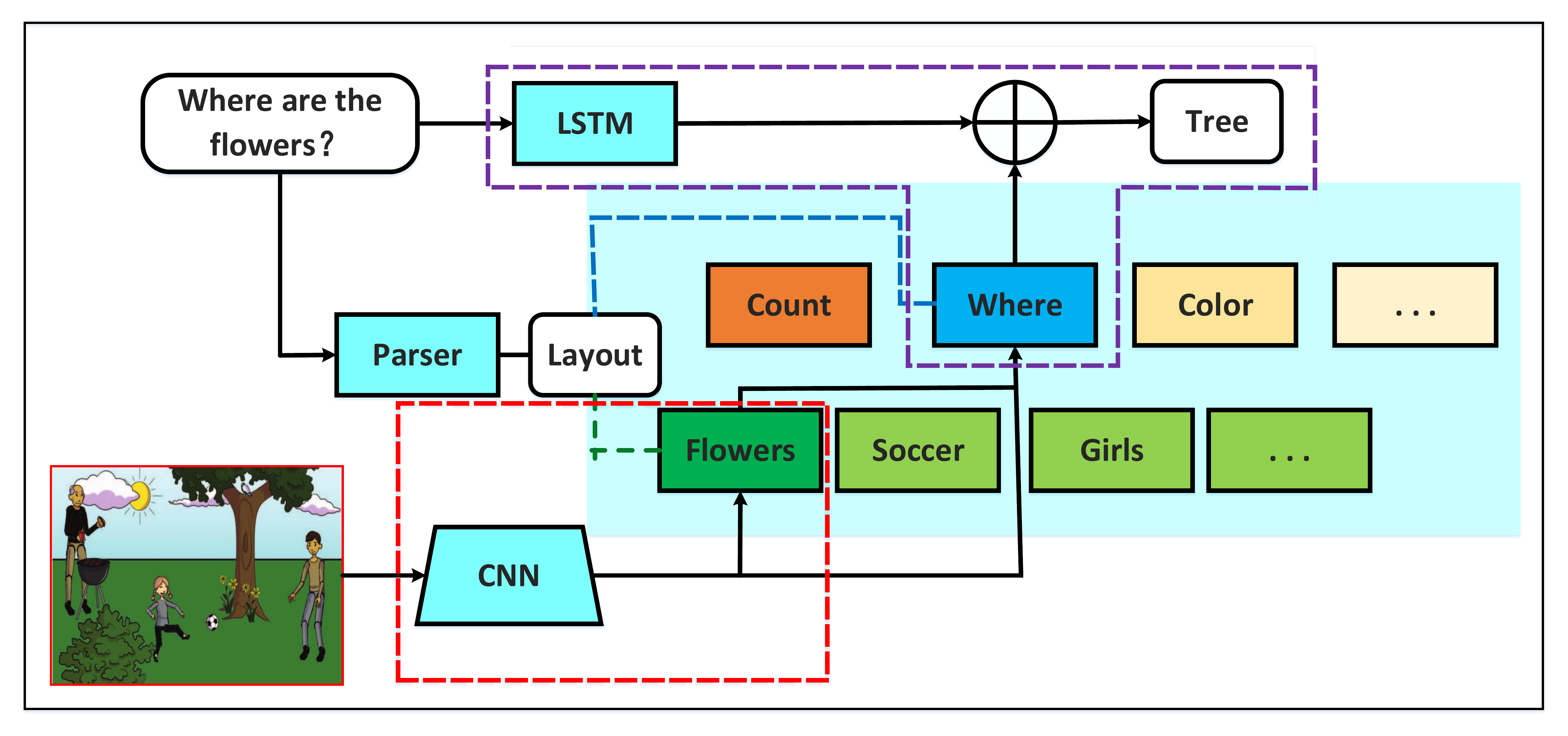}}
  \subfigure[Sets of NMN]{
  \label{fig12b}
  \includegraphics[width=2.9in]{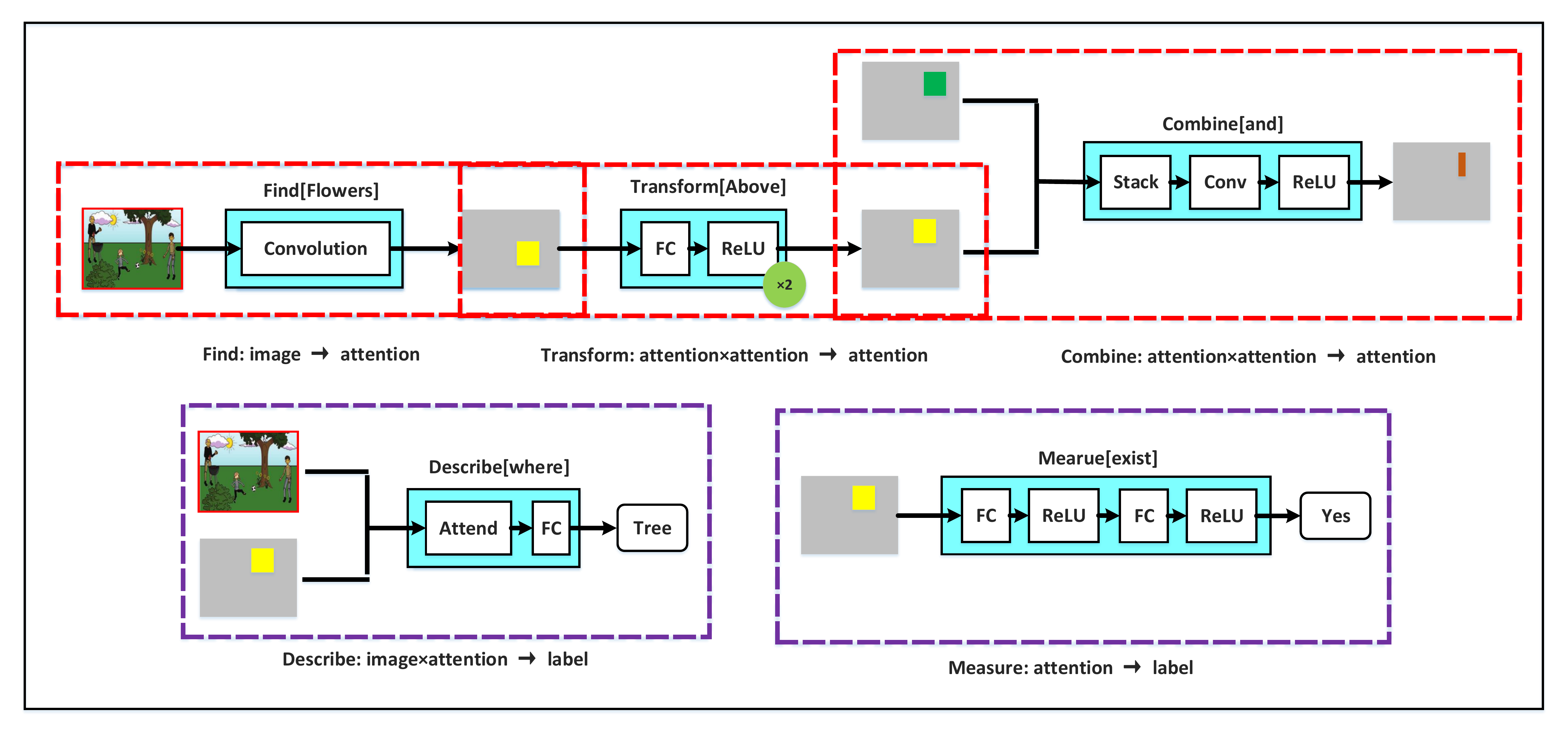}}\vspace{-3mm}
  \caption{\small Illustration of Neural Module Networks(NMN) (images are reproduced from \citep{andreas2016neural}). (a) Diagram of NMN. The green area is a neural module network, and the parser NMN dynamically lays out a deep network composed of reusable modules  by leveraging a natural language technique. (b) Sets of NMN. The red dotted box corresponds to that in (a). Particularly, \texttt{Find[flowers]} module produces a matrix whose entries should be large in regions of the image containing dogs, and small everywhere else; \texttt{Transform[Above]} module shifts the attention regions upward and \texttt{Combine[and]} modules should be active only in the regions that are active in both inputs. In terms of purple dotted box, \texttt{Describe[where]} module returns a representation where are the flowers in the region attended to, i.e., around the tree; \texttt{Mearue[exist]} module evaluates the existence of detected flowers and trees. The figure is better viewed in color and zoomed in on a computer screen.}\label{fig12}
\end{figure}
\paragraph{Neural Module Networks:}
Neural Module Networks \citep{andreas2016neural,andreas2016learning,hu2017modeling} are composed by collecting jointly-trained neural modules, which can be dynamically assembled into arbitrary deep networks. When we want to use the previously trained model on a new task, we can assemble these modules dynamically to produce a new network structure tailored to that task (an illustration of NMN architecture is depicted  in Fig.\ref{fig12}).
Along the line of NMN, relation networks~(RNs)~\citep{santoro2017simple}, end-to-end module networks (N2NMN)~\citep{hu2017learning}, program generator+execution engine (PG+EE)~\citep{johnson2017inferring}, thalamus gated recurrent module~(ThalNet) \citep{hafner2017learning} and feature-wise linear modulation~(FiLM)~\citep{perez2018film} have been developed. And FiLM is demonstrated that can generalize well to challenging, new data from few examples or even zero-shot settings.

\subsubsection{Metric-driven Small Sample Learning}\label{MSSL}
The metric learning idea along this research line is to learn a mapping from inputs to vectors in an embedding space to make the inputs of the same identity or category closer than those of different identities or categories (more discriminative than original input space)~\citep{kulis2013metric,lu2017deep}. Once the mapping is learned, at test time a nearest neighbors method can be used for retrieval and classification without retraining models for new categories that are unseen during training. Through putting emphasis on more high-quality samples while depressing those low-quality ones, the dependence of the method to the size of samples can be more or less reduced.

\begin{figure}
  \centering
  \subfigure[\scriptsize Siamese Networks for Chromosome Classification]{
  \label{fig:metrica}
  \includegraphics[width=2.7in]{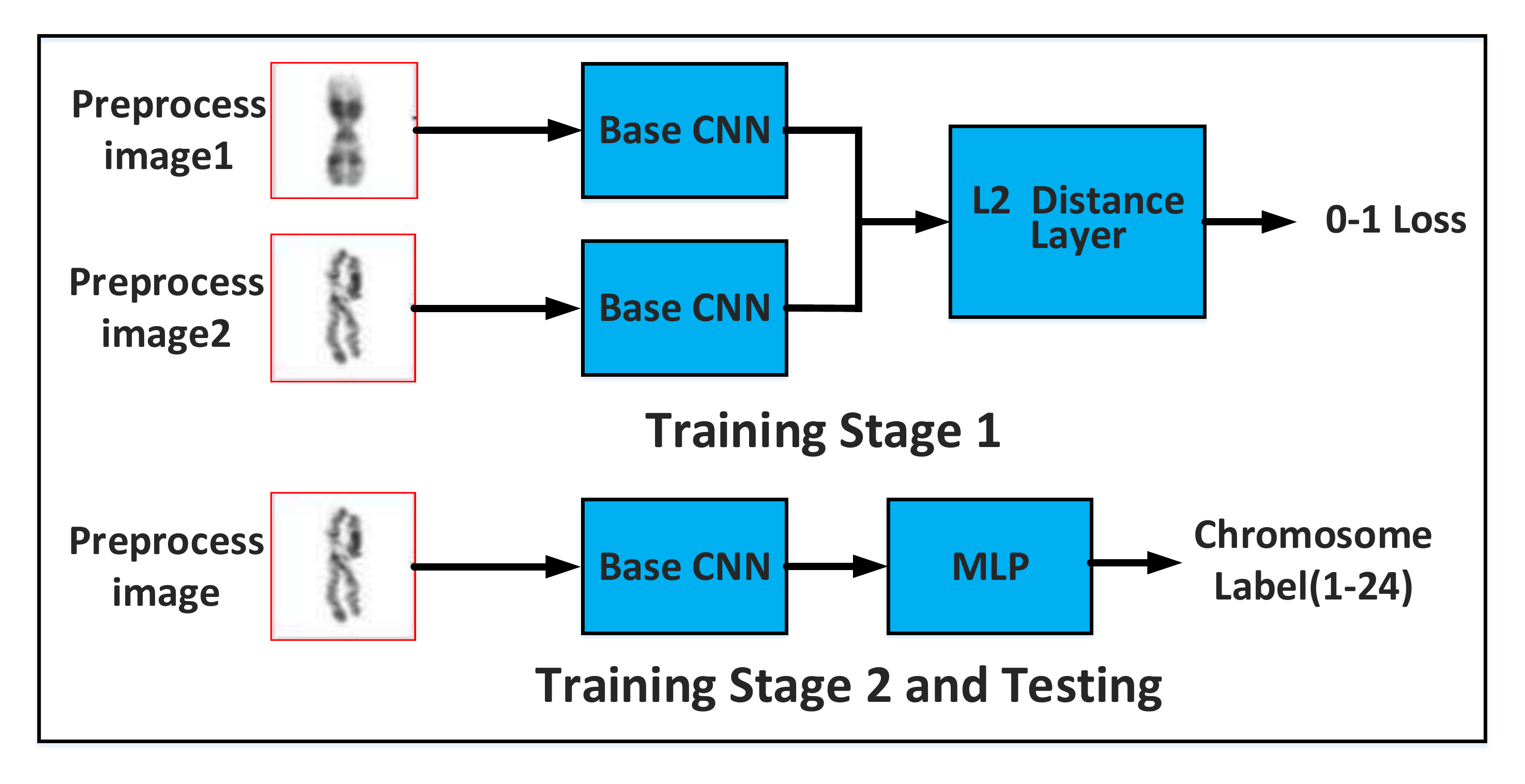}}
    \subfigure[\scriptsize Matching Network for Low Data Drug Discovery]{
  \label{fig:metricb}
   \includegraphics[width=2.7in]{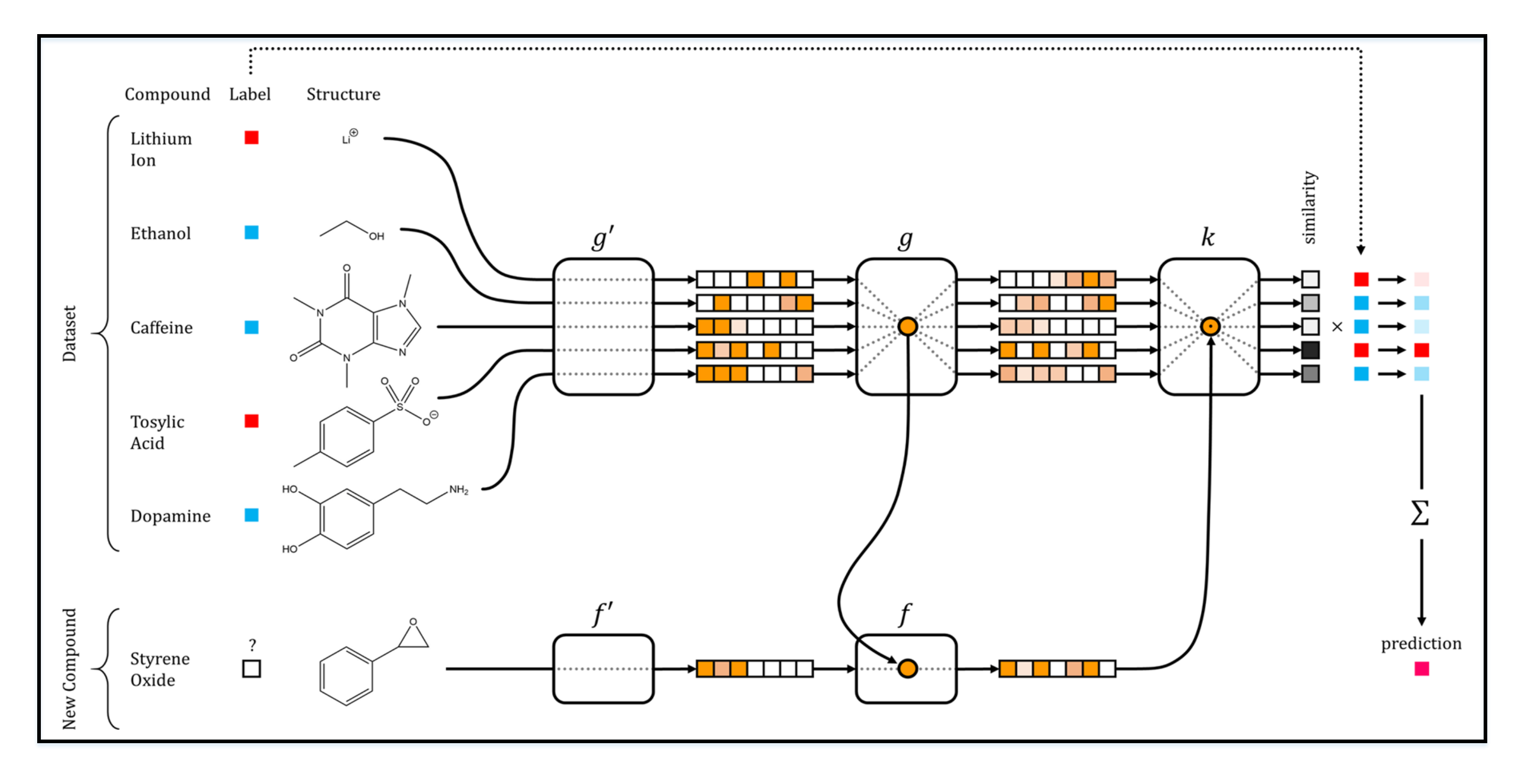}}
  \caption{\small Examples of metric learning in SSL.
  (a) Siamese Networks for Chromosome Classification(image is reproduced from \citep{gupta2017siamese}). Siamese networks are composed of two same neural networks, i.e, Base CNN, with shared parameters. In training stage 1, input is a similar/dissimilar pair, and the model parameters are learnt by optimizing the contrastive loss function. In training stage 2 and testing stage, a k-nearest neighbour(KNN) approach is utilized in the embeddings space learnt by the base CNN. (b) Matching Network for Low Data Drug Discovery (image is reproduced from \citep{altae2017low}). The core idea of this method is to use an attLSTM to generate both query embedding $f$ and support embedding $g$ that embed input examples (in small molecule space) into a continuous representation space. Then based on initial embeddings $f'$ and $g'$, it can construct $f$ and $g$ through iteratively evolving both embeddings simultaneously using a similarity measure $k(¡¤,¡¤)$, where the support embedding $g$ defines $f$. Finally, the prediction can be casted by the siamese one-shot learning problem.}
\end{figure}

The pioneer work was proposed by \citep{wolf2009one}, which applied One-Shot Similarity (OSS) measure as a kernel basis used with SVM, learning a similarity kernel for image classification of insects. \cite{wan2013one} proposed a new spatio-temporal feature representation (3D EMoSIFT) by fusing RGB-D data, which was invariant to scale and rotation, and then used nearest neighbor classifier for one-shot learning gesture recognition. Benefited from deep neural networks, the deep metric learning is gradually more popular, which explicitly learns a nonlinear mapping to map data points into a new feature space by exploiting the architecture of deep neural networks. The main techniques are siamese networks~\citep{bromley1994signature} and triplet networks~\citep{hoffer2015deep}.
In \cite{koch2015siamese},  powerful discriminative features were generalized for one-shot image recognition without any retraining, which were learned via a supervised metric-based approach with siamese neural networks for the first time. A similar architecture proposed in \citep{hilliard2017dynamic} could handle arbitrary example sizes dynamically as the system was used.
As shown in Fig.\ref{fig:metrica}, \cite{gupta2017siamese} augmented vanilla siamese networks for chromosome classification. In order to reduce the dependency of using actual class labels annotated by human experts, \cite{chung2017learning} proposed a deep siamese CNN to learn fixed-length latent image representation from solely image pair information. Alternatively, the siamese network methods emphasize less to the inter-class and intra-class variations, and thus \cite{ye2018deep} developed deep triplet ranking networks for one-shot image classification with larger capacity in handling inter- and intra-class image variations.  The triplet ranking loss can separate the instance pair that belongs to the same class from the instance pair that belongs to different classes in the relative distance metric space computed from the image embeddings. Furthermore, \cite{dong2017quadruplet} tried to add more instances into a tuple, and connected them with a novel loss combining a pair-loss and a triplet based contractive-loss.

Note that the above deep metric learning approaches do not offer a natural mechanism to solve $K$-shot $N$-way tasks (recognize $N$ objects with $K$ samples) for $K > 1$ and just focus on one-shot learning.
Recently, some novel metric learning methods for SSL have been proposed to tackle the more general few-shot learning problem, namely, matching networks~\citep{vinyals2016matching}, prototypical networks~\citep{snell2017prototypical}, relation network~\citep{sung2017learning}. Typically,
\cite{vinyals2016matching} learned a network called matching networks with an episodic training strategy. In each episode, the algorithm learns the embedding of the few labeled examples (the support set) to predict classes for the unlabeled points (the query set). Mathematically, we denote the support set $\mathcal{S}$, query set $\mathcal{Q}$, and $\mathcal{S}$ containing $N_{\mathcal{S}}$ (e.g., 1 or 5) exemplar images per category. The query set $\mathcal{Q}$ is coupled with $\mathcal{S}$ (has the same categories), but has no overlapped images. Each category of $\mathcal{Q}$ contains $N_{\mathcal{Q}}$ query images. During training, $\mathcal{S}$ will be fed into the to-be-learned embedding function $\mathcal{F}$ to generate the category classifiers $f_{\mathcal{S}}$. Then, $f_{\mathcal{S}}$ is subsequently applied to $\mathcal{Q}$ for evaluating the classification loss. The training objective then amounts to learning the embedding function by minimizing the classification loss. This process can be mathematically expressed as follows:
\begin{align}
  \min_{w} \mathbb{E}_{(\mathcal{S},\mathcal{Q})}\{\mathcal{L}(f_{\mathcal{S}}\circ \mathcal{Q})\},
\end{align}
where $w$ denotes the model parameters of the embedding function $f_{\mathcal{S}}$, and $\mathcal{L}$ is the loss function. $(f_{\mathcal{S}}\circ \mathcal{Q})$ denotes applying the category classifiers $f_{\mathcal{S}}$ on the query set $\mathcal{Q}$.
The purpose of episodic training is to mimic the real test environment containing few-shot support set and unlabeled query set, whose process can be viewed as meta-training (Section \ref{L2L}). The consistency between training and test environment alleviates the distribution gap and improves generalization, capable of obtaining state-of-the-art performance on a variety of one-shot classification tasks. Then $f_{\mathcal{S}}$ can be interpreted as a weighted nearest-neighbor classifier. To enhance the capacity of memory,  \cite{cai2018memory} further incorporated memory module into matching networks learning process, additionally integrating the contextual information across support samples into the deep embedding architectures. Some works extended matching networks~\citep{vinyals2016matching} to various applications, like low data drug discovery~\citep{altae2017low} (see Fig.\ref{fig:metricb}), video action recognition~\citep{kim2017matching}, one-shot part labeling~\citep{choi2017structured} and one-shot action localization~\citep{yang2018one}.
Alternatively, \cite{snell2017prototypical} established prototypical networks to learn a metric space where classification could be performed by computing distances to prototype representations of each class. Further, \cite{fortgaussian2017gaussian} improved prototypical networks architecture with interpretation of encoder outputs and construction way of metric on the embedding space. As an extension of \citep{vinyals2015show,snell2017prototypical}, \cite{sung2017learning} provided a learnable rather than fixed metric, or non-linear rather than linear classifier. Based on \citep{sung2017learning}, \cite{long2018object} learned an object level representation and exploited rich object-level information to infer image similarity.

Other related methods are developed in \citep{triantafillou2017few,oreshkin2018tadam,scott2018adapted}. Specifically, \cite{triantafillou2017few} adopted an information retrieval perspective on the problem of few-shot learning, i.e., each point acted as a `query' that ranked the remaining ones based on its predicted relevance to them. The mean average precision objective function was used that aimed to extract as much information as possible from each training batch by direct loss minimization over all relative orderings of the batch points simultaneously. To find more effective similarity measures for SSL, \cite{oreshkin2018tadam} proposed metric scaling and metric task conditioning to boost the performance of few-shot algorithms.
Furthermore, a hybrid approach was  proposed in \citep{scott2018adapted} to combine deep embedding losses for training (metric learning) on the source domain with weight adaptation (domain adaptation) on the target domain for $k$-shot learning.

\subsubsection{Knowledge-driven Small Sample Learning} \label{kdssl}
From the perspective of traditional Bayesian, regularization can be considered as prior knowledge, which is often the purely subjective assessment for learning tasks of an experienced expert. Following this understanding, \cite{tenenbaum2011grow} showed that when humans or machines made inferences that went far beyond the data available, strong prior knowledge must be making up the difference. In the big data era, the prior has more extensive meanings, like knowledge extracted from the environment, events and activities.  This knowledge may contain prior of learning tasks~\citep{stewart2017label}, domain knowledge~\citep{pan2010survey} or side information~\citep{vapnik2009new}, human/world knowledge~\citep{Lake2016,song2017machine} (e.g., human-level concepts~\citep{Lake2015}, common sense~\citep{davis2015commonsense}, and intuitive physics ~\citep{smith2013sources,battaglia2016interaction,hamrick2017metacontrol}), and so on.

In the pioneer work in \citep{Fei-Fei2006}, authors verified that learning need not start from scratch, while a key insight was that knowledge of previously learned classes could be considered as prior knowledge.
Recently, bringing specific domain knowledge to learning tasks has been causes widespread attention, such as physical laws~\citep{stewart2017label,battaglia2013simulation}, low rank, sparsity, side information~\citep{vapnik2009new}, domain noise distribution~\citep{xie2017robust} and so on.
For example, \cite{stewart2017label} introduced a new method for using physics and other domain constraints to supervise neural networks, detecting and tracking objects without any labeled examples.
To fuse side information into data representation learning, \cite{tsai2017improving} introduced two statistical approaches to improve one-shot learning.
Alternatively, \cite{ji2017combining} aimed to identify the related prior knowledge from different sources and to systematically encode them into visual learning tasks though joint bottom-up and top-down inference. Specifically, they demonstrated how to identify permanent theoretical knowledge and circumstantial knowledge for different vision tasks and how to represent and integrate them with the image data, maintaining good recognition performance and excellent generalization ability with minimal or even no data. On the other hand, some novel theories of incorporating knowledge by Bayesian methods have been developed recently. Regularized Bayesian inference~\citep{zhu2017big} improved the flexibility of Bayesian framework via posterior regularization, providing a novel approach to  incorporate knowledge. Another attempt called Bayesian deep learning \citep{wang2016towards} integrated deep learning and Bayesian models within a
principled probabilistic framework. In this unified framework, the interaction between data-driven deep learning and knowledge-driven Bayesian learning creates synergy and further boosts the performance.

Through employing domain knowledge, another research topic draws attention on unsupervised feature learning \citep{srivastava2015unsupervised} and self-supervised feature learning~\citep{pathak2016context}. Unsupervised feature learning aims to learn video representations to generate future target sequence by learning from the historical frames, where spatial appearances and temporal variations are two crucial structures. Sometimes CNN-based networks can predict one frame at a time and generate future images recursively, which are prone to focus on spatial appearances but RNN-based networks focus on temporal dynamics. Thus Convolutional LSTM (ConvLSTM) model becomes popular \citep{xingjian2015convolutional,lotter2016deep,villegas2017decomposing,wang2017predrnn}. Self-supervised feature learning learns invariance features, which does not require manually annotations (human intervention) but is still utilized in supervised learning by inferring supervisory signals from data structure. Recent methods mainly employ context information. For example, \cite{doersch2015unsupervised} explored the spatial consistency of image as context prediction task to learn feature representation. Further, \cite{noroozi2016unsupervised} created an extension by solving the jigsaw configuration. Alternatively, Temporal ordering of patches were investigated in \citep{lee2017unsupervised}. Recently, \cite{nathan2018improvements} developed a set of methods to improve performance of self-supervised learning using context.

Researches on human/world knowledge are popular for cognitive science to inspire SSL strategies. Here we will review some popular progresses on causality~\citep{zhang2017learning} and compositionality~\citep{Lake2016}, attention mechanism~\citep{desimone1995neural} and curiosity~\citep{gottlieb2013information}.

\paragraph{Causality and compositionality.} Causality is about using knowledge of how real world processes produce perceptual observations, which helps influence how people learn new concepts. Compositionality allows for reuse of a finite set of primitives (addressing the data efficiency) across many scenarios by recombining them to produce an exponentially large number of novel yet coherent and potentially useful concepts (addressing the overfitting problem). Benefiting from the ideas of causality and compositionality, some novel models have been proposed to perform SSL. For example, a representative work is like \citep{Lake2015}. They established the framework of Bayesian program learning (BPL) to mimic human writing, which inferred the next stroke from the current stroke using causality and compositionality, achieving one-shot generating characters.
Another work was investigated in \citep{George2017}. They established a generative vision model that were compositional, factorized, hierarchical, and flexibly queryable, achieving excellent generalization and occlusion-reasoning capabilities, and outperformed deep neural networks on a challenging scene text recognition benchmark while 300-fold more data efficient.
Alternatively,
\cite{higgins2017scan} described a new framework of symbol-concept association network (SCAN), able to discover and learn an implicit hierarchy of abstract concepts from as few as five symbol-image pairs per concept. Crucially, SCAN can imagine and learn novel concepts that have never been experienced during training with compositional abstract hierarchical representations. Through assuming that complex visual concepts could be composed using primitive visual concepts,
\cite{misra2017red} presented an approach to compose classifiers to generate classifiers for new complex concepts.

\paragraph{Attention.}
Attention describes the tendency of visual processing to be confined largely to stimuli that are relevant to behavior (addressing the data efficiency). This topic has become an active research in image capationing~\citep{xu2015show}, image generation~\citep{gregor2015draw}, VQA~\citep{xiong2016dynamic}, machine translation~\citep{bahdanau2014neural,johnson2016google}, and speech recognition~\citep{chorowski2015attention}. Specifically,
\cite{gregor2015draw} began the early work in small sample learning with the deep recurrent attentive writer (DRAW) neural network architecture for image generation, where attention helped the system to build up an image
incrementally, attending to one portion of a ``mental canvas" at a time.
Moreover, \cite{rezende2016one} developed new deep generative models building on the principles of feedback and attention, which could generate compelling and diverse samples after observing new examples just once.
Likewise, \cite{johnson2016google} utilized a single neural machine translation (NMT) model with attention module to translate between multiple languages achieving zero-shot translation.
Recently, \cite{wang2017multi} designed a neural network, that took the semantic embedding of the class tag to generate attention maps and used those attention maps to create the image features for one-shot learning.
Besides, \cite{he2017single} presented a fast yet accurate text detector with attention mechanism encoding strong supervised information of text in training that predicted word-level bounding boxes in one shot.

\begin{figure}
  \centering
   \includegraphics[width=6in]{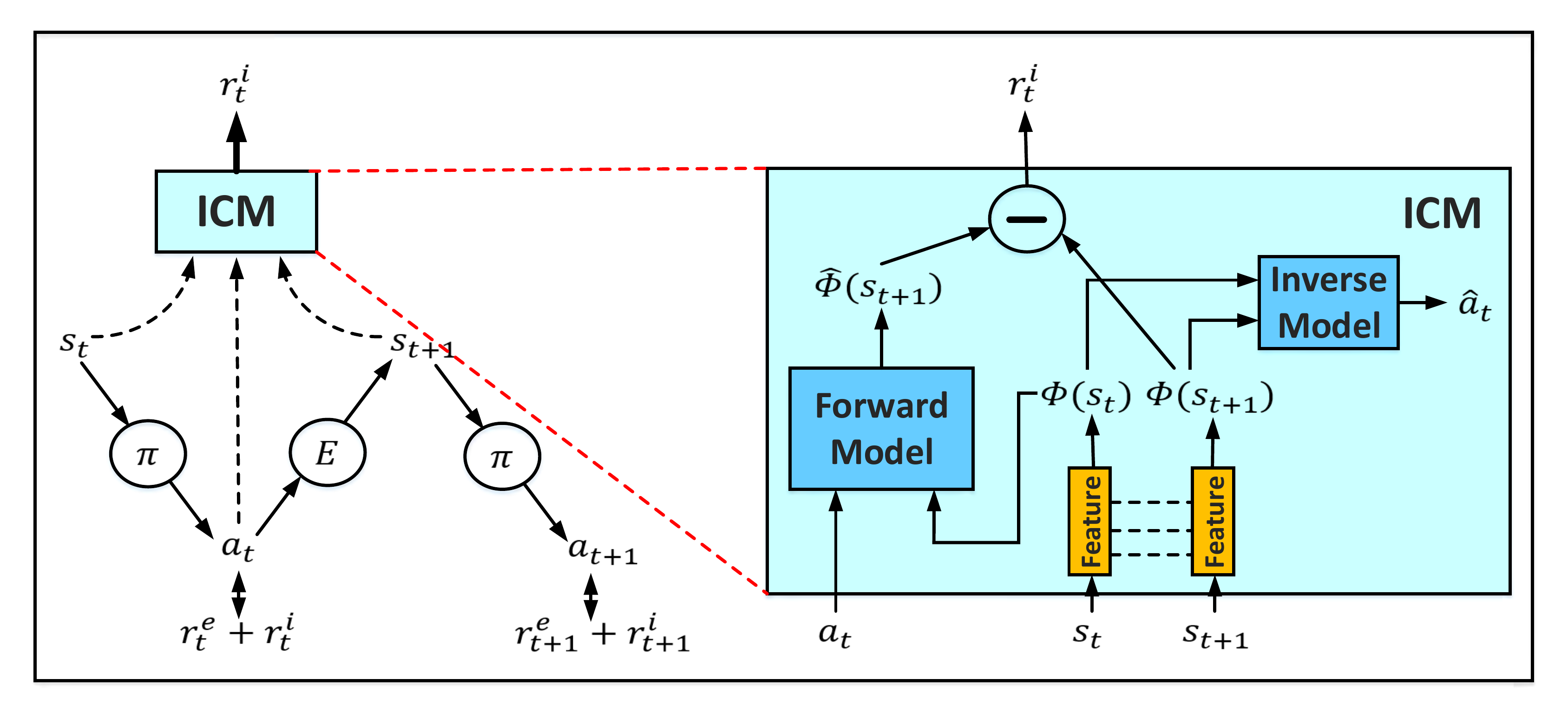}\vspace{-8mm}
  \caption{Illustration for Curiosity-Driven Exploration (image is reproduced from \citep{pathak2017curiosity}).
   The agent in state $s_t$ executes two subsystems: a reward generator (intrinsic curiosity module, ICM) that outputs a curiosity-driven intrinsic reward signal $r_t^i$ and a policy $\pi$ that outputs actions $a_t$ to maximize that reward signal. In addition to intrinsic rewards $r_t^i$, the agent optionally may also receive some extrinsic reward $r_t^e$ from the environment. ICM encodes the states $s_t,s_{t+1}$ to predict $\hat{a}_t$ by Inverse Model, and the prediction error between feature representation $\Phi(s_{t+1})$ of $s_{t+1}$ and $\hat{\Phi}(s_{t+1})$ produced by Forward Model is used as the curiosity based intrinsic reward signal. }\label{figcurious}
\end{figure}

\paragraph{Curiosity.}
The role of curiosity has been widely studied in the context of solving tasks with sparse rewards~\citep{gottlieb2013information,hester2017intrinsically}.
In general, a learning agent with curiosity can explore its environment in the quest for new knowledge and  learning skills that might be helpful in future scenarios with rare or deceptive rewards. \cite{pathak2017curiosity} firstly proposed a mechanism for generating curiosity-driven intrinsic reward signal that scales to high-dimensional continuous state spaces like images, achieving gradually learning more and more complex skills with few data rewards or even no data rewards(see Fig.\ref{figcurious}). 
Curiosity-driven exploration system, sometimes also called intrinsic motivation system~\citep{oudeyer2016intrinsic}, and recently \cite{forestier2017intrinsically} have presented intrinsically motivated goal exploration processes (IMGEP) algorithmic approach to establish unsupervised multi-goal reinforcement learning formal framework. Further, \cite{pere2018unsupervised} extended IMGEP to add a unsupervised goal space learning stage (UGL), where an unsupervised representation learning algorithm was used to learn a lower-dimensional latent space representation, and then the representation was applied to a standard IMGEP. Readers are suggested to read \citep{oudeyer2018computational} to review computational frameworks and theories of curiosity-driven learning.

\subsection{Approach 4: Meta learning}\label{L2L}

The explanation of meta learning in computer science on Wikipedia is to use metadata to understand how automatic learning can become flexible in solving learning problems, and hence to improve the performance of existing learning algorithms or to learn (induce) the learning algorithm itself~(\url{https://en.wikipedia.org/wiki/Meta_learning_(computer_science)}). In other words, meta learning helps a learning system capable of learning to learn by itself, which can achieve the adaptive perception and cognition of the environment. From the cognition learning perspective, one way human acquires prior knowledge (Section \ref{kdssl}) is through meta learning with a long research history \citep{harlow1949formation,thrun2012learning}.
Meta learning works through learning the common/shared methodology in accomplishment of a family of tightly related tasks in current researches, which sometimes is closely related to the machine learning notions of transfer learning~\citep{pan2010survey} or multi-task learning~\citep{zhang2017overview}. Then the common/shared methodology can very easily adapt the novel task as much stronger prior, while this prior is learning to learn by itself, which forces the learning system to learn new tasks as rapidly and
flexibly as humans do. For example, \cite{Lake2015} proposed Bayesian program learning (BPL) to develop hierarchical priors that allowed previous experience with related concepts to ease learning of new concepts. Specifically, meta learning helps BPL learn the generation process of handwritten characters, which is the common/shared methodology to understand and explain characters. Therefore when encountering novel types of handwritten characters, BPL can perform one-shot learning in classification tasks at human-level accuracy. Also, if the common/shared methodology corresponds to other types of symbolic concepts/knowledge, the learning system can perform broader tasks, which may be particularly promising (more detailed discussions are described in Section \ref{metassl}).

Meta learning takes an important role in SSL and artificial intelligence~\citep{Lake2016}, and recent researches include learning to learn~\citep{santoro2016meta}, learning to reinforcement learn \citep{wang2016learning,duan2016rl,xu2018learning}, learning to transfer~\citep{Ying2018transfer}, learning to optimize~\citep{li2016learning,Rosenfeld2018Learning}, learning to infer~\citep{hu2017learning,marino2018learning}, learning to search~\citep{guez2018learning,balcan2018learning}, learning to control~\citep{duan2017meta} and so on. In the following we will review some typical methods in SSL along this research line.

\begin{figure}
  \centering
  \subfigure[Meta Networks]{
  \label{figmetaa}
  \includegraphics[width=2.7in]{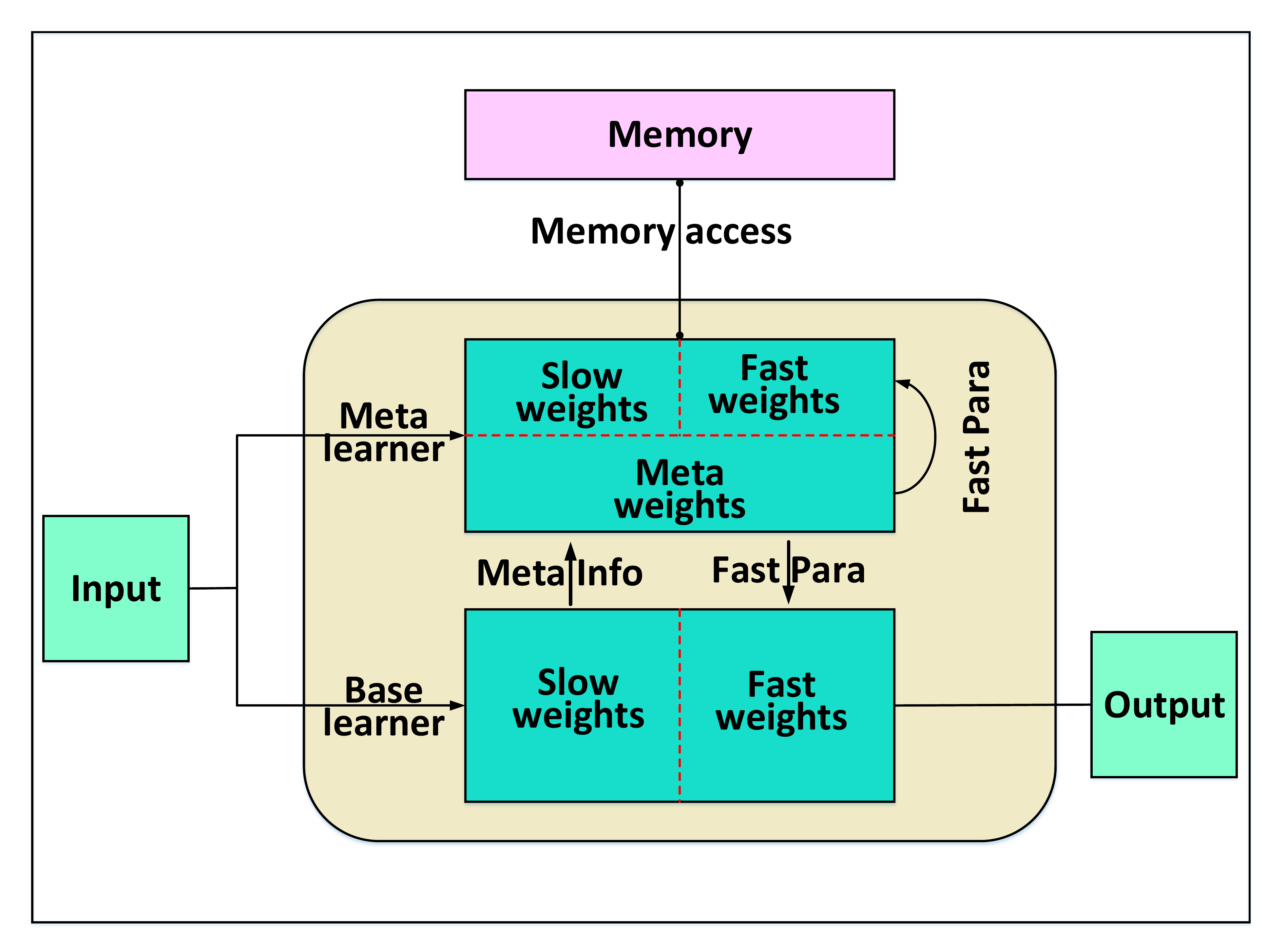}}
  \hspace{0.1in}
  \subfigure[Meta-Critic Networks]{
  \label{figmetab}
  \includegraphics[width=2.7in]{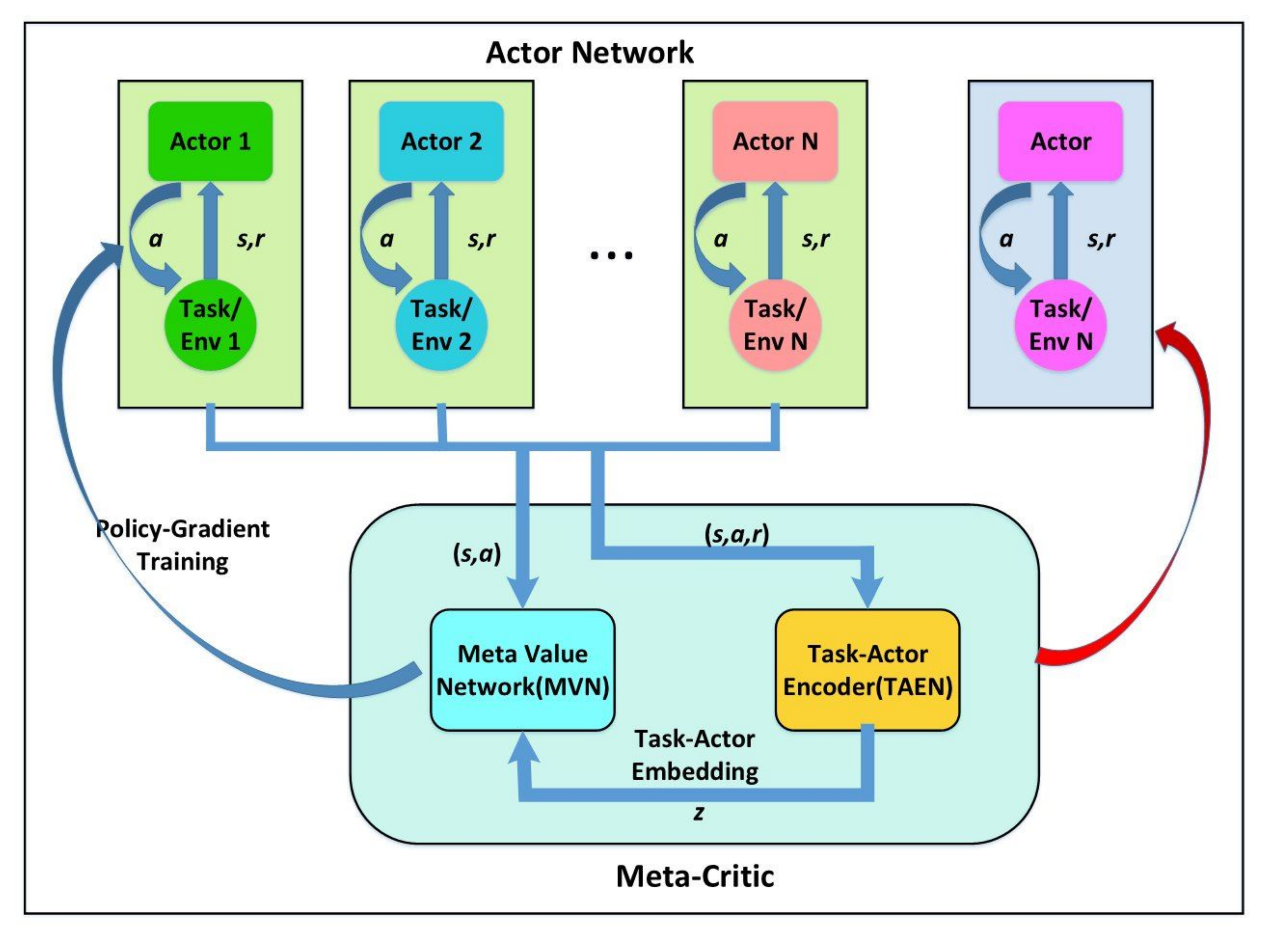}}
  \caption{Examples of architectures in meta learning (images are reproduced from \citep{munkhdalai2017meta} and \citep{sung2017learning}). (a) Meta networks. MetaNet consists of two main learning components, a base learner and a meta learner, and is equipped with an external memory. (b) Meta-critic networks. There exists a meta-learner--meta-critic network to guide actor network for each task, which is composed of two parts: meta value network (MVN) and task-actor encoder (TAEN). TAEN outputs task-actor embedding $z$ with state ($s$), action ($a$), reward ($r$) from multi-tasks simultaneously. Then $z$ and $a,r$ are delivered to MVN being trained to model the return of policy. When encountering new tasks, we can fix MVN whereas establish a new actor network to rapidly learn by leveraging learning critic.}
\end{figure}

\paragraph{Learning to learn.}  This strategy aims to adaptively determine the appropriate data, loss function, and hypothesis space in a machine learning model in meta-level learning manner. The pioneer work began at \citep{santoro2016meta}, which used LSTM and MANN meta-learners to learn quickly from data presented sequentially, binding data representations to their appropriate labels. The general scheme to map data representations to appropriate classes or function values boosts few-shot image classification performance. While \cite{vinyals2016matching} treated the data as a set, their episodic training strategy helped mimic the real test environment containing few-shot support set and unlabeled query set.
Also, they combined metric learning to judge image similarity, and similar strategy was employed in \citep{snell2017prototypical,sung2017learning}. \cite{romero2014fitnets} further extended \citep{snell2017prototypical} with unlabeled examples training.
Another attempt learning a meta-level network was made in \citep{wang2016learning}. The network operated on the space of model parameters, which was specifically trained to regress many-shot model parameters (trained on large datasets) from few-shot model parameters (trained on small datasets).
Recently, \cite{munkhdalai2017meta} proposed a meta networks (MetaNet), as shown in Fig.\ref{figmetaa}, where the base learner performed in the input task space whereas the meta learner operated in a task-agnostic meta space. The meta learner can continuously learn and perform meta knowledge acquisition across different tasks. When novel tasks coming, the base learner first analyzes the task, and then provides the meta learner with a feedback in the form of higher order meta information (knowledge) to explain its own status in the current task space. Based on the meta information, the meta learner rapidly parameterizes both itself and the base learner so that the MetaNet model can recognize the new concepts rapidly.
To exploit the domain-specific task structure, \cite{mishra2017simple} proposed a class of simple and generic meta-learner architectures combining  temporal convolutions and soft attention.
A  new framework called learning to teach was proposed by \cite{fan2018learning}. The teacher model leveraged the feedback from the student model to determine the appropriate data, loss function, and hypothesis space to facilitate the training of the student model. This technique can achieve almost the same accuracy as full-supervised training using much less training data and fewer iterations.

\paragraph{Learning to reinforcement learn.} \citep{wang2016learning,duan2016rl} firstly introduced meta learning into reinforcement learning (RL) to realize deep meta-reinforcement learning, motivated by developing deep RL methods that could adapt rapidly to new tasks.
In particular, learner used deep RL to train a recurrent network on a series of interrelated tasks, with the result that the network dynamics learned a second RL procedure which operated on a faster time-scale than the original algorithm. Some typical methods are introduced as follows.
\cite{sung2017learning} proposed to learn a meta-critic network that could be used to train multiple `actor' networks to solve specific problems, and the shared meta-critic provided the transferrable knowledge that allows actors to be trained with only a few trials on a new problem (see Fig.\ref{figmetab}). For hierarchically structured policy learning, \cite{frans2017meta} employed  shared primitives (sub-policies) to improve sample efficiency on unseen tasks. A generic neural mechanism was introduced by \cite{munkhdalai2017learning} to meta learning called conditionally shifted neurons, which could modify activation values with task-specific shifts retrieved from a memory module with limited task experience. For continuous adaptation in non-stationary environments, \cite{al2017continuous} developed a gradient-based meta-learning approach suitable. They regarded non-stationarity as a sequence of stationary tasks and trained agents to exploit the dependencies between consecutive tasks such that they can handle similar non-stationarities at execution time. Unlike model-free RL~\citep{wang2016learning,duan2016rl,sung2017learning,al2017continuous}, \cite{clavera2018learning} considered to learn online adaptation in the context of model-based reinforcement learning. They trained a global model such that, when combined with recent data, the model could be rapidly adapted to the local context.

\paragraph{Learning to transfer.} Meta-learner is able to implement a learning task on a large number of different tasks to automatically determine what and how to transfer should be appropriate.
Inspired by this capability, the model-agnostic meta-learning (MAML) approach~\citep{finn2017model} was proposed aiming to meta-learn an initial condition (set of neural network weights) that was suitable for fine-tuning on few-shot problems under model and task-agnostic conditions. Furthermore, \cite{kim2018bayesian} extended \citep{finn2017model} by introducing Bayesian mechanisms for fast adaptation and meta-update, quickly obtaining an approximate posterior of a given unseen task, as well a probabilistic framework was developed by~\citep{finn2018probabilistic}. To avoid a biased meta-learner like \citep{finn2017model},
\cite{jamal2018task}  proposed a task-agnostic meta-learning (TAML) algorithms to train a meta-learner unbiased towards a variety of tasks before its initial model was adapted to unseen tasks. To deal with the long-tail distribution in big data, \cite{wang2017learning} introduced a meta-network that learned to progressively transfer meta-knowledge from the head to the tail classes, where meta-knowledge was encoded with a meta-network trained to predict many-shot model parameters from few-shot model parameters.
Another attempt to learn to transfer across domains and tasks was made by \citep{hsu2017learning}. They learned a pairwise similarity (i.e., meta-knowledge) to perform both domain adaptation and cross-task transfer learning, which was realized using a neural network trained by using the output of the similarity.
For model agnostic training procedure,
\cite{li2017learning} made meta-learning on simulated train/test split with domain-shift for domain generalisation, which could be applied to different base network types.
The latest work was investigated by \citep{Ying2018transfer}. They proposed a framework of learning to transfer (L2T) to enhance transfer learning effectiveness by leveraging previous transfer learning experiences. In particular, L2T learns a reflection function mapping a pair of domains and the knowledge transferred between them to the performance improvement ratio. When a new pair of domains arrives, L2T optimizes what and how to transfer by maximizing the value of the learned reflection function.

\begin{figure}
  \centering
  \subfigure[Forward graph for the meta-learner]{
  \label{figmeta1a}
  \includegraphics[width=2.7in]{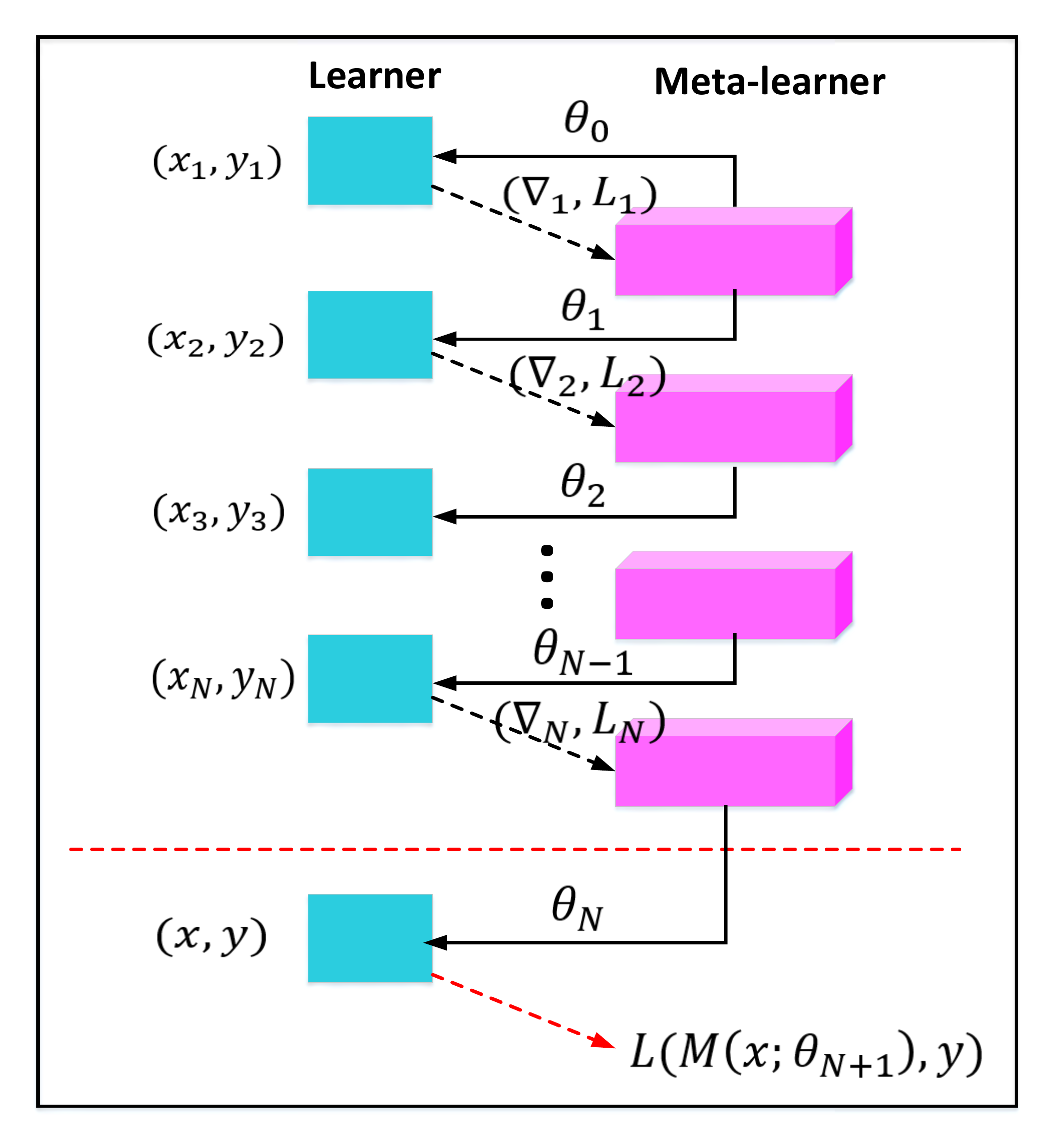}}
  \hspace{0.1in}
  \subfigure[Diagram of optimization process]{
  \label{figmeta1b}
  \includegraphics[width=2.7in]{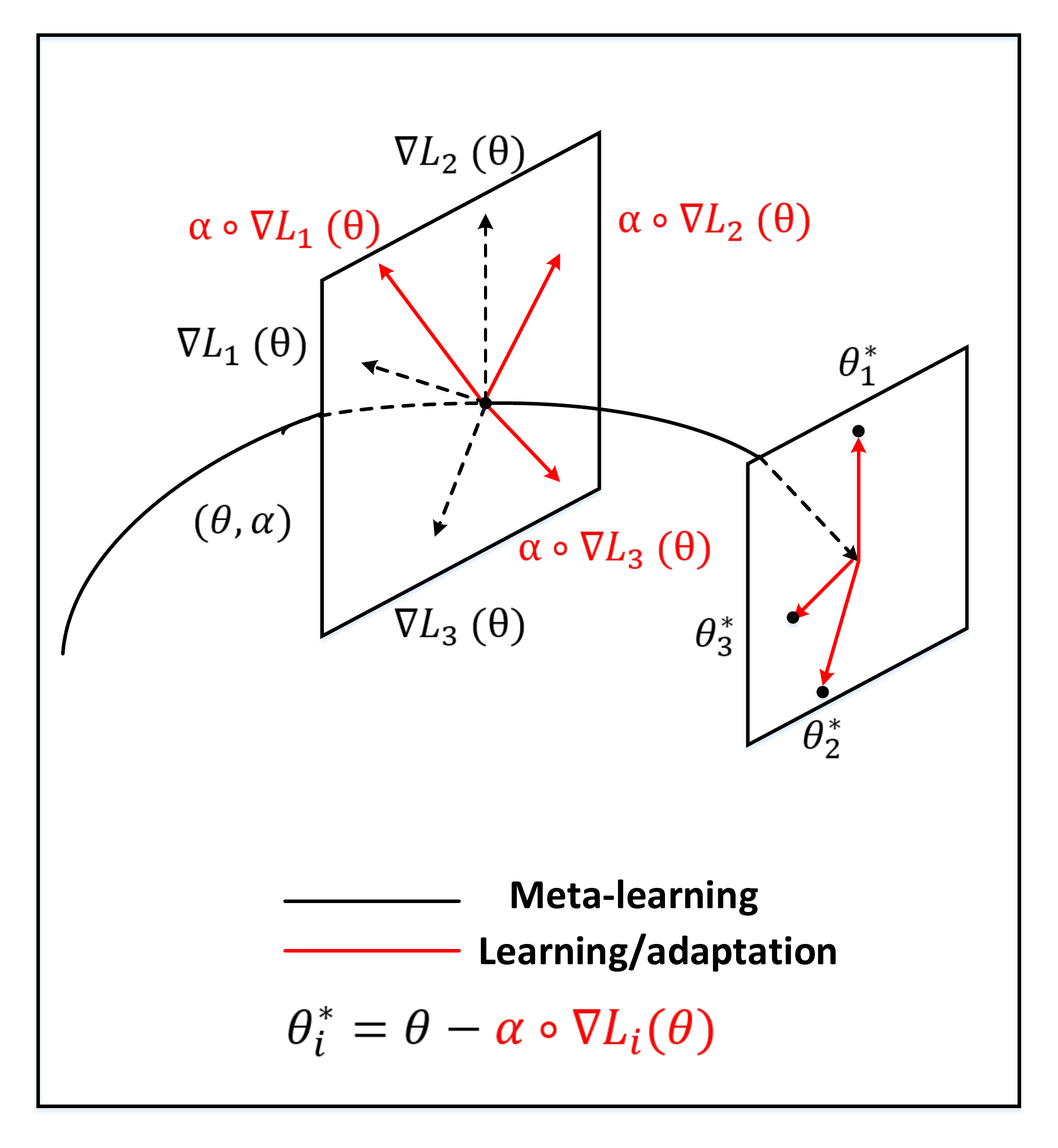}}
  \caption{Illustrations of learning process in meta learning (images are reproduced from \citep{ravi2016optimization,li2017learning}). (a) Forward graph for the meta-learner. The red dashed line divides examples from the training set $D_{train}$ and test data $D_{test}$. Each $(x_i,y_i)$ is the $i^{th}$ batch from the training set whereas $(x,y)$ is all the elements from the test set. The dashed arrows indicate that we do not back-propagate through that step when training the meta-learner and red dashed arrow implies that learner performs rapid adaptation in the test stage. Here we use $\nabla_t$ as a shorthand for $\nabla_{\theta_{t-1}}L_t$. (b) Diagram of optimization process. Gradual learning aims to learn the meta-learner across tasks in the meta-space $(\theta,\alpha)$. Rapid learning is performed by the meta-learner in the learner space $\theta$ that learns task-specific learners.}\label{metaprocess}
\end{figure}

\paragraph{Learning to optimize.} Casting optimization algorithm design as a learning problem allows us to specify the class of problems we are interested in through data as well as automatic generate optimizers. The learning process is shown in Fig.\ref{metaprocess}.
\cite{bertinetto2016learning} firstly proposed a method to minimize a one-shot classification objective in a learning-to-learn formulation. Particularly, they optimized a pupil network through constructing the learner, called a learnet, which predicted the parameters of pupil network from a single exemplar. Followed by \citep{bertinetto2016learning}, \cite{ravi2016optimization} proposed an LSTM-based meta-learner model to learn the exact optimization algorithm used to train pupil neural network classifier in the few-shot regime as well as episodic training idea. Recently, some works have improved LSTM-based meta-learner model~\citep{ravi2016optimization}.
For example, \cite{finn2017model} proposed a model-agnostic meta-learning (MAML) learner, which was compatible with any model trained with gradient descent and applicable to a variety of different learning problems. Alternatively, \cite{li2017metasgd} insisted that the choice of meta-learners was crucial and developed an easily trainable meta-learner, Meta-SGD, that could initialize and adapt any differentiable learner in just one step. Notably, compared with LSTM-based learner~\citep{ravi2016optimization}, Meta-SGD was conceptually simpler, easier to be implemented, and could be learned more efficiently. To help learning to learn able to scale to larger problems and generalize to new tasks, \cite{wichrowska2017learned} introduced  a hierarchical RNN architecture ensemble of small and diverse optimization tasks capturing common properties of loss landscapes.
In theory, \cite{finn2018meta} stated that a meta-learner was able to approximate any learning algorithm in terms of its ability to represent functions of the dataset and test inputs independent of the type of meta-learning algorithm. Furthermore, a bridge between gradient-based hyperparameter optimization and learning to learn in some setting was explored by \citep{franceschibridge2017}.

\section{Beyond Small Sample Learning} \label{beyond}
In this section, we will introduce some research topics closely related to SSL, and discuss their relationships to SSL.

\subsection{Weakly-Supervised Learning} \label{WSL}
Different from SSL with few annotated samples, weakly-supervised learning usually contains more annotated information, while is coarse-grained or noisy, whose supervised information is inexact, inaccurate or incomplete~\citep{zhou2017brief}. For example, semantic segmentation needs pixel-wise labels for supervised learning, while collecting large-scale annotations is significantly labor intensive and limited for some applications. To alleviate this annotation quality issue and make semantic segmentation more scalable and generally applicable, weakly supervised learning has attracted much attention recently. The challenge in this issue is that weak labels provide part (or even inaccurate) information of the supervision~\citep{hong2017weakly}, such as image-level label, bounding box, point supervision, scribble, and so on. To reduce human intervention required for training further, some approaches design exploitation regimes of an additional source of data. For example, \cite{hong2017weakly2} made use of web videos as additional data, while the annotations of web videos returned by a search engine tend to be inevitably noisy since the query keywords may not be consistent with the visual content of target images, and thus the problem is evidently weakly supervised.
Sometimes, weakly-supervised information help boost performance of SSL. For example, \cite{Niu_2018_CVPR} recently designed a new framework, which can jointly leverage both web data and auxiliary labeled categories (zero-shot learning) for fine-grained image classification. Their model can tackle the label noise and domain shift issue to a certain extent.

\subsection{Developmental Learning and Lifelong Learning}
To avoid the issue of experience catastrophic forgetting~\citep{kirkpatrick2017overcoming}, human can learn and remember many different tasks that are encountered over multiple timescales. Recently, developmental learning~\citep{sigaud2016towards} and lifelong learning~\citep{thrun1995lifelong,mitchell2018never} try to alleviate this issue. SSL focused on learning with few observations, which sometimes meets the need of developmental learning and lifelong learning sometimes accounting for new tasks containing few data. On the other hand, the techniques and ideas of developmental learning and lifelong Learning may inspire SSL solving strategy, like fine-tuning (Section \ref{section311}). For example, \cite{kaiser2017learning} proposed life-long one-shot learning, which firstly tries to make deep models learn to remember rare events through their lifetime.

\subsection{Open Set Learning}
Open set learning was firstly proposed by \citep{scheirer2013toward}, which tries to identify whether the testing images come from the training classes or some unseen classes. Unlike zero-shot learning, open set learning does not need to explicitly predict the class labels. Recently, this setting is also known as incremental learning~\citep{rebuffi2017icarl}, where learning systems learn more and more concepts over time from a stream of data. This learning paradigm distinguishes with zero-shot learning in that it deploys data in a dynamic way. Furthermore, when we consider the data features/classes are both incremental and decremental, the setting is called online learning~\citep{hou2017one}. To summarise, open set learning can be considered as a specific SSL problem with certain constraints, because the novel classes are coming with few observations, while it is dynamic, evolving and infeasible to keep the whole data. Recently, \cite{busto2017open} explored the field of domain adaptation in open sets, which is called open set domain adaptation. In this setting, both source and target domain contain classes that are not interested, and the target domain contains classes not related to that in the source domain and vice versa.

\section{Further Research}\label{section6}
The research on SSL is just in its very beginning period. The current developments are still needed to be further improved, empirically justified, theoretically evaluated and capability extended. In this section, we will try to list some promising and challenging research directions worthy to be investigated in future research.

\subsection{More Neuroscience-Inspired Researches}
SSL stems from mimicking the mechanism of human being that recognizes and forms concepts. Though many works have focused on computer simulation of the human's mechanism, more intrinsic simulations deserve to be further explored. Especially, the following issues should be necessary to be considered:
\begin{itemize}
  \item Faster information process mechanism.
  \item SSL with capability of episodic memory and experience replay.
  \item Generating data as the cognitive manners like imagination, planning, and synthesis.
  \item Continual and incremental SSL regimes.
\end{itemize}

\subsection{Meta Knowledge \& Meta Learning} \label{metassl}
The core strategy of SSL consists in skillful use of the knowledge existed. The knowledge mostly used so far is concerned with the method itself for specific problem-solving, while not with the methodology of how to develop the methods. The latter knowledge is the knowledge in meta-level.  How to effectively use the meta-level knowledge in SSL deserves to be further studied.

Meta-learning aims at learning methodology of doing things (namely, learning to learn, optimize, transfer, and so on).  This should be one of the next important focuses in AI research. To achieve this goal may go through the following stages:
\begin{itemize}
  \item To accomplish a family of highly related tasks through learning the common methodology of solving the family of tasks.
  \item To accomplish a set of weakly related tasks through learning the common methodology of solving the family of tasks.
  \item To accomplish more general tasks through learning methodology of doing things.
\end{itemize}
Realizing the above stages needs a progressive efforts. The current developments may be mainly paid on realization of the first two stage goals.

\subsection{Concept Learning: Main Challenges}
It also a critical issue of how to build a universal (generally applicable) mapping from the visual (V, image) space to semantics space (S, concept space). The existing methods are far from satisfied, and the problem to transform from V to S is still very challenging, e.g., domain shift problems\citep{fu2014transductive} and hubness problems\citep{radovanovic2010hubs,dinu2014improving}.

Another challenge is the new concept learning problem. How to justify a new concept is being formed, and how to properly formalize its intensional and extensional representations. Such research is less so far but imperative, e.g., subtype discovery of cancer.

\subsection{Experience Learning: Main Challenges}
Cross-domain synthesis is the most attractive manner to form augmented data. How to realize a proper transformation of an object from one representation to the another (say, from CT to MR from visual to brain signal) is still a challenging issue. The differential homomorphism approach provides a promising mathematical framework for alleviating this issue, its effectiveness is, however, far from satisfied.

Model/knowledge/metric-driven learning provides promising ways to relax the dependence of LSL on amount of samples, while some problems still need to be considered:
\begin{itemize}
  \item  How to determine a proper family of models?
  \item  How to define, represent, and embed knowledge into a model?
  \item  How to design metric learning methods more suitable for SSL?
\end{itemize}

\subsection{Promising SSL Applications}
There are many attractive applications that SSL is hopeful to be generalized to use. Some typical cases include:
\begin{itemize}
  \item New drug discovery, human-machine interaction, subtype discovery of disease, and outlier detection;
  \item Fast cognition and recognition: the applications needed to perceive environment and react in real time;
  \item Unmanned system;
  \item Experience learning with few samples: medical aid diagnosis, intelligent communication~\citep{suh2016label}.
\end{itemize}

\section{Conclusion}\label{section7}

This paper has provides a comprehensive survey on the current developments on small sample learning (SSL). The existing SSL techniques can be divided into to main categories of approaches, including experience learning and concept learning. Both concepts, as well as SSL, have been finely explained in mathematics in the paper, and most typical methods along both lines of research have been comprehensively reviewed.
Besides, biology plausibility has been provided to support the feasibility of SSL.
Furthermore, the relationship of some current related methodologies with SSL has also been discussed, and some meaningful research directions of SSL have been introduced for future research.

%


\acks{This research was supported by the China NSFC projects under contracts 61661166011, 11690011, 61603292, 61721002.}


%
%
%
%
%
%

\vskip 0.2in
\bibliography{sample}

\end{document}